\documentclass[]{lowlevelbanana}
\definecolor{w_blue}{RGB}{52,204,204}
\definecolor{w_yellow}{RGB}{255,192,0}

\usepackage{amsfonts}
\usepackage{amsmath}
\usepackage{amssymb}

\usepackage{colortbl}

\usepackage{makecell}
\usepackage{multicol}
\usepackage{multirow}
\usepackage{pifont}

\definecolor{red}{rgb}{0.8,0,0}  
\definecolor{green}{RGB}{0, 133, 21}  
\definecolor{grey}{rgb}{0.5,0.5,0.5}

\definecolor{w_1}{RGB}{52,204,204}
\definecolor{w_2}{RGB}{70,203,187}
\definecolor{w_3}{RGB}{95,202,161}
\definecolor{w_4}{RGB}{119,200,136}
\definecolor{w_5}{RGB}{155,199,101}
\definecolor{w_6}{RGB}{185,197,70}
\definecolor{w_7}{RGB}{227,194,28}
\definecolor{w_8}{RGB}{243,193,12}
\definecolor{w_9}{RGB}{255,192,0}

\makeatletter
\def\blfootnote{\xdef\@thefnmark{}\@footnotetext}
\DeclareRobustCommand\onedot{\futurelet\@let@token\@onedot}
\def\@onedot{\ifx\@let@token.\else.\null\fi\xspace}

\makeatother

\def\eqref#1{Equation~\ref{#1}}

\usepackage{soul}

\usepackage{titletoc}
\usepackage{caption}

\usepackage{float}
\usepackage[utf8]{inputenc}
\usepackage{hyperref}

\title{Is Nano Banana Pro a Low-Level Vision All-Rounder? \\ A Comprehensive Evaluation on 14 Tasks and 40 Datasets}

\author[]{Jialong Zuo}
\author[]{Haoyou Deng}
\author[]{Hanyu Zhou}
\author[]{Jiaxin Zhu}
\author[]{Yicheng Zhang}
\author[]{Yiwei Zhang}
\author[]{Yongxin Yan}
\author[]{Kaixing Huang}
\author[]{Weisen Chen}
\author[]{Yongtai Deng}
\author[]{Rui Jin}
\author[]{Nong Sang}
\author[]{Changxin Gao}
\affiliation[]{National Key Laboratory of Multispectral Information Intelligent Processing Technology,\\ School of Artificial Intelligence and Automation, Huazhong University of Science and Technology}
\abstract{
The rapid evolution of text-to-image generation models has revolutionized visual content creation. While commercial products like Nano Banana Pro have garnered significant attention, their potential as generalist solvers for traditional low-level vision challenges remains largely underexplored. In this study, we investigate the critical question: \textbf{Is Nano Banana Pro a Low-Level Vision All-Rounder?} We conducted a comprehensive zero-shot evaluation across 14 distinct low-level tasks spanning 40 diverse datasets. By utilizing simple textual prompts without fine-tuning, we benchmarked Nano Banana Pro against state-of-the-art specialist models. Our extensive analysis reveals a distinct performance dichotomy: while \textbf{Nano Banana Pro demonstrates superior subjective visual quality}, often hallucinating plausible high-frequency details that surpass specialist models, \textbf{it lags behind in traditional reference-based quantitative metrics.} We attribute this discrepancy to the inherent stochasticity of generative models, which struggle to maintain the strict pixel-level consistency required by conventional metrics. This report identifies Nano Banana Pro as a capable zero-shot contender for low-level vision tasks, while highlighting that achieving the high fidelity of domain specialists remains a significant hurdle.
}

\metadata[
\raisebox{-0.2em}{\includegraphics[width=0.025\linewidth]{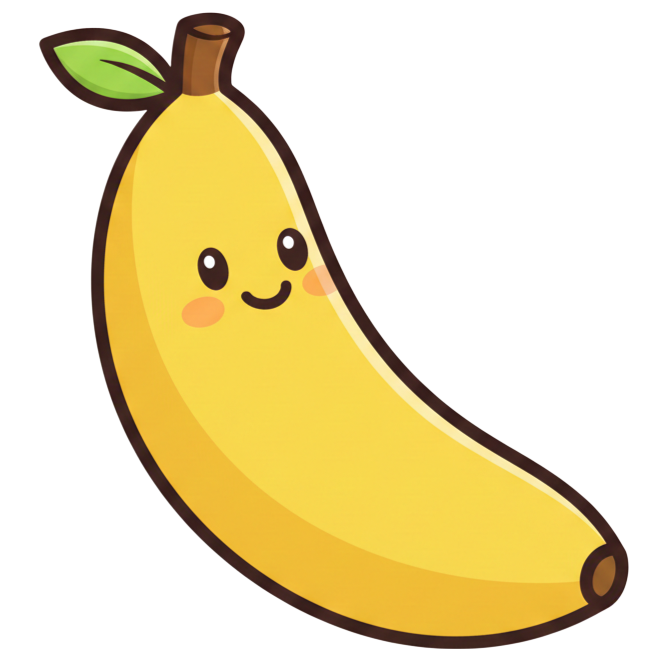}}~~Project Page]{\href{https://lowlevelbanana.github.io/}{\texttt{https://lowlevelbanana.github.io}}
\\[-1.5ex]}

\metadata[
\raisebox{-0.2em}{\includegraphics[width=0.025\linewidth]{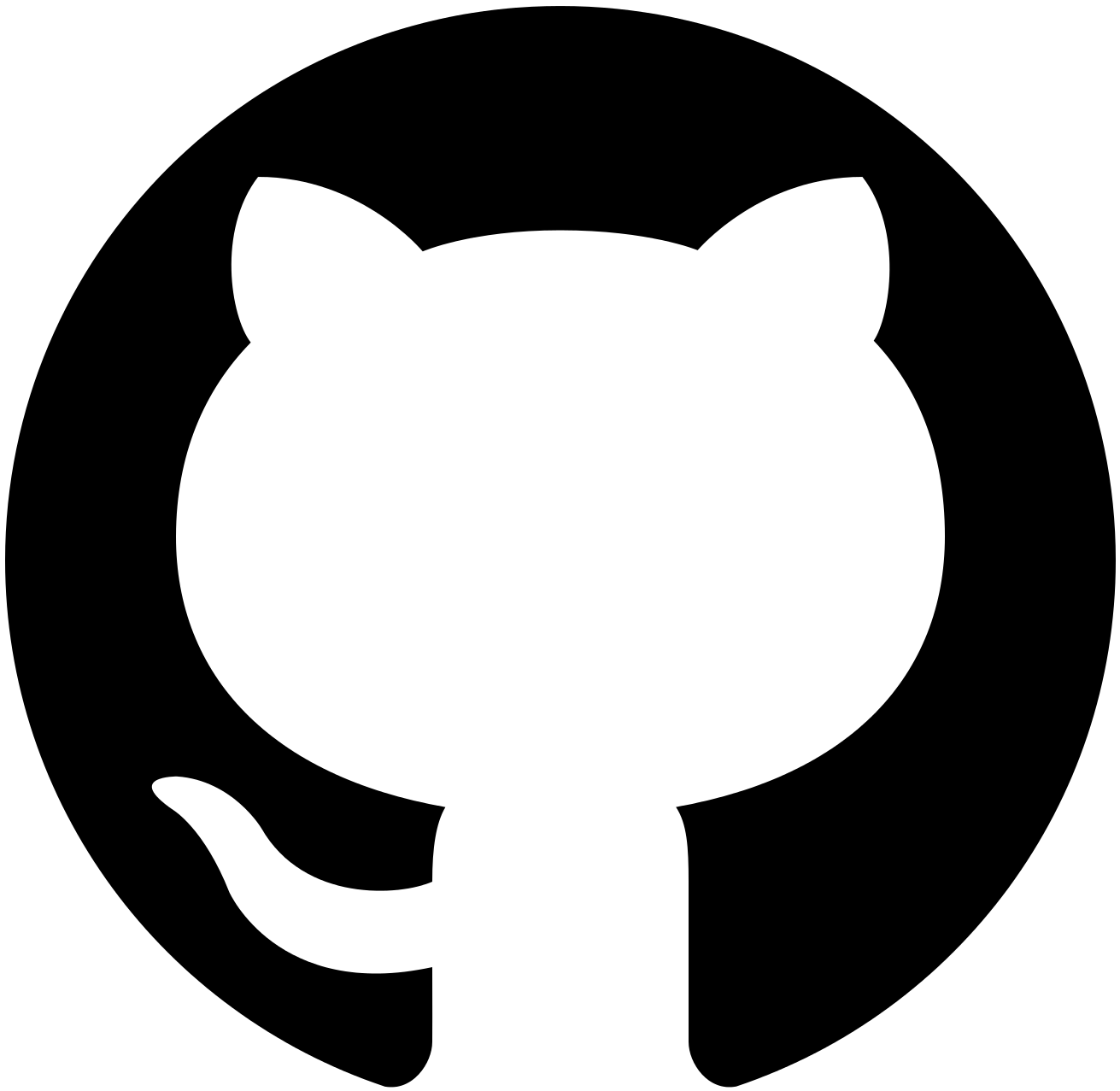}}~~GitHub]{\href{https://github.com/zplusdragon/LowLevelBanana}{\texttt{https://github.com/zplusdragon/LowLevelBanana}}
\\[-1.7ex]}

\metadata[
\raisebox{-0.3em}{\includegraphics[width=0.026\linewidth]{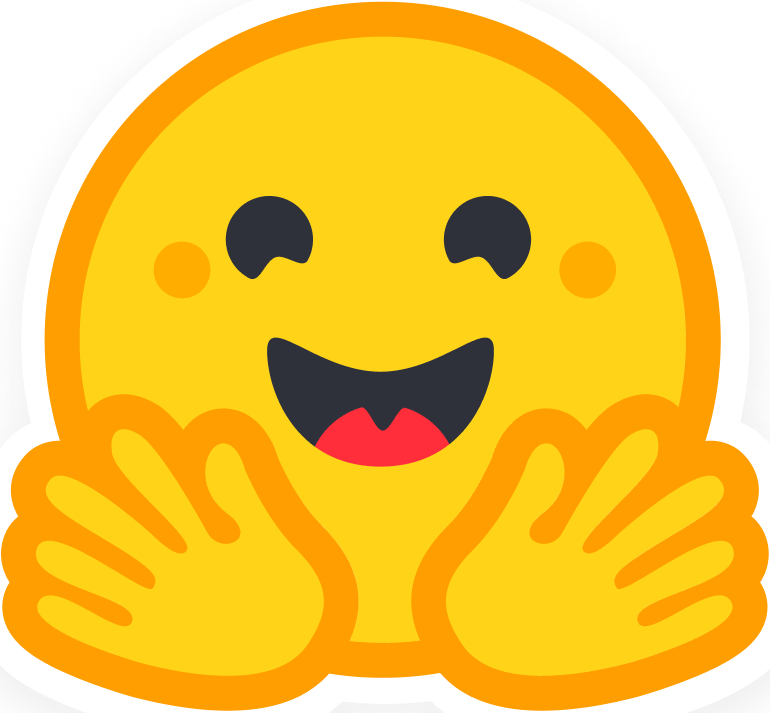}}~~HuggingFace Dataset]{\href{https://huggingface.co/datasets/jlongzuo/LowLevelEval}{\texttt{https://huggingface.co/datasets/jlongzuo/LowLevelEval}}}

\begin{document}

\maketitle

\section{Introduction}
\label{sec:intro}

The recent proliferation of Generative AI has fundamentally transformed the landscape of computer vision, with Text-to-Image (T2I) models demonstrating unprecedented capabilities in high-fidelity content creation. Among these, commercial products like Nano Banana Pro~\cite{team2023gemini} have emerged as standouts, garnering significant attention for their versatility. While its prowess in creative synthesis is well-documented, the extent to which such a large-scale foundation model can generalize to traditional low-level vision problems remains largely unexplored. This gap presents not only a challenge of capability but also one of evaluation, raising the pivotal research question: \textbf{Is Nano Banana Pro a Low-Level Vision All-Rounder?}

\begin{figure*}[htp]  
  \centering  
  \includegraphics[width=1.0\textwidth]{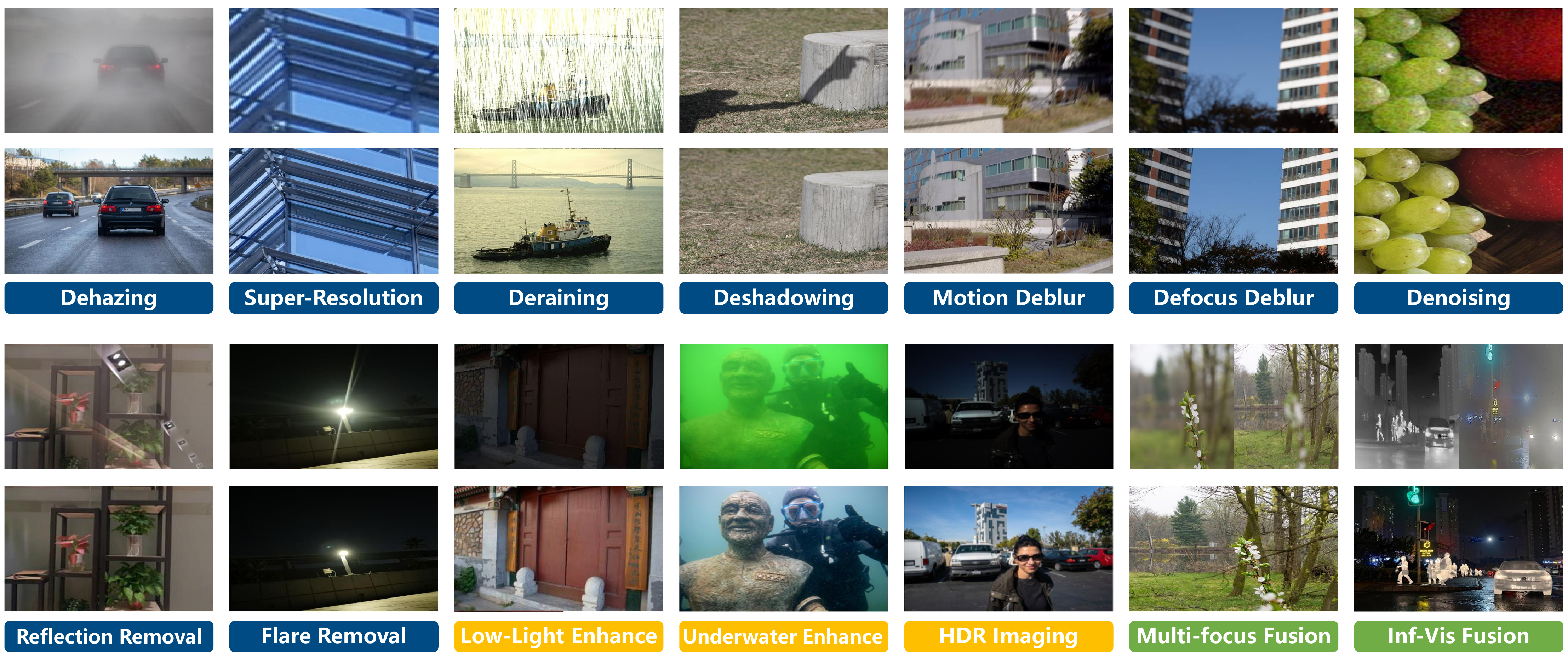}  
  \caption{Exemplary zero-shot results of Nano Banana Pro across 14 low-level vision tasks. For each task, the top row shows the degraded images, and the bottom row presents the corresponding restored outputs generated by Nano Banana Pro using simple text prompts. The visual results demonstrate the model's emerging capability for a diverse range of low-level vision tasks without task-specific training. The blue box represents image restoration tasks, the green box represents image enhancement tasks, and the yellow box represents image fusion tasks.} 
  \label{intro}  
\end{figure*}
The motivation for this study is rooted in a fundamental tension between human perception and traditional metrics. On one hand, the rich visual priors encapsulated within robust generative models should theoretically enable them to hallucinate plausible solutions for restoration, enhancement, and fusion tasks without task-specific training. On the other hand, this generative nature may conflict with the goal of achieving the strict, pixel-perfect fidelity that is valued by conventional evaluation metrics. To investigate this, we systematically evaluate the zero-shot capabilities of Nano Banana Pro using simple textual prompts, a stark contrast to the complex pipeline fine-tuning typically required for specialist models.

Our comprehensive empirical study spans 14 distinct low-level vision tasks across 40 datasets, covering image restoration, enhancement, and fusion. As visually exemplified in Fig.~\ref{intro}, Nano Banana Pro frequently produces outputs with remarkable perceptual quality. For instance, in tasks like dehazing or deraining, it can generate sharp edges and realistic textures that are often more aesthetically pleasing to a human observer than the results from specialist models. This initial observation immediately highlights a critical challenge: a model can be subjectively superior yet quantitatively inferior. Our work, therefore, aims not only to benchmark performance but also to delineate the boundaries of its current capabilities through rigorous quantitative and qualitative analysis. It is important to note that \textbf{the present evaluation reflects a conservative estimate of the model’s capability}, as we did not engage in meticulous prompt tuning or employ multi-round inference to cherry-pick optimal outputs. Our fixed, simple prompts represent a pragmatic but unoptimized use case.

Our findings uncover this anticipated dichotomy in stark detail: \textbf{Nano Banana Pro excels in perceptual quality but lags in metric-driven fidelity.} While it demonstrates remarkable zero-shot potential across a wide array of degradations, its outputs consistently score lower on reference-based metrics (e.g., PSNR, SSIM) when compared to domain-specific experts. We attribute this performance gap to the inherent stochasticity of generative models, which prioritize semantic plausibility over the strict pixel-wise alignment demanded by these metrics. In essence, while Nano Banana Pro has not yet achieved the status of a perfect all-rounder, it forces us to reconsider the traditional definition of success in low-level vision. It establishes a strong baseline for zero-shot restoration, highlighting both its emerging strengths and the critical need for new evaluation paradigms that can reconcile perceptual quality with pixel-level accuracy.

The remainder of this report is organized to systematically present these findings. We will detail our experimental results across Image Restoration, Image Enhancement, and Image Fusion, providing in-depth comparative analysis for each task. Finally, we conclude by summarizing the model's limitations and discussing potential future directions, including the development of more perception-aligned evaluation methods for generative low-level vision solvers.

\newpage

\tableofcontents  
\newpage

\textbf{\LARGE{Image Restoration}}
\section{Dehazing}
\subsection{Introduction}
Real-world Image Dehazing (RID) aims to recover clear and high-fidelity images from hazy observations captured in real-world environments. Unlike synthetic dehazing, where haze is generated under simplified physical assumptions, real-world haze is highly complex and exhibits strong spatial non-uniformity, large depth variations, severe color shifts, sensor noise, and compression artifacts. These factors make RID a long-standing and extremely challenging low-level vision problem. The goal of RID is not only to generate visually appealing results, but also to preserve accurate photometric and structural information to support reliable downstream vision tasks such as detection, tracking, and segmentation.

Early dehazing methods~\cite{song2023vision,qiu2023mb} relied on handcrafted statistical priors, such as the Dark Channel Prior (DCP)\cite{he2010single} and Non-Local Prior (NLP)\cite{berman2016non}, to constrain the solution space. While these physics-inspired approaches achieved initial success, they often fail to generalize across diverse real-world scenes and frequently introduce visible artifacts. With the rapid development of deep learning, numerous CNN-based and Transformer-based methods\cite{shao2020domain,chen2021psd,wu2023ridcp}, have significantly improved dehazing performance on synthetic benchmarks. However, collecting large-scale, perfectly aligned real-world hazy and clean image pairs remains nearly impossible. Although several real-world datasets have been constructed, their scale and diversity are still far from sufficient for training robust deep models. Consequently, most existing methods heavily rely on synthetic data and suffer from severe performance degradation when deployed in real-world scenarios.

To bridge this gap, recent studies have increasingly shifted their focus toward real-world dehazing. Some methods reintroduce physical priors to adapt pre-trained networks, while others modify inference strategies to improve generalization. Nevertheless, these approaches remain highly dependent on the quality of pre-training data. Moreover, heavily hazed images often contain severe information loss, fundamentally limiting the capability of traditional enhancement-based methods that lack generative flexibility to recover missing content.

Nano Banana is an image generation model developed by Google DeepMind. Its professional version, Nano Banana Pro, further enhances precision and world knowledge understanding. We applied it to real-world image dehazing tasks, with a focus on its effectiveness in removing haze, restoring blurred textures, and maintaining scene semantic integrity and tested it on mainstream dehazing benchmarks.

\subsection{Quantitative and Qualitative Results}
\begin{table}[ht]
\centering
\caption{Quantitative comparison of dehazing methods on multiple datasets. NB Pro achieves excellent NIMA scores on both datasets. It also demonstrates favorable FADE and BRISQUE scores on the RTTS dataset. However, its FADE and BRISQUE metrics on the Fattal’s dataset are unsatisfactory. The best results are in \textbf{black bold.}}
\label{tab:dehazing_results}
\resizebox{0.7\textwidth}{!}{%
\begin{tabular}{lcccccc}
\toprule
\multirow{2}{*}{\textbf{Method}} & \multicolumn{3}{c}{\textbf{RTTS}} & \multicolumn{3}{c}{\textbf{Fattal's}} \\
\cmidrule(lr){2-4} \cmidrule(lr){5-7} 
 & \textbf{FADE$\downarrow$} & \textbf{BRISQUE$\downarrow$} & \textbf{NIMA$\uparrow$} & \textbf{FADE$\downarrow$} & \textbf{BRISQUE$\downarrow$} & \textbf{NIMA$\uparrow$}  \\
\midrule
MBSDN~\cite{dong2020multi}  & 1.363  & 27.67 & 4.53 & 0.579 & \textbf{14.15} & 5.43  \\
Dehamer~\cite{guo2022image}   & 1.895 & 33.24 & 4.52 & 0.698 & 15.53 & 5.16  \\
DAD~\cite{shao2020domain} & 1.130 & 32.24 & 4.31 & 0.484 & 29.64 & \textbf{5.46}  \\
PSD~\cite{chen2021psd} & 0.920 & 27.71 & 4.60 & 0.416 & 23.61 & 4.99  \\
D4\cite{yang2022self}  & 1.358 & 33.21 & 4.48 & 0.411 & 20.33 & 5.44  \\
RIDCP\cite{wu2023ridcp}  & 0.944 & 17.29 & 4.97 & 0.408 & 20.05 & 5.43  \\
CORUN~\cite{fang2024real}  & \textbf{0.824} & \textbf{11.96} & \textbf{5.34} & \textbf{0.338} & 14.82 & 5.39  \\
\rowcolor{lightgray!50}
\textbf{\textcolor{orange}{NB Pro}} & 
\textbf{\textcolor{orange}{0.986}} & 
\textbf{\textcolor{orange}{27.21}} & 
\textbf{\textcolor{orange}{4.95}} & 
\textbf{\textcolor{orange}{0.683}} & 
\textbf{\textcolor{orange}{22.16}} & 
\textbf{\textcolor{orange}{5.44}}   \\
\bottomrule
\end{tabular}%
}
\end{table}
To intuitively present the dehazing outcomes of the Nano Banana (NB) Pro model, we provide quantitative and qualitative evaluations of its processing results across the RTTS~\cite{li2018benchmarking} and Fattal's~\cite{fattal2014dehazing} datasets, with comparisons to state-of-the-art baseline methods in real-world dehazing tasks. Tab.~\ref{tab:dehazing_results} show the performance comparison between NB Pro and current mainstream dehazing networks on the RTTS and Fattal's datasets, where we evaluated three real-world-oriented dehazing metrics: FADE , BRISQUE and NIMA. Integrating the evaluation results on both the RTTS and Fattal’s datasets, NB Pro demonstrates outstanding performance in terms of subjective visual quality, achieving top-tier NIMA scores on both benchmarks, which indicates that the generated images possess strong aesthetic appeal and favorable human perceptual quality. However, it performs poorly in terms of image naturalness. Specifically, NB Pro exhibits a significantly higher BRISQUE score on the Fattal’s dataset, suggesting that the outputs may suffer from over-enhancement artifacts. 

\begin{figure*}[ht]  
  \centering  
  \includegraphics[width=0.8\textwidth]{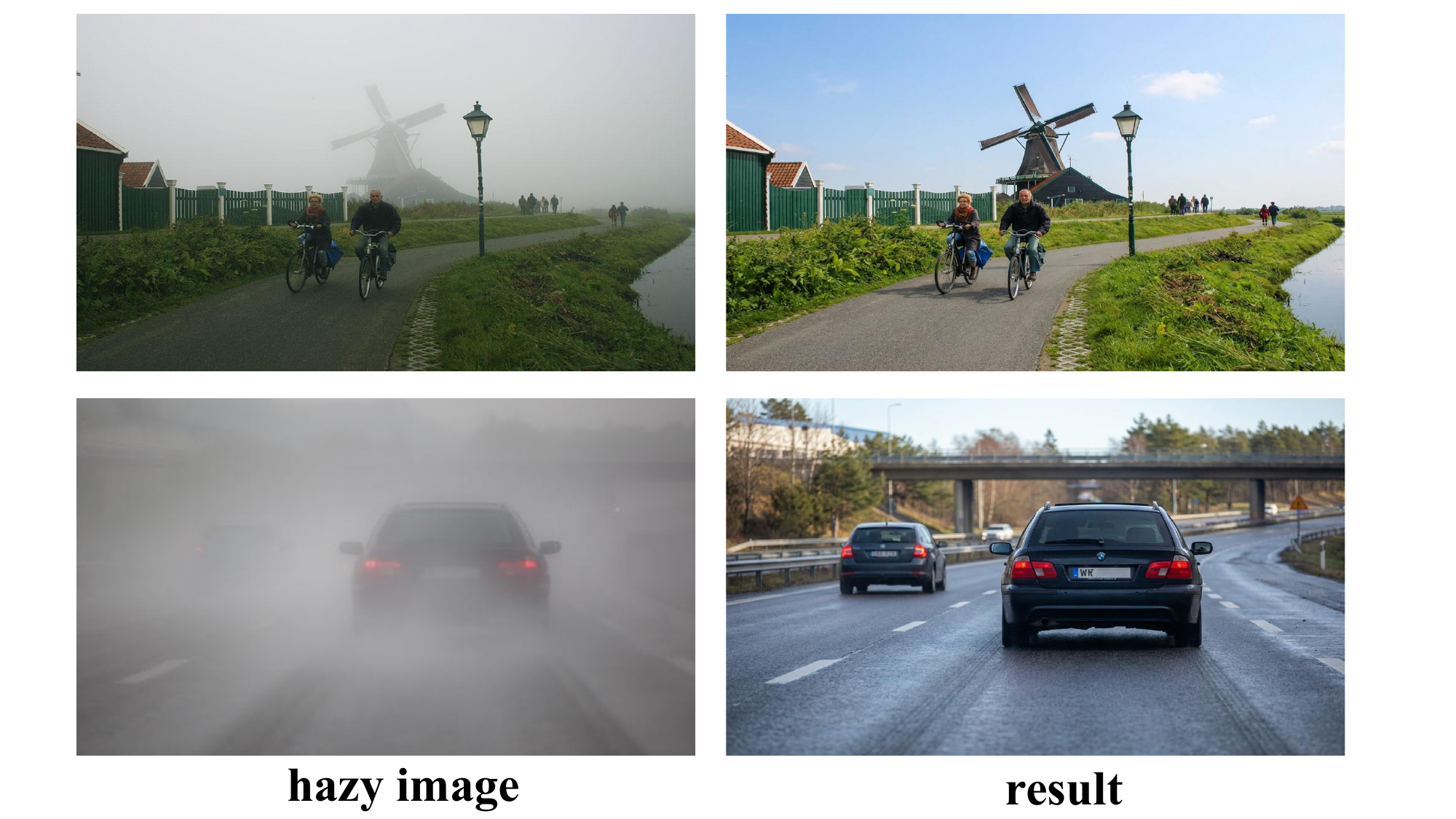}  
  \caption{NB Pro dehazing visual results on the RTTS dataset. Especially under heavy hazy conditions, NB Pro can effectively recover and enhance blurred details.} 
  \vspace{-5pt}
  \label{dehaz_figure}  
\end{figure*}
\begin{figure*}[ht]
  \centering  
  \includegraphics[width=1.0\textwidth]{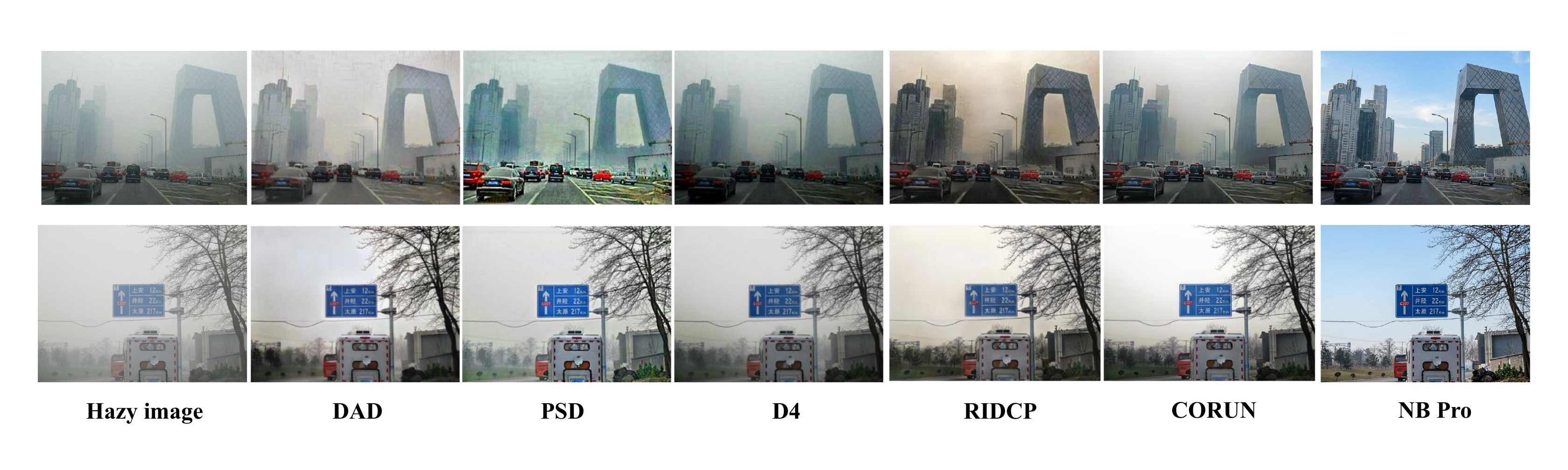}  
  \vspace{-8mm}
  \caption{A comparison of visual results between NB pro and other methods. It can be observed that the results produced by NB Pro are noticeably clearer and exhibit superior visual quality; however, they also suffer from obvious over-enhancement artifacts.} 
  \vspace{-5pt}
  \label{dehazcase_figure}  
\end{figure*}

Qualitative experimental results demonstrate that for extreme degradation scenarios such as dense fog with severe visibility loss, heavy atmospheric scattering, and complex urban or natural environments, NB Pro can generate perceptually enhanced results. Fig.~\ref{dehaz_figure} displays the visualization of exemplary dehazing cases for NB Pro on the RTTS dataset, highlighting both successful and challenging scenarios. For certain severely hazy images, benefiting from its powerful generative capabilities, NB Pro effectively restores intricate details—such as fine textures in buildings, vehicles, or vegetation—that are heavily obscured in the input, producing clear, high contrast outputs with impressive visual recovery of distant structures and overall scene coherence. Similarly, for some lightly foggy images, NB Pro performs well in optimization, selectively removing haze from distant backgrounds while preserving foreground sharpness and natural tones, resulting in balanced enhancements that improve visibility without introducing artifacts. Fig.~\ref{dehazcase_figure} shows the visual comparison of NB and other methods on the RTTS dataset. 

\begin{figure*}[ht]  
  \centering  
  \includegraphics[width=1.0\textwidth]{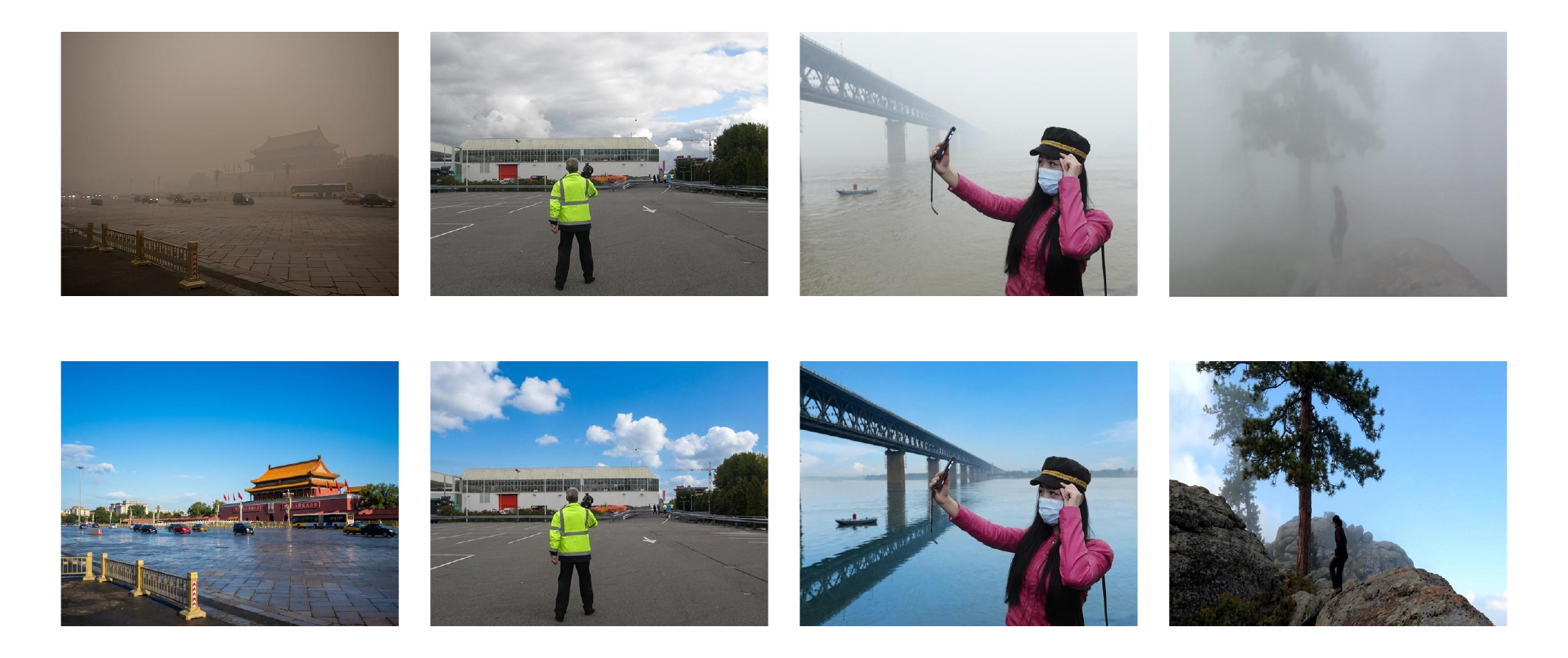}  
  \caption{Anamorphic example images of Nano Pro in dehazing on the RTTS dataset. The original hazy images is on top, and below is the results after hazy removal. The dehazed results exhibit poor color fidelity.} 
  \vspace{-5pt}
  \label{dehazfai_figure}  
\end{figure*}

However, NB Pro also exhibits numerous failure cases, as illustrated in Fig.~\ref{dehazfai_figure}. In these examples, spanning both moderate and heavy haze conditions, NB Pro often restores the images into distorted outputs characterized by unnatural color shifts, such as over-saturated or overly vibrant hues and hallucinatory weather elements, most notably forcing intensely blue skies into scenes that were originally overcast or neutral. These distortions undermine the authenticity of the atmospheric conditions, leading to results that deviate significantly from realistic dehazing expectations.

Overall, while NB Pro offers inspirational generative capabilities for ill-posed real-world dehazing, demonstrating the potential of semantic-driven priors in tackling highly ambiguous degradations—its limitations in color fidelity, physical realism, and consistent handling across varying haze densities suggest it is better suited for creative enhancements rather than precise low-level restoration. This prompts future research into hybrid approaches, such as combining NB Pro's zero-shot generative strengths with task-specific physical constraints or refined prompt engineering, to better constrain its tendencies toward perceptual appeal over faithful recovery.

\subsection{Analyses}
Stemming from a misalignment between training distributions and restoration objectives, NB Pro struggles to maintain spectral fidelity and atmospheric consistency. The model frequently over-corrects naturally muted, hazy tones into saturated, vibrant colors, introducing artificial chromatic biases that alter the scene's intrinsic mood. In severely hazy scenarios where high-frequency details are obliterated, NB Pro’s generative priors dominate the restoration process. Rather than solving the inverse physical scattering model, it hallucinates details based on learned statistical patterns. A defining characteristic of this behavior is the systematic rendering of vivid blue skies in originally overcast scenes. While visually striking and aligned with subjective preferences for ``clear weather'', this deviation undermines the scene's authenticity and temporal consistency.

Consequently, NB Pro is currently best positioned for creative content generation—prioritizing perceptual appeal and "wow-factor"—rather than forensic restoration tasks demanding strict adherence to physical constraints. However, its value remains significant. In the ill-posed domain of real-world dehazing, where traditional physics-based methods often falter due to unknown degradation parameters, NB Pro demonstrates the power of semantic priors to reconstruct plausible details in information-deficient scenarios. This suggests a pivotal direction for future research: developing hybrid frameworks. By integrating NB Pro-like generative backbones with task-specific physical constraints (e.g., atmospheric scattering laws) and fidelity anchors, we can bridge the gap between creative hallucination and realistic restoration, aiming for a paradigm that balances visual delight with physical truth.

\newpage

\section{Super-Resolution}

\subsection{Introduction}
Real-World Image Super-Resolution (Real-ISR) aims to restore high-fidelity, high-resolution content from low-resolution inputs degraded by complex, unknown physical processes. Unlike synthetic super-resolution, where degradations are modeled by simple bicubic downsampling, real-world scenarios involve intricate combinations of blur, sensor noise, compression artifacts, and varying camera response functions~\cite{cai2019toward, wei2020component}. This complexity renders traditional regression-based methods—which rely on pixel-wise optimization (e.g., MSE loss)—ineffective, often resulting in over-smoothed textures and a lack of high-frequency details~\cite{dong2014learning, zhang2018image}.

The landscape of generative Real-ISR has evolved rapidly. Generative Adversarial Networks (GANs)~\cite{goodfellow2014generative, ledig2017photo}, represented by methods such as BSRGAN~\cite{zhang2021designing} and Real-ESRGAN~\cite{wang2021real}, introduced high-order degradation modeling to synthesize realistic training pairs, significantly improving visual perceptual quality over PSNR-oriented baselines. More recently, Denoising Diffusion Probabilistic Models (DDPMs)~\cite{ho2020denoising} have emerged as the new state-of-the-art. Methods like StableSR~\cite{wang2023exploiting} and DiffBIR~\cite{lin2023diffbir} leverage strong priors from large-scale pre-trained text-to-image models (e.g., Stable Diffusion~\cite{rombach2022high}) to generate intricate textures. However, these multi-step diffusion models often suffer from high computational costs and slow inference speeds. To mitigate this, acceleration techniques have been proposed, exemplified by SinSR~\cite{wang2024sinsr}, which distills complex diffusion priors into single-step inference models. Despite these advancements, a critical challenge remains: the inherent trade-off between perceptual quality and signal fidelity~\cite{blau2018perception}, often leading to artifacts or structural hallucinations in the pursuit of sharpness.

In this work, we conduct a comprehensive quantitative and qualitative evaluation of Nano Banana Pro, a novel generative ISR framework, benchmarking it against a spectrum of industry-standard algorithms, including GAN-based approaches (BSRGAN~\cite{zhang2021designing}, Real-ESRGAN~\cite{wang2021real}), multi-step diffusion models (StableSR~\cite{wang2023exploiting}, DiffBIR~\cite{lin2023diffbir}), and accelerated diffusion methods (SinSR~\cite{wang2024sinsr}). Our goal is to rigorously assess where Nano Banana Pro stands within the current Perception-Distortion landscape.

To ensure a robust evaluation across varying degradation complexities, our experiments are conducted on the large-scale DIV2K-Val dataset (2,994 images)~\cite{agustsson2017ntire} as well as the authentic RealSR~\cite{cai2019toward} and DRealSR~\cite{wei2020component} benchmarks. Recognizing that pixel fidelity alone is insufficient for characterizing generative performance, we employ a comprehensive set of evaluation metrics, incorporating not only standard Full-Reference indicators (PSNR, SSIM~\cite{wang2004image}, LPIPS~\cite{zhang2018unreasonable}) but also widely-adopted No-Reference perceptual metrics (NIQE~\cite{mittal2012making}, MUSIQ~\cite{ke2021musiq}, CLIPIQA~\cite{wang2023exploring}). Under this rigorous testing framework, we systematically assess the reconstruction capabilities of Nano Banana Pro in comparison to established GAN-based and diffusion-based baselines. The resulting analysis provides a detailed characterization of the model's behavior regarding the perception-distortion trade-off, offering valuable insights into its suitability for real-world applications.

\subsection{Quantitative Results}

\begin{table*}[t]
\centering
\caption{Quantitative comparison on synthetic (DIV2K-Val\cite{agustsson2017ntire}) and real-world (RealSR\cite{cai2019toward}, DRealSR\cite{wei2020component}) benchmarks. The best and second best results are highlighted by \textbf{black bold} and \underline{underline}.}
\normalsize
\setlength{\heavyrulewidth}{1.2pt}
\setlength{\lightrulewidth}{1pt}
\renewcommand{\arraystretch}{1.2}
\resizebox{0.8\textwidth}{!}{

\begin{tabular}{l l ccc ccc}
\toprule

\multirow{2}{*}{Test Dataset} & \multirow{2}{*}{Method} & \multicolumn{3}{c}{Full-Reference Metrics} & \multicolumn{3}{c}{No-Reference Metrics} \\
\cmidrule(lr){3-5} \cmidrule(lr){6-8}
& & PSNR$\uparrow$ & SSIM$\uparrow$ & LPIPS$\downarrow$ & NIQE$\downarrow$ & MUSIQ$\uparrow$ & CLIPIQA$\uparrow$ \\
\midrule

\multirow{6}{*}{DIV2K-Val}
& BSRGAN~\cite{zhang2021designing} & \textbf{24.58} & \underline{0.6269} & 0.3351 & 4.75 & 61.20 & 0.5071 \\
& Real-ESRGAN~\cite{wang2021real} & 24.29 & \textbf{0.6371} & \textbf{0.3112} & \underline{4.68} & 61.06 & 0.5501 \\
& StableSR~\cite{wang2023exploiting} & 23.26 & 0.5726 & \underline{0.3113} & 4.76 & \textbf{65.92} & \underline{0.6192} \\
& DiffBIR~\cite{lin2023diffbir} & 23.64 & 0.5647 & 0.3524 & 4.70 & \underline{65.81} & \textbf{0.6210} \\
& SinSR~\cite{wang2024sinsr} & \underline{24.41} & 0.6018 & 0.3240 & 6.02 & 62.82 & 0.5386 \\
& \textbf{\textcolor{orange}{Nano Banana Pro}} & \textbf{\textcolor{orange}{20.29}} & \textbf{\textcolor{orange}{0.4720}} & \textbf{\textcolor{orange}{0.3645}} & \textbf{3.52} & \textbf{\textcolor{orange}{65.40}} & \textbf{\textcolor{orange}{0.5257}} \\
\midrule
\multirow{6}{*}{RealSR}
& BSRGAN & \textbf{26.39} & \textbf{0.7654} & \textbf{0.2670} & 5.66 & 63.21 & 0.5001 \\
& Real-ESRGAN & 25.69 & \underline{0.7616} & \underline{0.2727} & 5.83 & 60.18 & 0.4449 \\
& StableSR & 24.70 & 0.7085 & 0.3018 & 5.91 & \textbf{65.78} & \underline{0.6221} \\
& DiffBIR & 24.75 & 0.6567 & 0.3636 & \underline{5.53} & \underline{64.98} & \textbf{0.6246} \\
& SinSR & \underline{26.28} & 0.7347 & 0.3188 & 6.29 & 60.80 & 0.5385 \\
& \textbf{\textcolor{orange}{Nano Banana Pro}} & \textbf{\textcolor{orange}{23.56}} & \textbf{\textcolor{orange}{0.6649}} & \textbf{\textcolor{orange}{0.2978}} & \textbf{4.39} & \textbf{\textcolor{orange}{60.18}} & \textbf{\textcolor{orange}{0.5199}} \\
\midrule
\multirow{6}{*}{DRealSR}
& BSRGAN & \textbf{28.75} & \underline{0.8031} & \underline{0.2883} & 6.52 & 57.14 & 0.4915 \\
& Real-ESRGAN & \underline{28.64} & \textbf{0.8053} & \textbf{0.2847} & 6.69 & 54.18 & 0.4422 \\
& StableSR & 28.03 & 0.7536 & 0.3284 & 6.52 & 58.51 & \underline{0.5601} \\
& DiffBIR & 26.71 & 0.6571 & 0.4557 & \underline{6.31} & \textbf{61.07} & \textbf{0.5930} \\
& SinSR & 28.36 & 0.7515 & 0.3665 & 6.99 & 55.33 & 0.4884 \\
& \textbf{\textcolor{orange}{Nano Banana Pro}} & \textbf{\textcolor{orange}{23.97}} & \textbf{\textcolor{orange}{0.6323}} & \textbf{\textcolor{orange}{0.3809}} & \textbf{5.03} & \underline{\textbf{\textcolor{orange}{59.00}}} & \textbf{\textcolor{orange}{0.5145}} \\
\bottomrule
\end{tabular}
}
\label{tab:quantitative_comparison}
\end{table*}

To comprehensively evaluate Nano Banana Pro's performance in Real-ISR tasks, we quantitatively compared it against a range of advanced GAN-based and diffusion-based image super-resolution methods. We employed standard full-reference metrics: PSNR and SSIM to evaluate signal fidelity, and LPIPS to assess perceptual similarity. Additionally, No-Reference (NR) metrics NIQE, MUSIQ, and CLIPIQA were utilized to quantify the statistical naturalness and aesthetic quality of the generated images. Results are shown in Tab.~\ref{tab:quantitative_comparison}. Nano Banana Pro significantly underperformed against the comparison methods in terms of traditional fidelity metrics. On the DIV2K-Val dataset, Nano Banana Pro achieved significantly lower PSNR and SSIM than the optimal method, lagging by over 4 dB. A similar trend, though less severe, was observed on the RealSR and DRealSR datasets, where its fidelity scores remained consistently behind both GAN-based and diffusion-based baselines. This result clearly indicates that under the standard full-reference evaluation framework, which prioritizes pixel-level accurate reconstruction, Nano Banana Pro's generated results exhibit systematic deviations from the ground-truth reference images.

Nano Banana Pro fundamentally differs from traditional super-resolution models optimized for specific degradation kernels. The latter typically undergo end-to-end training targeting minimization of pixel-level loss—thus inherently excelling in metrics like PSNR and SSIM. In contrast, Nano Banana Pro's generation process prioritizes semantic coherence and visual cleanliness. Its outputs can be viewed as plausible reconstructions of the input image rather than strict pixel-to-pixel mappings. Consequently, generated images may exhibit deviations in local texture alignment and structural positioning compared to reference images, leading to comprehensive score reductions across full-reference metrics. However, notably, on the NIQE metric, Nano Banana Pro consistently achieved the best scores across all three datasets (e.g., 3.52 on DIV2K-Val vs. 4.75 for BSRGAN). This suggests its outputs possess superior statistical naturalness, effectively suppressing artifacts, even though the low-level pixel arrangements have been altered from the original reference.

\begin{figure}[H]  
  \centering  
  \includegraphics[width=0.8\textwidth]{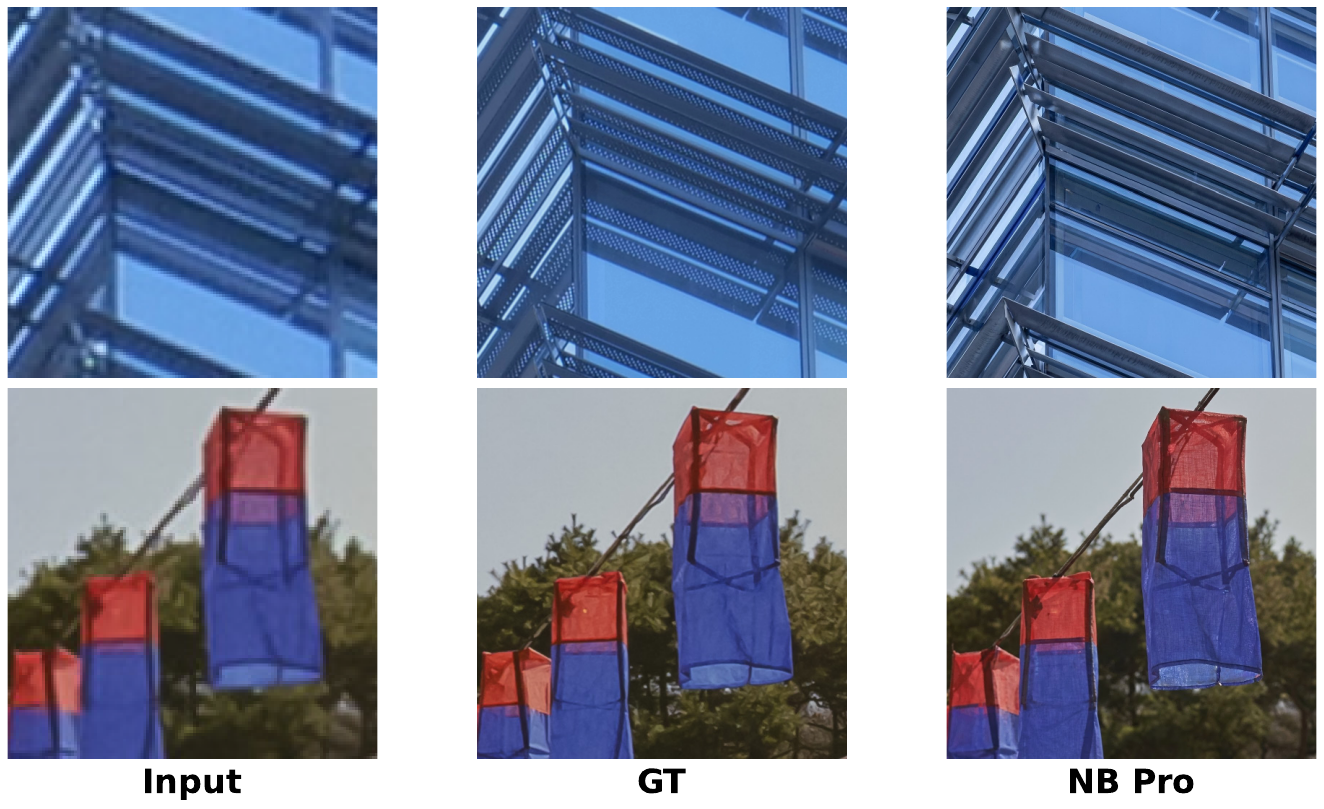}  
  \caption{Qualitative visualization of structural recovery in Real-ISR tasks using Nano Banana Pro.} 
  \vspace{-5pt}
  \label{SR1}  
\end{figure}

\begin{figure}[ht]  
  \centering  
  \includegraphics[width=0.8\textwidth]{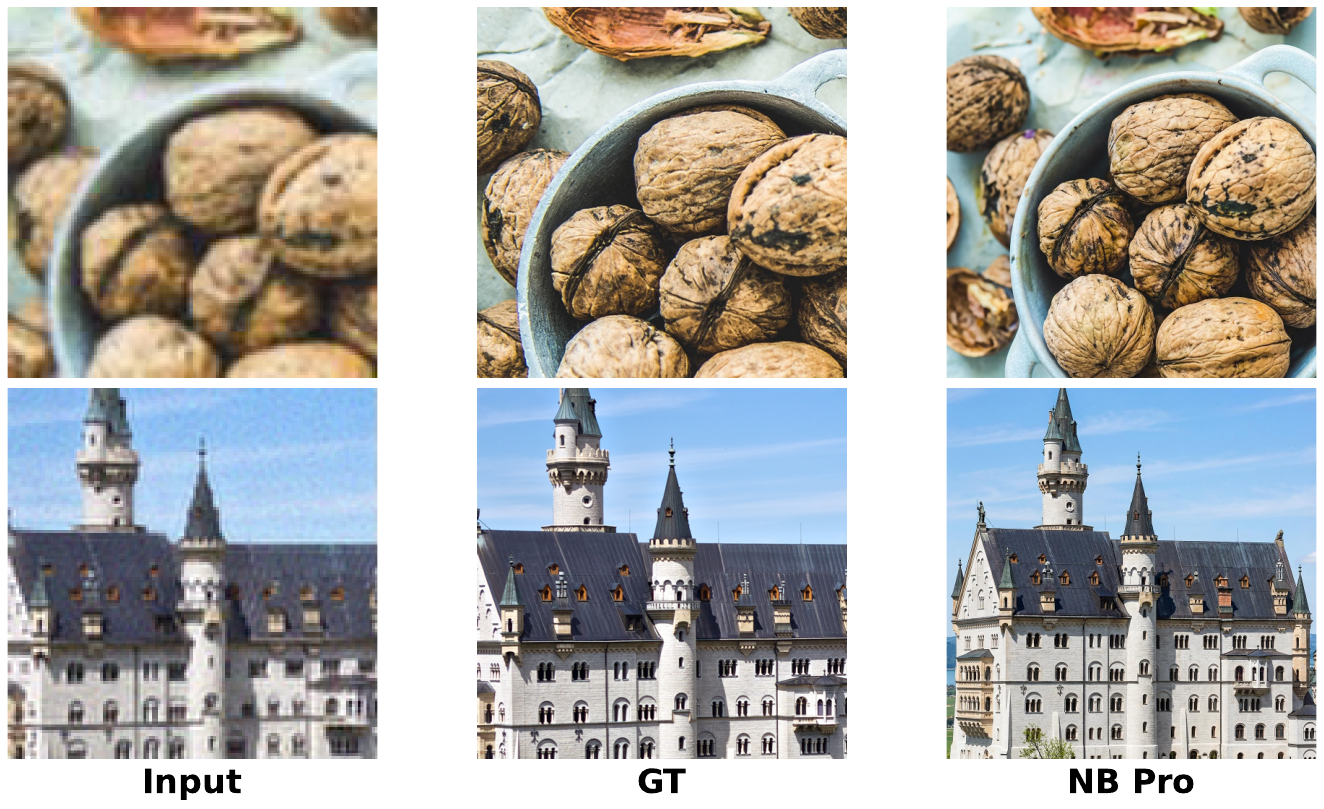}  
  \caption{ Visualization of unintended image boundary extension cases in Real-ISR tasks using Nano Banana Pro.} 
  \vspace{-5pt}
  \label{SR2}  
\end{figure}

\begin{figure}[H]  
  \centering  
  \includegraphics[width=0.8\textwidth]{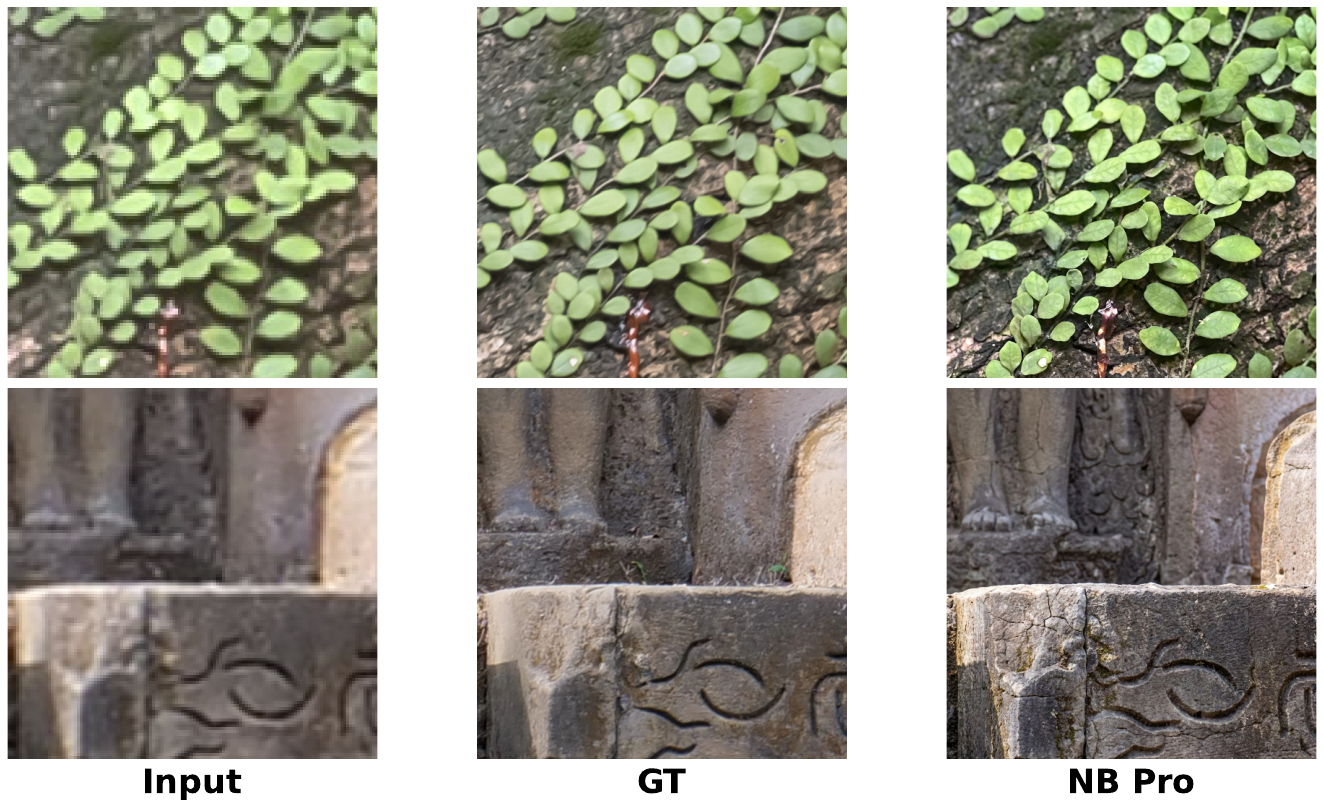}  
  \caption{Visualization of generative texture deviations in Real-ISR tasks using Nano Banana Pro.} 
  \vspace{-5pt}
  \label{SR3}  
\end{figure}

\begin{figure*}[ht]  
  \centering  
  \includegraphics[width=0.8\textwidth]{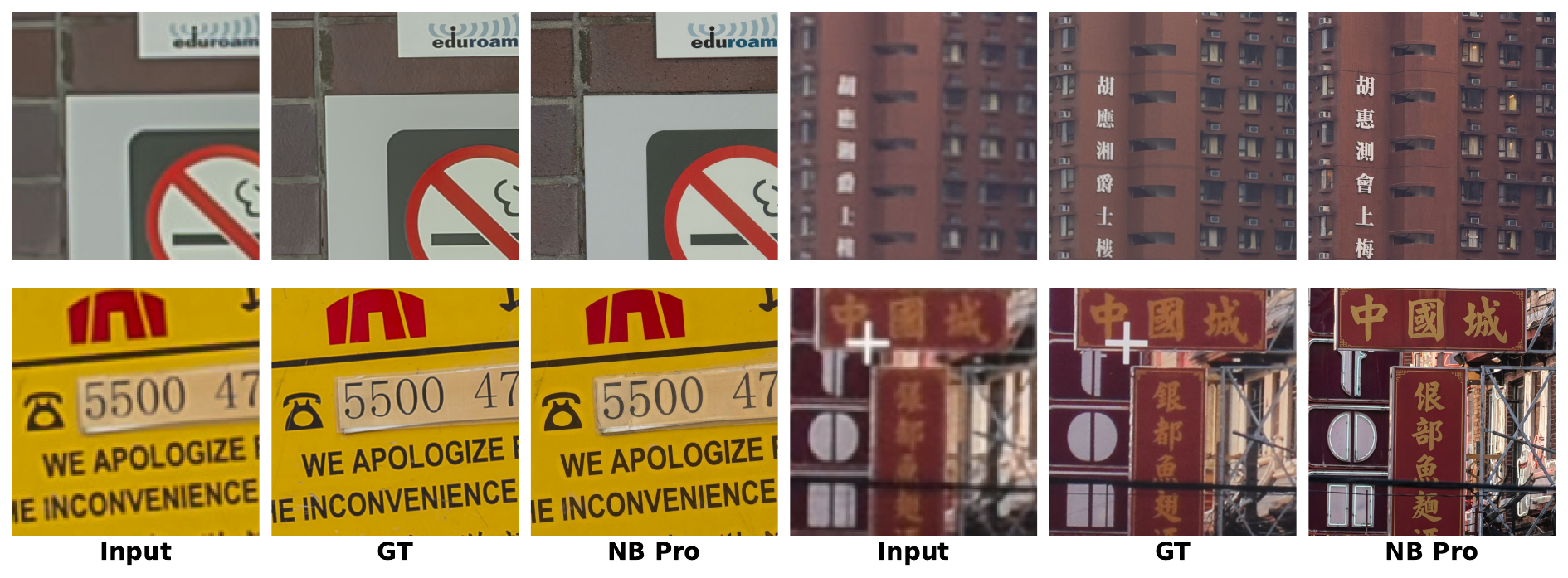}  
  \caption{This visualization illustrates text reconstruction, showing successful character recovery in the left panel and erroneous character hallucination in the right panel.} 
  \vspace{-5pt}
  \label{SR4}  
\end{figure*}

\subsection{Qualitative Results}
In this section, we examine the generative characteristics of Nano Banana Pro across the DIV2K, RealSR, and DRealSR datasets. The qualitative evaluation is organized into four key scenarios to highlight both the strengths and failure modes of the model: geometric clarity, field-of-view artifacts, texture fidelity, and character reconstruction (Figs.~\ref{SR1}--\ref{SR4}).

Fig.~\ref{SR1} displays the super-resolution results on scenes with distinct geometric structures. In the examples of the architectural facade and hanging lanterns, Nano Banana Pro effectively sharpens the blurred edges and recovers the linear patterns lost in the low-resolution inputs. The resulting images maintain structural coherence and exhibit reduced noise, offering a noticeable improvement in clarity compared to the inputs.

Fig.~\ref{SR2} illustrates a distinct structural anomaly observed in Nano Banana Pro: unintended Field-of-View (FOV) expansion. Comparing the low-resolution input, the Ground Truth (GT), and the Nano Banana Pro generated result reveals that the model fails to strictly adhere to the original spatial boundaries of the input image. Instead of merely super-resolving the existing pixels, Nano Banana Pro erroneously hallucinates additional content along the image periphery. 

Fig.~\ref{SR3} illustrates the tendency of Nano Banana Pro to synthesize plausible but non-existent details in areas with complex stochastic textures. In the foliage and stone carving examples, while the model produces sharp, high-frequency patterns, these generated textures often deviate from the Ground Truth. Specifically, the arrangement of leaf veins and the granular surface of the stone are reconstructed with altered local structures rather than being faithfully restored, leading to pixel-level discrepancies that lower fidelity scores.

Fig.~\ref{SR4} highlights the dependency of Nano Banana Pro on semantic recognizability for text reconstruction. In the examples on the left, where the low-resolution input retains discernible structural features (such as the logo and digits), the model accurately reconstructs sharp and legible characters. Conversely, the examples on the right demonstrate failure cases where severe degradation obscures the original glyphs, particularly with complex Chinese characters. In these instances, the model fails to recover the correct semantic content and instead hallucinates incorrect strokes or non-existent characters, resulting in high-contrast but semantically erroneous text.

\subsection{Analyses}
Our comprehensive evaluation elucidates the operational characteristics of Nano Banana Pro within the Real-ISR domain. Quantitatively, the model trails significantly behind mainstream GAN and diffusion-based methods in fidelity metrics (PSNR/SSIM) across the DIV2K-Val and RealSR datasets; however, it achieves remarkable performance in the no-reference NIQE metric. This discrepancy suggests that, as evidenced in Fig.~\ref{SR3}, the model prioritizes learned generative priors to synthesize texture details rather than strictly adhering to the low-resolution input.

Qualitatively, unlike traditional restoration models that maintain strict spatial consistency, Nano Banana Pro lacks precise pixel-level alignment with the reference image. Consequently, unintended Field-of-View (FOV) expansion is observed, as shown in Fig.~\ref{SR2}. Furthermore, the text reconstruction failures in Fig.~\ref{SR4} reveal the model's heavy reliance on feature recognizability. When encountering degraded features such as blurred text, the model exhibits a tendency to aggressively generate sharp outputs. This behavior causes feature recognition errors to have a catastrophic impact on the result, leading to the hallucination of sharp but semantically incorrect text.

Synthesizing these findings, Nano Banana Pro is highly suitable for perception-centric applications—such as artistic upscaling, old photo restoration, or casual photography—where visual purity and noise elimination are paramount. However, due to its propensities for texture hallucination, spatial misalignment, and semantic alteration, it is unsuitable for fidelity-critical scenarios.
\newpage
\section{Deraining}

\subsection{Background}
Rain is a common yet challenging weather degradation that severely obscures scene content and alters the structural appearance of images. Such degradations significantly reduce visual quality and adversely affect the reliability of numerous outdoor vision systems, including intelligent transportation, UAV-based monitoring, and autonomous navigation. Single image deraining, which seeks to restore a clean background from a rain-contaminated observation, has therefore become an essential task in low-level vision. In recent years, a variety of deraining algorithms and benchmark datasets have been developed, achieving remarkable progress in modeling rain streaks, raindrops, and atmospheric veils.

Recent advances in large-scale multimodal generative models offer a promising new direction for image restoration. Among these models, Nano Banana Pro, Google’s latest high-fidelity visual generation system built upon the powerful Gemini 3 Pro multimodal reasoning engine, excels in semantic comprehension, fine-grained visual modeling, and precise structural control. Designed for professional image creation and editing, Nano Banana Pro supports high-resolution synthesis, multi-image fusion, accurate text rendering, and semantically coherent scene manipulation. These capabilities suggest that the model is not only adept at generating novel visual content, but also inherently possesses strong prior knowledge for reconstructing clean structures and textures, rendering it a promising candidate for image restoration tasks such as deraining.

In this study, we investigate the feasibility of adapting Nano Banana Pro to single-image deraining. In contrast to traditional restoration networks that rely on explicit task-specific modeling, Nano Banana Pro utilizes its rich world model and multimodal reasoning ability to interpret degraded regions, infer plausible background structures, and produce natural, artifact-free reconstructions. By framing deraining as a guided generative reconstruction problem, we aim to harness the model’s semantic priors and sophisticated visual synthesis capabilities to achieve rain removal across diverse scenarios.

\subsection{Experiment Setup}
To thoroughly evaluate the performance and generalization ability of Nano Banana Pro on the single image deraining task, we conduct experiments on three widely used benchmark datasets: two synthetic datasets, Rain200L and Rain200H\cite{yang2017deep}, and one real-world dataset, SPA\cite{Wang_2019_CVPR}. These datasets cover a broad range of rain patterns and scene complexities, enabling a comprehensive assessment of the model’s restoration capability.
\begin{itemize}
    \item \textbf{Rain200L}\cite{yang2017deep}: Contains 1,800 pairs of synthetic training images and 200 test pairs. The rain streaks in this dataset exhibit a single predominant direction and relatively low density, making it suitable for assessing the model’s basic restoration ability under simple rain conditions.
    \item \textbf{Rain200H}\cite{yang2017deep}: Provides 1,800 training pairs and 200 test pairs, but features rain streaks with higher density and more diverse orientations. This dataset is designed to evaluate the robustness of deraining models when faced with heavy and structurally complex rain degradations.
    \item \textbf{SPA}\cite{Wang_2019_CVPR}: A large-scale real-world rainy image dataset comprising 638,492 training pairs and 1,000 test images. The rain patterns in SPA are highly diverse, and the background scenes vary significantly, making it an appropriate benchmark for measuring cross-domain generalization from synthetic to real rainy conditions.
\end{itemize}

All images from all datasets are given the prompt: \textit{``This is a rainy image. Please remove the rain streaks and raindrops while keeping all other elements, the original color tone, lighting, and atmosphere unchanged.''}

It is worth emphasizing that Nano Banana Pro is evaluated in a strictly zero-shot manner: no training images are used for optimization, fine-tuning, or adaptation, and the model is directly applied to the test images via a fixed textual prompt.

\subsection{Quantitative and Qualitative Results}
For fair comparison, all metrics are computed under exactly the same evaluation protocol as NeRD-Rain, including image resolution, color space, and PSNR/SSIM computation.
Based on the quantitative results reported in Tab.~\ref{tab:results_deraining}, we systematically evaluate the image deraining performance of Nano Banana Pro on two synthetic datasets, Rain200L\cite{yang2017deep} and Rain200H\cite{yang2017deep}, as well as the real-world SPA-Data dataset\cite{Wang_2019_CVPR}, and compare it with a wide range of representative methods. The compared approaches cover prior-based methods (DSC\cite{luo2015removing}, GMM\cite{li2016rain}), CNN-based methods (DDN\cite{fu2017removing}, RESCAN\cite{li2018recurrent}, PReNet\cite{ren2019progressive}, MSPFN\cite{jiang2020multi}, RCDNet\cite{wang2020model}, MPRNet\cite{zamir2021multi}, DualGCN\cite{fu2021rain}, SPDNet\cite{yi2021structure}), and Transformer-based methods (Uformer\cite{wang2022uformer}, Restormer\cite{zamir2022restormer}, IDT\cite{xiao2022image}, DRSformer\cite{chen2023learning}, NeRD-Rain\cite{NeRD-Rain}).

\begin{table}[h]
\centering
\caption{Quantitative comparison results on three representative benchmarks. The best results are in \textbf{black bold.}}
\label{tab:results_deraining}
\resizebox{0.6\textwidth}{!}{
\begin{tabular}{lcccccc}
\toprule
\multirow{2}{*}{Method} & \multicolumn{2}{c}{Rain200L} & \multicolumn{2}{c}{Rain200H} & \multicolumn{2}{c}{SPA-Data} \\ \cmidrule(lr){2-3} \cmidrule(lr){4-5} \cmidrule(lr){6-7}
 & PSNR $\uparrow$ & SSIM $\uparrow$ & PSNR $\uparrow$ & SSIM $\uparrow$ & PSNR $\uparrow$ & SSIM $\uparrow$ \\ \midrule
DSC\cite{luo2015removing} & 27.16 & 0.8663 & 14.73 & 0.3815 & 34.95 & 0.9416 \\
GMM\cite{li2016rain} & 28.66 & 0.8652 & 14.50 & 0.4164 & 34.30 & 0.9428 \\
DDN\cite{fu2017removing} & 34.68 & 0.9671 & 26.05 & 0.8056 & 36.16 & 0.9457 \\
RESCAN\cite{li2018recurrent} & 36.09 & 0.9697 & 26.75 & 0.8353 & 38.11 & 0.9707 \\
PReNet\cite{ren2019progressive} & 37.80 & 0.9814 & 29.04 & 0.8991 & 40.16 & 0.9816 \\
MSPFN\cite{jiang2020multi} & 38.58 & 0.9827 & 29.36 & 0.9034 & 43.43 & 0.9843 \\
RCDNet\cite{wang2020model} & 39.17 & 0.9885 & 30.24 & 0.9048 & 43.36 & 0.9831 \\
MPRNet\cite{zamir2021multi} & 39.47 & 0.9825 & 30.67 & 0.9110 & 43.64 & 0.9844 \\
DualGCN\cite{fu2021rain} & 40.73 & 0.9886 & 31.15 & 0.9125 & 44.18 & 0.9902 \\
SPDNet\cite{yi2021structure} & 40.50 & 0.9875 & 31.28 & 0.9207 & 43.20 & 0.9871 \\
Uformer\cite{wang2022uformer} & 40.20 & 0.9860 & 30.80 & 0.9105 & 46.13 & 0.9913 \\
Restormer\cite{zamir2022restormer} & 40.99 & 0.9890 & 32.00 & 0.9329 & 47.98 & 0.9921 \\
IDT\cite{xiao2022image} & 40.74 & 0.9884 & 32.10 & 0.9344 & 47.35 & 0.9930 \\
DRSformer\cite{chen2023learning} & 41.23 & 0.9894 & 32.17 & 0.9326 & 48.54 & 0.9924 \\
NeRD-Rain\cite{NeRD-Rain} & \textbf{41.71} & \textbf{0.9903} & \textbf{32.40} & \textbf{0.9373} & \textbf{49.58} & \textbf{0.9940} \\ 
\rowcolor{lightgray!50}
\textbf{\textcolor{orange}{Nano Banana Pro}} & \textbf{\textcolor{orange}{26.05}} & \textbf{\textcolor{orange}{0.7954}} & \textbf{\textcolor{orange}{21.10}} & \textbf{\textcolor{orange}{0.6659}} & \textbf{\textcolor{orange}{32.25}} & \textbf{\textcolor{orange}{0.9142}} \\ \bottomrule
\end{tabular}}
\end{table}

The quantitative performance of Nano Banana Pro is significantly inferior to that of state-of-the-art deraining models across all three datasets. On Rain200L\cite{yang2017deep}, Nano Banana Pro achieves 26.05 dB PSNR and 0.7954 SSIM, which are substantially lower than those of supervised learning-based methods. On the more challenging Rain200H dataset with complex rain patterns, Nano Banana Pro obtains 21.10 dB PSNR and 0.6659 SSIM. Although it outperforms traditional prior-based methods, a considerable performance gap remains compared to CNN-based and Transformer-based approaches. On the real-world SPA-Data dataset\cite{Wang_2019_CVPR}, Nano Banana Pro reaches 32.25 dB PSNR and 0.9142 SSIM, still noticeably below the current best-performing methods.

\begin{figure}[htbp]
    \centering
    \includegraphics[width=1\textwidth]{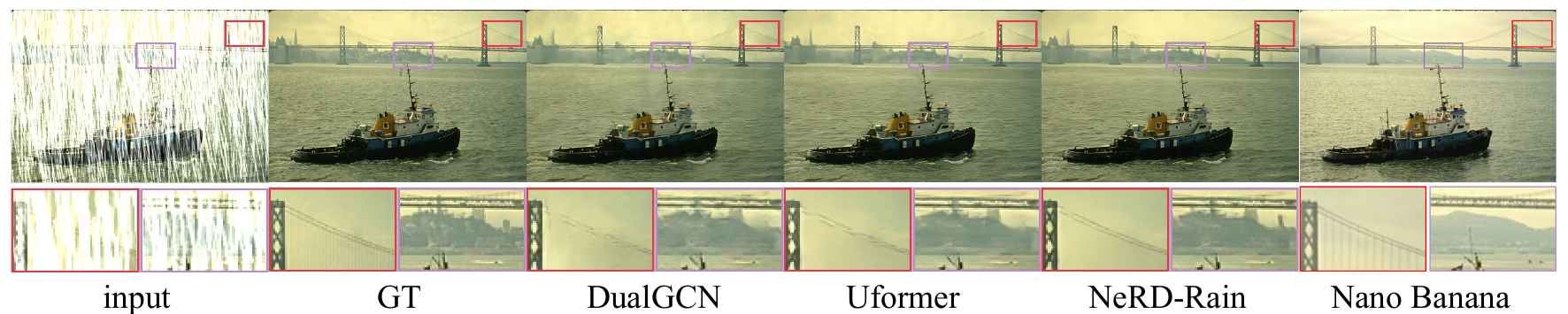}
    \centering
    \caption{Qualitative comparison on the Rain200H\cite{yang2017deep} dataset. Close-up views better illustrate the deraining capability.}
    \label{fig:deraining_fig1}
\end{figure}

\begin{figure}[htbp]
    \centering
    \includegraphics[width=0.7\textwidth]{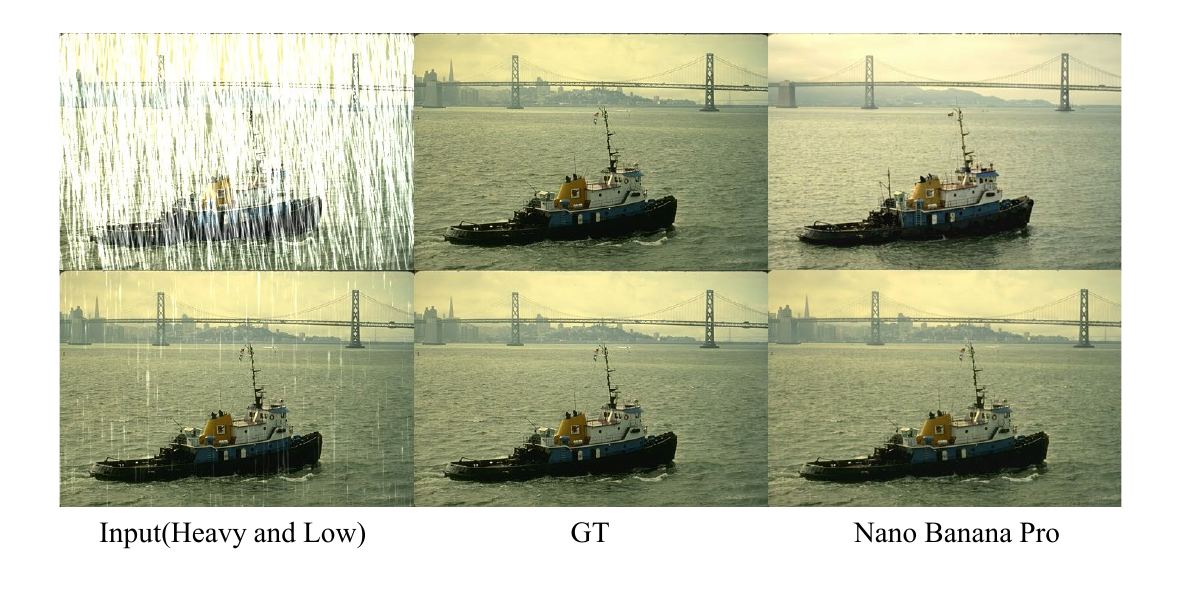}
    \centering
    \vspace{-5mm}
    \caption{Qualitative comparison under different rain intensities. Under lighter rain, Nano Banana Pro better preserves the original tone and fine details, while heavier rain leads to noticeable color shifts and detail loss.}
    \label{fig:deraining_fig2}
    \vspace{-3mm}
\end{figure}
This quantitative degradation is consistent with the visual results in Fig.~\ref{fig:deraining_fig1}, where large and dense synthetic rain streaks severely obscure background content, leading to color deviations and missing or oversmoothed fine details. Since Nano Banana Pro is not trained specifically for image deraining, local structures are often reconstructed via implicit generative hallucination rather than pixel-wise restoration, which negatively affects PSNR and SSIM. Nevertheless, the model demonstrates notable strength in recovering certain global structures; for example, the cable geometry of the suspension bridge is reconstructed with higher structural coherence and semantic plausibility than several supervised baselines. 

Furthermore, Fig.~\ref{fig:deraining_fig2} shows that the deraining performance of Nano Banana Pro is highly sensitive to rainfall intensity: under low-rain conditions, where more reliable visual information is preserved, both color fidelity and fine details are significantly improved, whereas heavy rainfall introduces severe occlusion and ambiguity, resulting in pronounced color shifts and detail degradation. Overall, these results indicate that while generative multimodal models are disadvantaged in pixel-level fidelity under zero-shot deraining, they retain strong semantic and structural priors, suggesting complementary potential in scenarios with limited supervision or severe information loss.

\begin{figure}[htbp]
    \centering
    \includegraphics[width=0.6\textwidth]{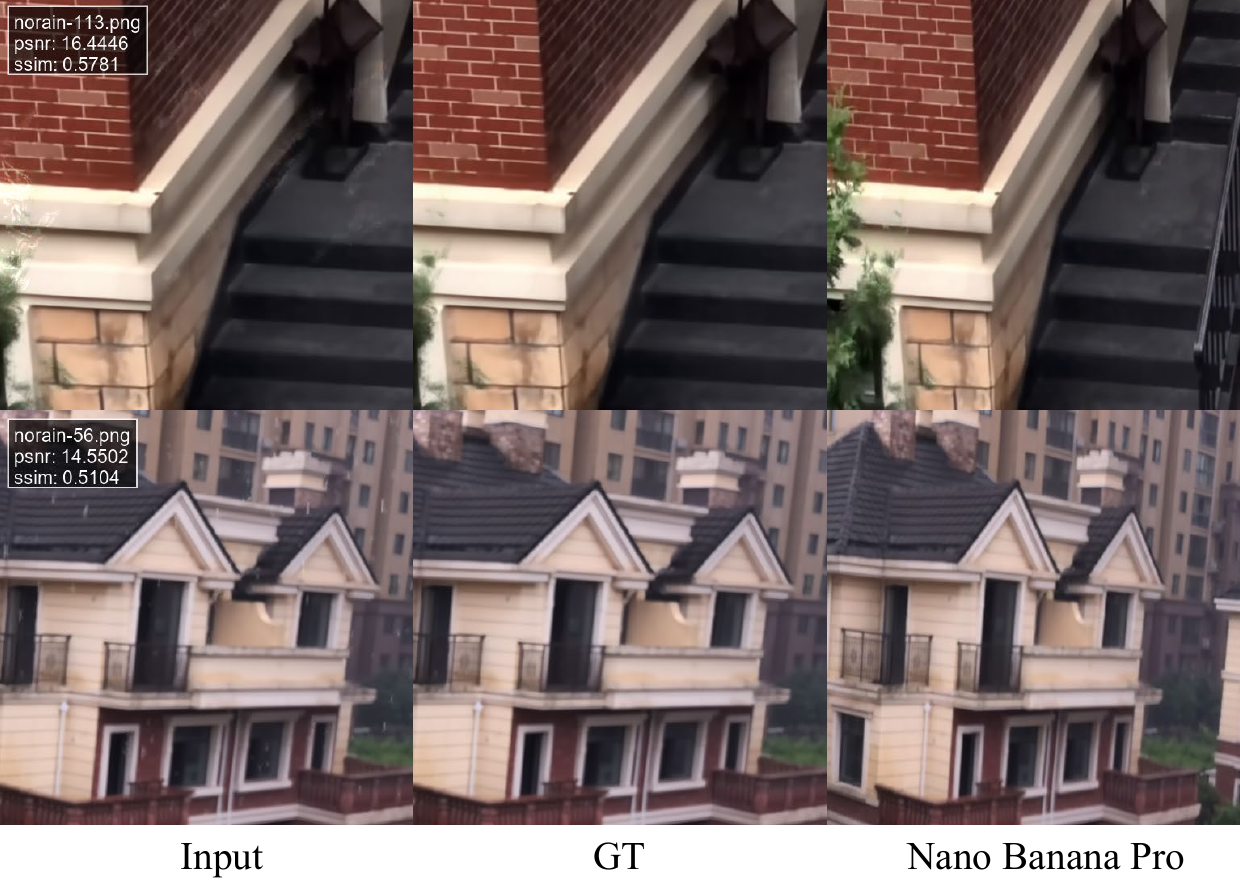}
    \centering
    \caption{Failure case of Nano Banana Pro. From left to right: input rainy image, ground truth (GT), and the output of Nano Banana Pro. The model hallucinates plausible content to fill severely corrupted regions, leading to a significant pixel-level discrepancy from the GT.}
    \label{fig:deraining_fig3}
    \vspace{-5mm}
\end{figure}

Our qualitative analysis reveals a fundamental limitation in the instruction-following behavior of prompt-conditioned multimodal generative models. As shown in Fig.~\ref{fig:deraining_fig3}, despite utilizing explicit prompts that constrain the model to preserve non-rain regions, we consistently observe unintended alterations in background elements.

\begin{figure}[htbp]
    \centering
    \includegraphics[width=0.6\textwidth]{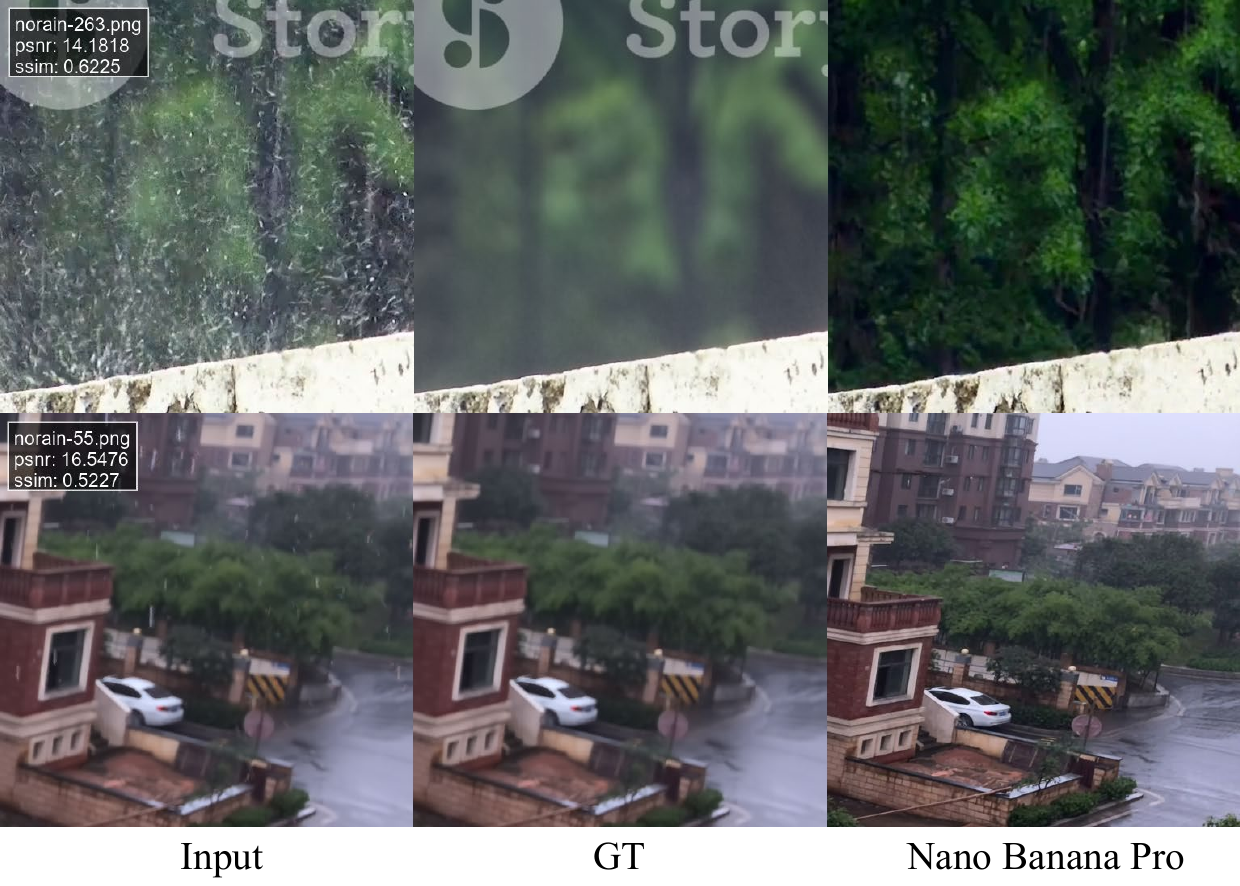}
    \centering
    \caption{Failure case of Nano Banana Pro. The model inadvertently removes the atmospheric haze accompanying rain streaks; while this enhances visual clarity, it results in lower quantitative metrics due to deviation from the GT.}
    \label{fig:deraining_fig4}
    \vspace{-3mm}
\end{figure}

This phenomenon stems from an inherent bias in fine-grained semantic understanding. As clearly illustrated in Fig.~\ref{fig:deraining_fig4}, the model conflates rain streaks with atmospheric haze (i.e., the rain-mist ambiguity). Consequently, it aggressively removes the mist alongside the rain, yielding an output image that appears clearer and visually superior in low-level details. However, this visual enhancement paradoxically leads to lower quantitative scores (e.g., PSNR), as the complete removal of haze introduces a significant pixel-wise deviation from the ground truth. This observation underscores the prevalent perception-distortion trade-off in image restoration tasks.

\subsection{Analyses}
In this study, we investigated the feasibility of Nano Banana Pro for single-image deraining under a zero-shot setting. Experimental results indicate that the application of generative models to image restoration presents a ``double-edged sword'' effect. On one hand, compared to specialized deraining models trained on extensive supervised data, such as NeRD-Rain and Restormer, Nano Banana Pro exhibits a significant gap in pixel-level objective metrics like PSNR and SSIM (e.g., achieving a PSNR of only 21.10 dB on the Rain200H dataset). This quantitative deficiency is primarily attributed to the inherent tendency of generative models to prioritize semantic reconstruction over strict pixel-wise restoration, resulting in deviations in high-frequency detail preservation and color fidelity. On the other hand, leveraging its robust world model and semantic reasoning capabilities, Nano Banana Pro demonstrates superior structural coherence and visual plausibility compared to traditional methods when handling regions with severe rain occlusion (such as bridge cables). This confirms the unique complementary advantages of generative models in addressing extreme degradation and information loss. Future work will focus on employing prompt tuning to further enhance pixel-level restoration accuracy for low-level vision tasks, while preserving the model's strong semantic generation capabilities.
\section{Shadow Removal}

\subsection{Background}
Shadow removal aims to eliminate cast shadows from images and restore consistent illumination between shadow and non-shadow regions~\cite{guo2012paired}. Shadows are caused by partial occlusion of light by scene objects and are ubiquitous in natural environments. Although shadows provide useful geometric cues for human perception, they often introduce strong intensity discontinuities, color distortions, and loss of texture details, which severely degrade the performance of downstream vision tasks such as object detection, tracking, and segmentation. Indeed, removing shadows from an image remains a fundamental yet highly challenging low-level vision task.

Early shadow removal methods~\cite{qu2017deshadownet} primarily relied on handcrafted priors and carefully designed statistical features, such as illumination consistency, gradient constraints, and region smoothness. These optimization-based approaches were built upon highly idealized physical and photometric assumptions, which rarely hold in real-world scenes with complex lighting, textured backgrounds, and soft shadow boundaries. Consequently, they often suffer from noticeable artifacts, especially around shadow boundaries, and fail to generalize to diverse real-world scenarios.

With the rapid development of deep learning, fully supervised CNN-based shadow removal methods have achieved remarkable progress by learning pixel-wise mappings from shadow images to their shadow-free counterparts using large-scale paired datasets. While these methods significantly improve visual quality on benchmark datasets, they heavily rely on expensive pixel-level annotations and often exhibit severe overfitting with limited generalization capability. More importantly, shadow removal is inherently a region-wise corrupted problem that involves strong contextual and structural priors.

The recent introduction of Nano Banana Pro demonstrates remarkable capabilities of generative models in visual tasks. In this context, we systematically evaluate the performance of Nano Banana Pro on the single-image shadow removal task. Through comprehensive comparisons with existing representative methods, we provide an in-depth analysis of its strengths and limitations in real-world applications.

\subsection{Qualitative and Quantitative Results}

\begin{figure*}[ht]  
  \centering  
  \includegraphics[width=0.7\textwidth]{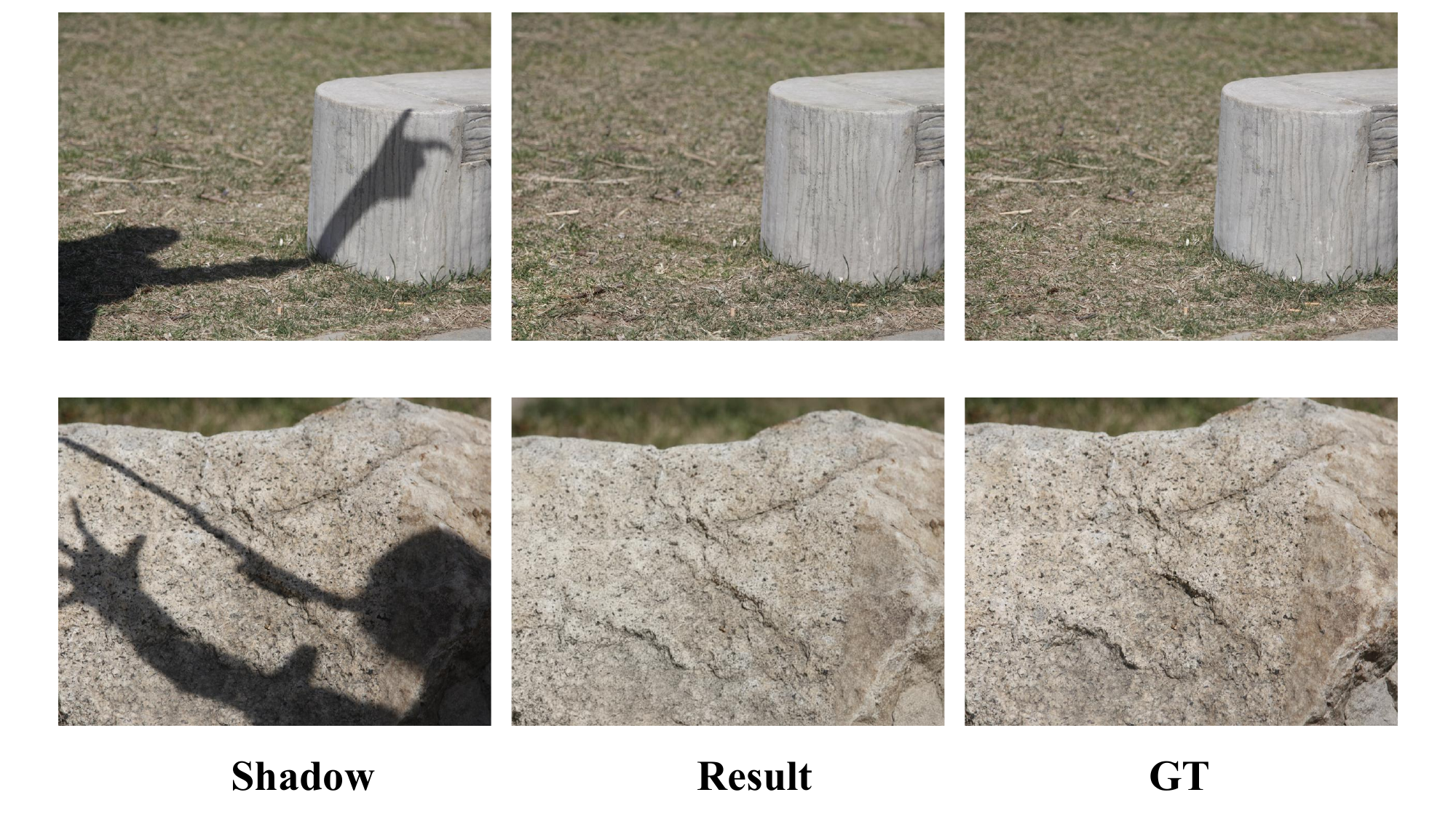}  
  \caption{Some well-performing visual examples of NB Pro on the SRD dataset~\cite{qu2017deshadownet} for the shadow removal task.} 
  \label{deshaow_figure}  
\end{figure*}

Fig.~\ref{deshaow_figure} presents the well-performing shadow removal results of NB Pro on the SRD dataset~\cite{qu2017deshadownet}. It can be observed that NB Pro effectively removes shadows from the image while highly preserving the original elements without alteration. Tab.~\ref{tab:shadow_results} presents the quantitative comparison on SRD dataset of NB Pro against state-of-the-art shadow removal methods, using PSNR and SSIM as the primary evaluation metrics.

\begin{table}[ht]
\centering
\small
\caption{Quantitative comparisons on the SRD dataset~\cite{qu2017deshadownet}. The best results are highlighted by \textbf{black bold.}}
\label{tab:shadow_results}
\resizebox{0.6\textwidth}{!}{%
\begin{tabular}{lcc}
\toprule
Method & PSNR $\uparrow$ & SSIM $\uparrow$ \\
\midrule
DSC \cite{hu2018direction} (TPAMI'19) & 27.76 & 0.903 \\
DHAN \cite{cun2020towards}(AAAI'20) & 30.51 & 0.949 \\
BMNet \cite{zhu2022bijective}(CVPR'22) & 31.69 & 0.956 \\
ShadowFormer\cite{guo2023shadowformer} (AAAI'23) & 32.90 & 0.958 \\
ShadowDiffusion \cite{guo2023shadowdiffusion} (CVPR'23) & 34.73 & 0.970 \\
HomoFormer \cite{xiao2024homoformer} (CVPR'24) & \textbf{35.37} & \textbf{0.972} \\
\rowcolor{lightgray!50}
\textbf{\textcolor{orange}{Nano Banana Pro}} & \textbf{\textcolor{orange}{20.67}} & \textbf{\textcolor{orange}{0.682}} \\
\bottomrule
\end{tabular}%
}
\end{table}
As shown in Tab.~\ref{tab:shadow_results}, a significant discrepancy exists between the visual quality discussed earlier and the numerical fidelity scores. While leading methods such as ShadowDiffusion~\cite{guo2023shadowdiffusion} and HomoFormer~\cite{xiao2024homoformer} achieve PSNR scores exceeding 34 dB and SSIM values above 0.97, NB Pro records comparatively lower scores, with a PSNR of 20.67 dB and SSIM of 0.682. This quantitative gap can be attributed to the inherent characteristics of generative models: NB Pro prioritizes perceptual plausibility and visual naturalness over strict pixel-wise alignment with ground truth references. Unlike traditional methods that focus on precise reconstruction, generative approaches like NB Pro tend to produce images with enhanced visual appeal, which may deviate from the exact pixel values of ground truth images, resulting in lower scores on fidelity-based metrics.

\begin{figure*}[ht]  
  \centering  
  \includegraphics[width=0.7\textwidth]{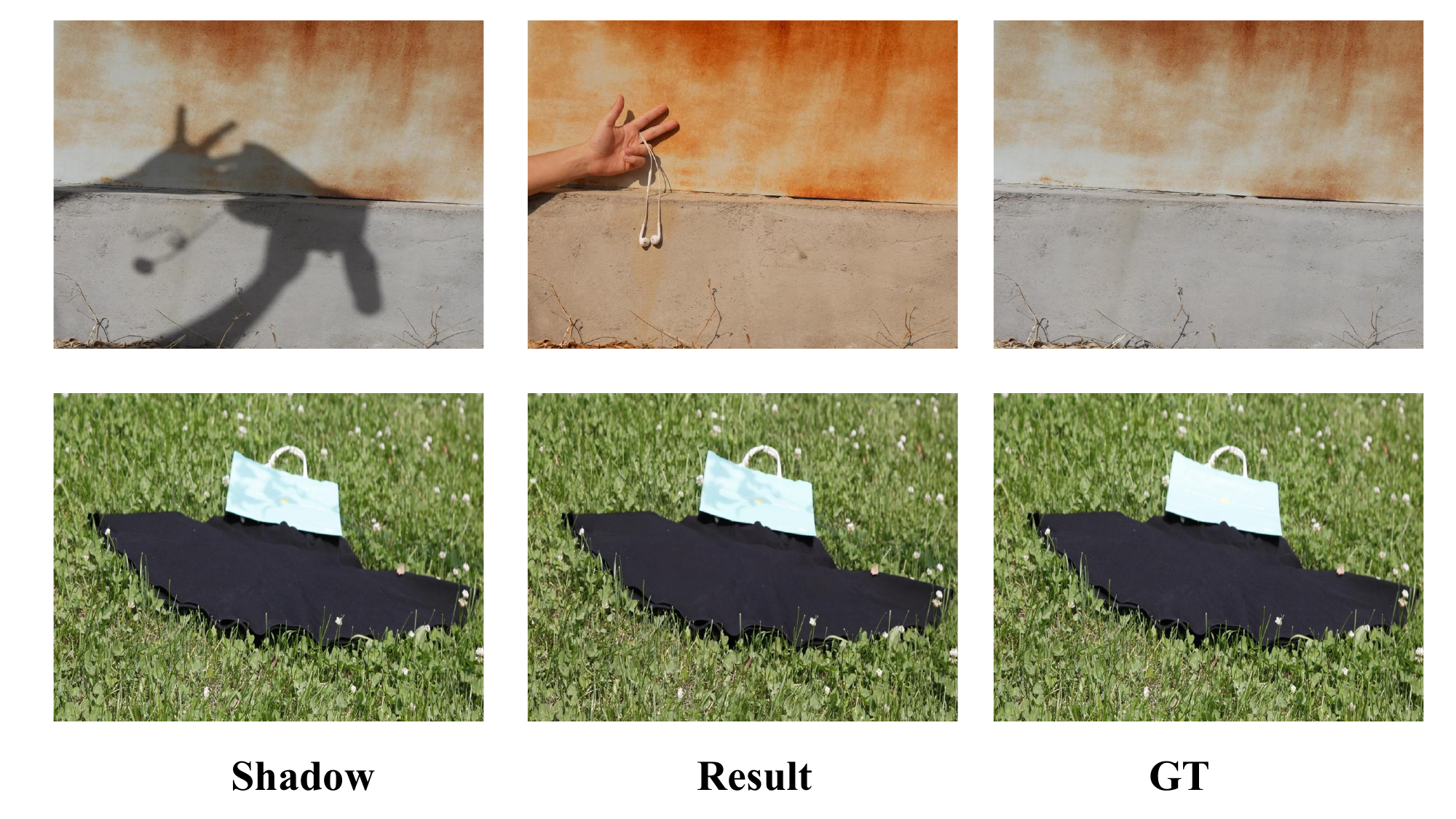}  
  \caption{Some visual failure examples of NB Pro on the SRD dataset~\cite{qu2017deshadownet} for the shadow removal task.} 
  \vspace{-5pt}
  \label{deshaow_figure2}  
\end{figure*}

In addition to its successful cases, we systematically document representative failure modes of NB Pro in shadow removal, as compiled in Fig.~\ref{deshaow_figure2}. These examples reveal two core, recurring limitations that stem from the model's generative nature: a propensity for over-generation and a blindness to subtle shadows.

A consistent pattern of failure emerges across the SDR dataset, revealing critical limitations in NB Pro’s approach. \textit{The first failure mode stems from the model's inherent generative bias.} Driven by a compulsion to produce visually 'complete' scenes, NB Pro often prioritizes hallucinated content over fidelity. As illustrated in Fig.~\ref{deshaow_figure2}, while the cast shadow is successfully removed, the model erroneously synthesizes a new hand to fill the void, fundamentally compromising the scene's semantic integrity. This highlights a tension between the model’s creative instinct and the strict fidelity required for low-level vision, where structural preservation is paramount. \textit{The second failure mode exposes a lack of sensitivity in shadow detection.} NB Pro frequently overlooks soft, low-contrast, or faint shadows (as seen in the second example), leaving them entirely untreated. This suggests a potential bias in the training data or optimization objectives that under-weights subtle illumination changes. Furthermore, the model struggles with color fidelity, frequently exhibiting shifts in tone and saturation that deviate from the ground truth. Collectively, these structural hallucinations, detection failures, and color shifts lead to unsatisfactory quantitative performance.

\subsection{Analysis}
Nano Banana Pro demonstrates a remarkable ability to decouple shadow components from the underlying reflectance. However, its effectiveness is fundamentally constrained by its generative nature, which presents a significant paradigm mismatch with the strict fidelity requirements of shadow removal.

First, the model’s inherent generative bias prioritizes perceptual plausibility over structural fidelity. As a generative model, NB Pro tends to hallucinate details, alter textures, or even synthesize new objects to create visually complete scenes. While this enhances aesthetic appeal, it compromises pixel-level accuracy, causing the output to deviate from a truthful reconstruction of the original reflectance. Consequently, NB Pro is better suited for creative applications requiring visual realism rather than low-level vision tasks demanding precise photometric or geometric correspondence.

Second, the model exhibits limited sensitivity in shadow detection. This limitation is likely attributable to biases in its training data or optimization objectives. If the training distribution under-represents subtle, soft, or low-contrast shadows, the model fails to learn the necessary features to identify them. As a result, NB Pro often leaves faint shadows untreated or, conversely, erroneously alters well-lit areas in an attempt to enforce uniform illumination, introducing artifacts in non-shadow regions.

Finally, these limitations are exacerbated by the incompatibility between generative outputs and traditional evaluation metrics. NB Pro typically generates images at fixed, high resolutions (e.g., 1K or 4K). Downsampling these outputs to match benchmark ground-truth resolutions smooths out high-frequency details, artificially depressing scores on pixel-wise metrics like PSNR and SSIM. More critically, a paradox arises where the model’s generative enhancements, such as implicit super-resolution or denoising, are heavily penalized as deviations from the ground truth. This stark divergence between perceptual quality and quantitative scores underscores the inadequacy of current fidelity-based frameworks for evaluating generative restoration models.
\section{Motion Deblurring}
\subsection{Background}
Motion blur, caused by camera shake or fast-moving objects, remains one of the most pervasive artifacts degrading image quality in photography and computer vision tasks. Over the past decade, deep learning approaches have revolutionized dynamic scene deblurring. Early CNNs, such as DeepDeblur~\cite{nah2017deep} and DeblurGAN~\cite{kupyn2018deblurgan}, paved the way for more sophisticated architectures. Recently, Transformer-based models like Uformer~\cite{wang2022uformer} and Restormer~\cite{zamir2022restormer} have dominated the field, achieving record-breaking scores in peak PSNR by effectively modeling long-range dependencies. However, these regression-based methods, typically optimized via MSE loss, tend to produce overly smooth results, often sacrificing high-frequency textures in favor of minimizing pixel-level error.

To overcome the "smoothing effect" and restore realistic details, the focus has shifted toward generative models, including GANs and Diffusion Models (e.g., ID-CDM~\cite{wang2026zero}, HI-Diff~\cite{chen2023hierarchical}). These approaches leverage strong generative priors, creating plausible textures for missing details. While they significantly enhance perceptual quality, they introduce a critical challenge: the perception-distortion trade-off. As the model strives to generate sharper and more visually pleasing images, it risks drifting away from the ground truth fidelity, potentially creating artifacts or "hallucinating" content that does not exist in the original scene.

In this technical report, we present a comprehensive evaluation of Nano Banana Pro (NB Pro). Our investigation reveals a fundamental limitation in this aggressive generative approach. While NB Pro demonstrates exceptional capability in synthesizing sharp textures for static environments and text, specifically in challenging low-light scenarios, it suffers from severe semantic instability. Our quantitative analysis (Tab.~\ref{tab:motiondeblur}) and visual inspection confirm that the model's pursuit of visual sharpness often comes at the cost of fidelity to the input signal.

Specifically, we observe that NB Pro struggles with complex motion trajectory reduction, often misinterpreting motion cues as structural elements, which leads to ghosting artifacts. Furthermore, the model exhibits a tendency to alter semantic information, such as facial identities and text characters. These generative behaviors result in comparatively low quantitative scores (PSNR/SSIM). By analyzing NB Pro's performance across synthetic (GoPro~\cite{nah2017deep}, HIDE~\cite{shen2019human}) and real-world (RealBlur~\cite{rim2020real}) benchmarks, this report aims to dissect the failure modes of generative deblurring, highlighting the significant gap between producing visually plausible images and maintaining structural accuracy.

\subsection{Qualitative Results}
We conduct a visual analysis of NB Pro on standard benchmark datasets(GoPro~\cite{nah2017deep}, HIDE~\cite{shen2019human} and RealBlur~\cite{rim2020real}) to evaluate its performance in restoring structural details and handling complex degradations.

\subsubsection{Performance on Synthetic Datasets}
\begin{figure}[htb]
    \centering
    \includegraphics[width=\linewidth]{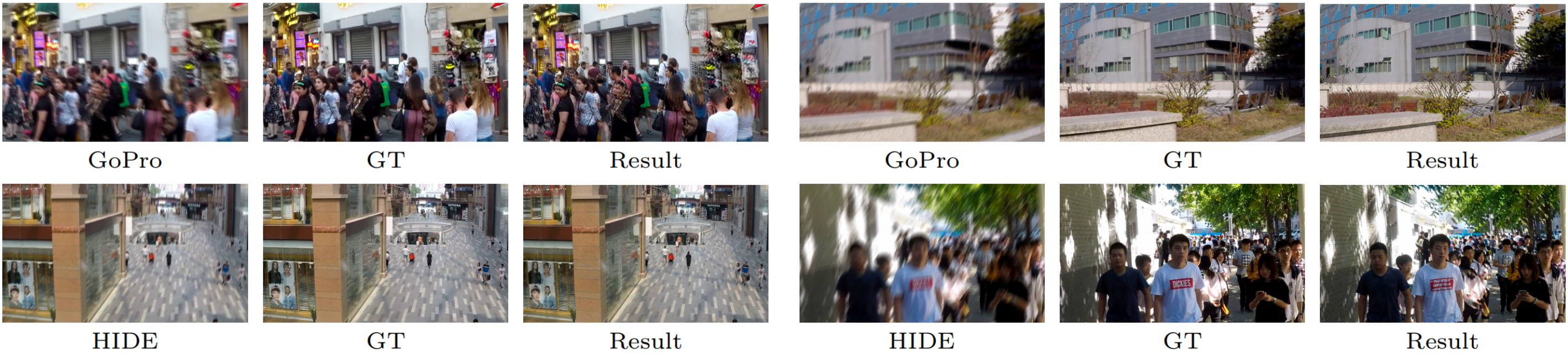}
    \caption{Visual results of Nano Banana Pro on synthetic blur datasets (GoPro and HIDE).}
    \label{fig:gopro&hide}
\end{figure}

NB Pro demonstrates impressive deblurring capabilities on synthetic datasets, particularly in recovering static environmental details. As observed in Fig.~\ref{fig:gopro&hide}, in the second column of GoPro and the first column of HIDE, the model effectively suppresses severe motion blur and restores high-frequency structures with great precision. Architectural elements, such as the building facades and pavement textures, are reconstructed with high fidelity. Notably, the model excels at text preservation in these scenarios; for instance, the "SEPHORA" signboard in the HIDE dataset is rendered clearly. This indicates strong spatial adaptability in handling rigid motion and structural edges.

However, significant limitations become apparent when processing highly dynamic scenes involving humans. The model struggles to fully eliminate complex synthetic motion trajectories, leading to noticeable residual artifacts. In the first column of GoPro, for example, the clothes hanging on the wall appear duplicated, and the woman’s headscarf exhibits a double-layer ghosting effect, suggesting an incomplete resolution of the motion path. Furthermore, the model tends to hallucinate facial details. While the restored faces in both GoPro (1st column) and HIDE (2nd column) appear visually sharp, they suffer from semantic inconsistencies. The facial features are altered to the extent that the identity of the pedestrians no longer matches the Ground Truth (GT), highlighting a critical lack of fidelity in semantic reconstruction.

\subsubsection{Performance on Real-World Datasets}
\begin{figure}[htb]
    \centering
    \includegraphics[width=\linewidth]{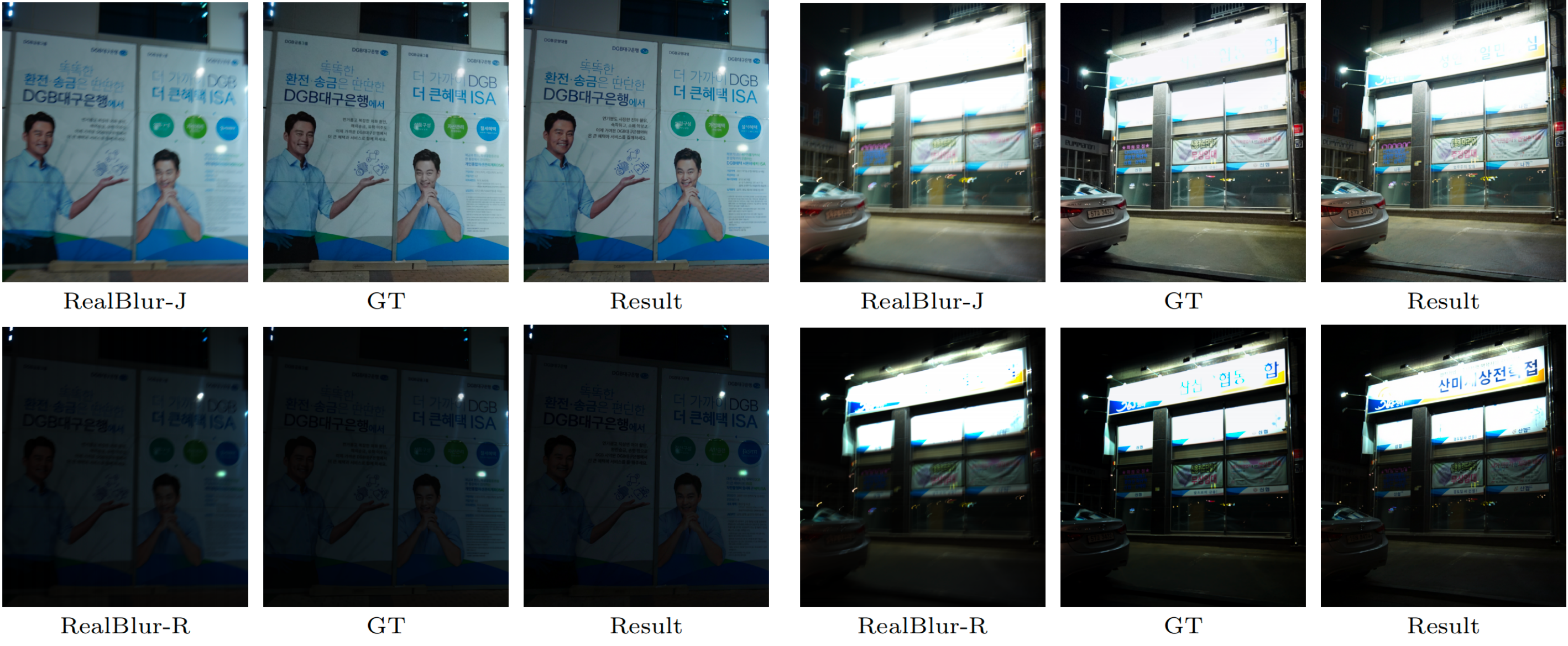}
    \caption{Visual results of Nano Banana Pro on real-world blur dataset (RealBlur-J and RealBlur-R).}
    \label{fig:realblur}
    \vspace{-3mm}
\end{figure}

On real-world datasets, NB Pro exhibits strong robustness against complex degradations such as low light and overexposure. As seen in Fig.~\ref{fig:realblur}, in the RealBlur-J dataset, the model successfully recovers the legibility of text on posters and signboards. Its ability to handle high dynamic range scenes is particularly noteworthy, in the storefront examples (2nd columns of both RealBlur-J and RealBlur-R), the model manages high-contrast lighting effectively. However, the result generated by NB Pro in the second column of RealBlur-R deviates significantly from the GT properties. While the output appears cleaner, it aggressively removes noise and alters lighting textures, resulting in a synthesized appearance that loses the atmosphere of the original scene.

Moreover, the model's reliance on generative priors introduces substantial perceptual deviations from the ground truth. In the poster examples of RealBlur-J (1st column), the restored facial features differ from the original image, creating a "hallucinated" face that does not preserve the subject's identity. Similar discrepancies are observed in the text content of the RealBlur-J (2nd column) result, where the generated characters deviate from the GT, such as pink characters on the window. This can be also observed in the second column of RealBlur-R, where the characters on the illuminated sign are significantly altered compared to GT. Additionally, color fidelity is occasionally compromised. For instance, in the first column of RealBlur-R, the skin tone of the person in the result exhibits a noticeable color shift compared to the target image. These issues indicate that while NB Pro excels at producing visually pleasing results, it sacrifices faithfulness to the original semantic content.

\subsection{Quantitative Results}
Tab.~\ref{tab:motiondeblur} presents the quantitative comparison of NB Pro against state-of-the-art deblurring methods, including transformer-based models like Uformer and Restormer, as well as recent diffusion-based approaches like HI-Diff and ID-CDM. The evaluation is performed on four standard benchmarks: GoPro~\cite{nah2017deep}, HIDEHIDE~\cite{shen2019human} and RealBlur~\cite{rim2020real}, using PSNR and SSIM as the primary metrics.

As observed in Tab.~\ref{tab:motiondeblur}, a significant divergence exists between the previously discussed visual sharpness and the numerical fidelity scores. While top-performing methods such as ID-CDM and HI-Diff achieve PSNR scores exceeding 33 dB on the GoPro dataset and 36 dB on RealBlur-R, NB Pro records comparatively lower values, such as 21.41 dB on GoPro and 21.35 dB on HIDE. Similarly, the SSIM scores for NB Pro range between 0.645 and 0.778, whereas competing methods consistently score above 0.90. This quantitative gap can be primarily attributed to the model's heavy reliance on strong generative priors, which prioritizes perceptual plausibility over strict pixel-wise alignment with the ground truth.

The lower PSNR and SSIM scores directly corroborate the limitations identified in the qualitative analysis. Standard metrics like PSNR are highly sensitive to pixel-level deviations. As noted in the visual evaluation, NB Pro tends to hallucinate high-frequency details, such as altering facial identities or modifying text characters, to maximize sharpness. These generated features, while appearing visually coherent, act as "errors" regarding the reference image, leading to heavy penalties in signal-to-noise calculations. Furthermore, the reported semantic inconsistencies, such as color shifts and the "double-layer ghosting" effects caused by misinterpreted motion trajectories, significantly disrupt structural similarity, resulting in the observed drop in SSIM. This confirms that the model's generated content often diverges from the underlying ground truth signal.

\begin{table*}[htb]
    \centering
    \caption{Quantitative comparison results of Nano Banana Pro and other representative methods on four benchmarks.}
    \label{tab:motiondeblur}
    \setlength{\tabcolsep}{7pt} 
    \renewcommand{\arraystretch}{1.2} 
    \begin{tabular}{l l l l l l l l l}
        \toprule 
        \multirow{2}{*}{Method} & \multicolumn{2}{l}{GoPro} & \multicolumn{2}{l}{RealBlur-R} & \multicolumn{2}{l}{RealBlur-J} & \multicolumn{2}{l}{HIDE} \\
        \cmidrule(r){2-3} \cmidrule(lr){4-5} \cmidrule(lr){6-7} \cmidrule(l){8-9} 
        & PSNR $\uparrow$ & SSIM $\uparrow$ & PSNR $\uparrow$ & SSIM $\uparrow$ & PSNR $\uparrow$ & SSIM $\uparrow$ & PSNR $\uparrow$ & SSIM $\uparrow$ \\
        \midrule 
        
        DeepDeblur~\cite{nah2017deep} & 29.08 & 0.914 & 32.51 & 0.841 & 27.87 & 0.827 & 25.73 & 0.874 \\
        GAMD~\cite{luan2024gyroscope} & 33.14 & 0.9284 & 34.00 & 0.9265 & -- & -- & -- & -- \\
        DeblurGAN~\cite{kupyn2018deblurgan} & 28.70 & 0.858 & 33.79 & 0.903 & 27.97 & 0.834 & 24.51 & 0.871 \\
        DeblurGAN-v2~\cite{kupyn2019deblurgan} & 29.55 & 0.934 & 35.26 & 0.944 & 28.70 & 0.866 & 26.61 & 0.875 \\
        DBGAN~\cite{zhang2020deblurring} & 31.10 & 0.942 & 33.78 & 0.909 & 24.93 & 0.745 & 28.94 & 0.915 \\
        Uformer-B~\cite{wang2022uformer} & 32.97 & 0.967 & 36.22 & 0.957 & 29.06 & 0.884 & 30.83 & \textbf{0.952} \\
        Stripformer~\cite{tsai2022stripformer} & 33.08 & 0.962 & 36.08 & 0.954 & 28.82 & 0.876 & 31.03 & 0.940 \\
        Restormer~\cite{zamir2022restormer} & 32.92 & 0.961 & 36.19 & 0.957 & 28.96 & 0.879 & 31.22 & 0.942 \\
        IR-SDE~\cite{luo2023image} & 30.70 & 0.901 & 33.96 & 0.918 & 24.21 & 0.729 & -- & -- \\
        DiffIR~\cite{xia2023diffir} & 33.20 & 0.963 & -- & -- & -- & -- & 31.55 & 0.947 \\
        HI-Diff~\cite{chen2023hierarchical} & \textbf{33.33} & 0.964 & 36.28 & \textbf{0.958} & \textbf{29.15} & \textbf{0.890} & 31.46 & 0.945 \\
        ID-CDM~\cite{wang2026zero} & 33.19 & \textbf{0.970} & \textbf{36.34} & 0.955 & 28.96 & 0.887 & \textbf{31.53} & 0.950 \\
        
        \rowcolor{lightgray!50}
\textbf{\textcolor{orange}{Nano Banana Pro}} & \textbf{\textcolor{orange}{21.41}} & \textbf{\textcolor{orange}{0.645}} & \textbf{\textcolor{orange}{27.43}} & \textbf{\textcolor{orange}{0.778}} & \textbf{\textcolor{orange}{24.51}} & \textbf{\textcolor{orange}{0.747}} & \textbf{\textcolor{orange}{21.35}} & \textbf{\textcolor{orange}{0.662}}\\
        
        \bottomrule 
    \end{tabular}
    
\end{table*}

\subsection{Analysis}
The discrepancy between NB Pro's superior visual sharpness and its lower quantitative scores stems primarily from the perception-distortion trade-off. While regression-based methods minimize pixel-wise error to maximize PSNR, often resulting in over-smoothed textures, NB Pro leverages generative priors to create high-frequency details. This strategy produces visually realistic textures but introduces stochastic pixel deviations from the ground truth. Since standard metrics like PSNR penalize any deviation equally, NB Pro receives lower scores despite operating at a higher level of perceptual quality.

Furthermore, the evaluation on real-world datasets reveals the specific limitations of the model's reconstruction capability. While the ground truth in RealBlur sometimes contains noise or light scattering, NB Pro's tendency to completely "clean" these elements represents a deviation from the scene's authentic characteristics. Rather than simply restoring the signal, the model synthesizes a new, idealized version of the image. This leads to low quantitative scores not just because of metric limitations, but because the model fails to preserve the original distribution of the input data, effectively altering the scene's atmosphere.

Consequently, this reliance on generative priors introduces significant risks regarding semantic fidelity. When motion blur obliterates structural information, the model synthesizes plausible but factually incorrect content, leading to the observed "identity swaps" in human faces and ghosting artifacts in complex motion paths. While NB Pro excels in perceptual synthesis, making it suitable for visually-oriented restoration, these semantic inconsistencies highlight its unsuitability for applications requiring strict adherence to the original input signal, such as forensic analysis or high-fidelity surveillance.
\section{Defocus Deblurring}
\subsection{Introduction}
Defocus blur, an inherent optical phenomenon resulting from limited depth of field and aperture configurations, presents one of the most complex challenges in computational photography and low-level vision. Unlike uniform degradations such as global motion blur, defocus acts as a spatially varying aberration where the point spread function (PSF) changes according to the scene depth. This results in a non-uniform loss of high-frequency details and edge information, making the restoration process an ill-posed inverse problem that requires estimating spatially adaptive kernels.

The field of single-image defocus deblurring has advanced significantly with the advent of deep learning. Early data-driven approaches, such as DPDNet~\cite{abuolaim2020defocus}, established strong baselines by leveraging dual-pixel data to supervise defocus removal. Subsequent architectures, including IFANet~\cite{lee2021iterative} and KPAC~\cite{son2021single}, introduced iterative filtering and kernel prediction mechanisms to better handle spatially varying blur. More recently, the introduction of Transformer-based architectures, such as Restormer~\cite{zamir2022restormer}, and multi-stage networks like MPRNet~\cite{mehri2021mprnet}, has pushed the boundaries of restoration fidelity by capturing long-range dependencies and global context. Current state-of-the-art methods, such as GGKMNet~\cite{quan2024deep}, further refine this process by integrating grouped kernel modeling to precisely invert the blurring process across complex depth maps.

In this section, we extend our evaluation of the Nano Banana Pro (NB Pro) to the domain of defocus deblurring. Unlike the aforementioned supervised methods, which are trained specifically on paired defocus datasets, our investigation focuses on assessing the NB Pro model in a zero-shot inference setting. By benchmarking the model against standard datasets such as DPDD~\cite{abuolaim2020defocus} and RealDOF~\cite{ruan2021aifnet}, we aim to analyze its efficacy in handling the inverse problem of deblurring without domain-specific fine-tuning. Specifically, we examine whether the model's processing pipeline can genuinely recover lost structural information comparable to established specialized networks, or if it merely relies on superficial enhancement techniques.

\subsection{Quantitative Results}
\begin{table}[htbp]
  \centering
  \caption{Quantitative comparison on DPDD and RealDOF datasets. The best results are highlighted in \textbf{black bold.}}
  \label{tab:defocusdeblur}
  \begin{tabular}{lcccc}
    \toprule
    \multirow{2}{*}{Model} & \multicolumn{2}{c}{DPDD} & \multicolumn{2}{c}{RealDOF} \\
                           & PSNR $\uparrow$ & SSIM $\uparrow$ & PSNR $\uparrow$ & SSIM $\uparrow$ \\
    \midrule
    DPDNet~\cite{abuolaim2020defocus}                 & 24.348 & 0.747 & 22.870 & 0.670 \\
    AIFNet~\cite{ruan2021aifnet}                 & 24.213 & 0.742 & 23.093 & 0.680 \\
    IFANet~\cite{lee2021iterative}                 & 25.366 & 0.789 & 24.712 & 0.748 \\
    KPAC~\cite{son2021single}                   & 25.221 & 0.774 & 23.975 & 0.762 \\
    GKMNet~\cite{quan2021gaussian}                 & 25.468 & 0.789 & 24.254 & 0.732 \\
    MDP~\cite{abuolaim2022improving}                    & 25.347 & 0.763 & 23.500 & 0.681 \\
    DRBNet~\cite{ruan2022learning}                 & 25.485 & 0.792 & 24.884 & 0.751 \\
    MPRNet~\cite{mehri2021mprnet}                 & 25.730 & 0.792 & 24.541 & 0.736 \\
    Restormer~\cite{zamir2022restormer}              & 25.980 & \textbf{0.811} & 25.091 & 0.762 \\
    INIKNet~\cite{quan2023single}                & 26.055 & 0.803 & 25.231 & 0.765 \\
    NRKNet~\cite{quan2023neumann}                 & 26.109 & 0.810 & 25.148 & 0.768 \\
    GGKMNet~\cite{quan2024deep}                & \textbf{26.272} & 0.810 & \textbf{25.355} & \textbf{0.770} \\
    \rowcolor{lightgray!50}
\textbf{\textcolor{orange}{NB Pro}}                 & \textbf{\textcolor{orange}{20.180}} & \textbf{\textcolor{orange}{0.635}} & \textbf{\textcolor{orange}{20.821}} & \textbf{\textcolor{orange}{0.641}} \\
    \bottomrule
  \end{tabular}
\end{table}

Tab.~\ref{tab:defocusdeblur} presents the quantitative evaluation on the
DPDD~\cite{abuolaim2020defocus} and RealDOF~\cite{ruan2021aifnet} datasets. The results indicate a substantial performance gap between the Nano Banana Pro and established defocus deblurring methods. On the DPDD dataset, the Nano Banana Pro yields a PSNR of 20.180 dB and an SSIM of 0.635, significantly trailing the state-of-the-art GGKMNet by over 6 dB. A similar deficiency is evident on the RealDOF dataset, where NB Pro lags behind even early baselines like DPDNet. These low metrics align consistently with our qualitative findings: the depressed PSNR reflects the model's general failure to restore pixel-level sharpness, while the low SSIM corroborates the structural hallucinations and inability to remove blur observed in the visual comparison. However, it is worth noting that the competing methods, such as Restormer and DPDNet, are supervised models trained directly on the DPDD dataset, whereas NB Pro is evaluated here in a zero-shot setting without domain-specific training.

\subsection{Qualitative Results}
We evaluated the perceptual performance of the Nano Banana Pro by conducting a comprehensive visual analysis on the DPDD and RealDOF datasets. The results indicate a consistent limitation in the model’s ability to recover high-frequency details from severe defocus, with the model frequently prioritizing global contrast enhancement over effective blur removal.

\begin{figure}[htb]
    \centering
    \includegraphics[width=\linewidth]{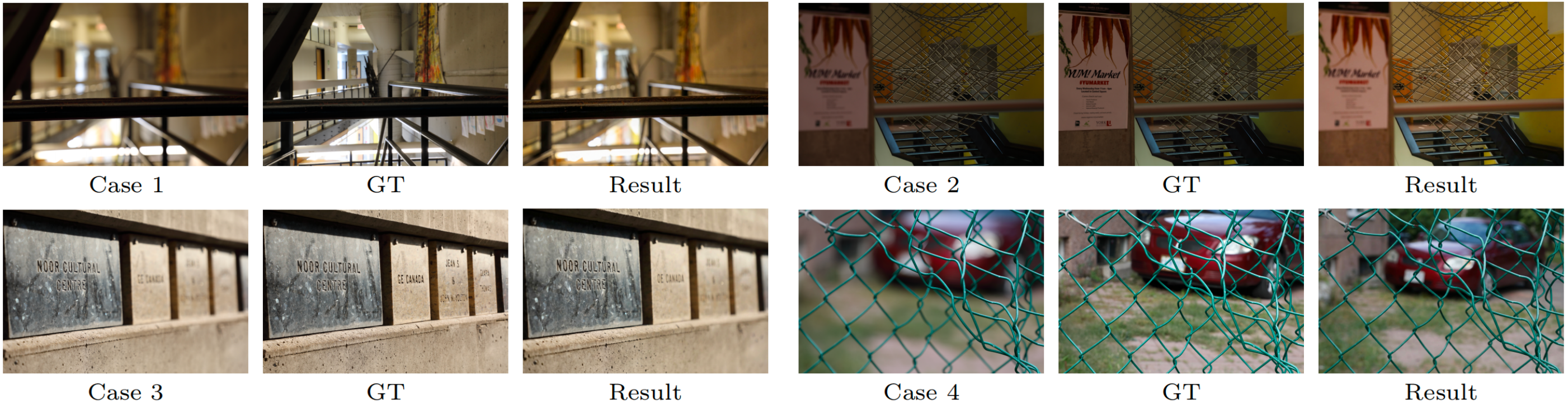} 
    \caption{Some representative qualitative results of Nano Banana Pro on the DPDD dataset.}
    \label{fig:dpdd}
\end{figure}

On the DPDD dataset, NB Pro behaves more akin to an image enhancement filter than a specialized deblurring network. As seen in Fig.~\ref{fig:dpdd}, in scenarios such as Case 1 and Case 2, the primary modification to the input is a global increase in luminance and contrast. While this improves the visual punch of the image, it fails to address the underlying degradation. Specifically, the severely defocused foreground in Case 2 remains blurry, and the background bokeh in Case 1 is only marginally reduced in spread. Furthermore, the model exhibits instability in structural reconstruction. This is evident in Case 3, where the restoration process hallucinates semantic details, incorrectly recovering the text "GE CANADA" as "OE CANADA" despite only a slight improvement in sharpness. Similarly, in Case 4, while the foreground fence is adequately sharpened, it compromises geometric fidelity, resulting in an unexplained scale alteration of the red vehicle in the background.

\begin{figure}[htb]
    \centering
    \includegraphics[width=\linewidth]{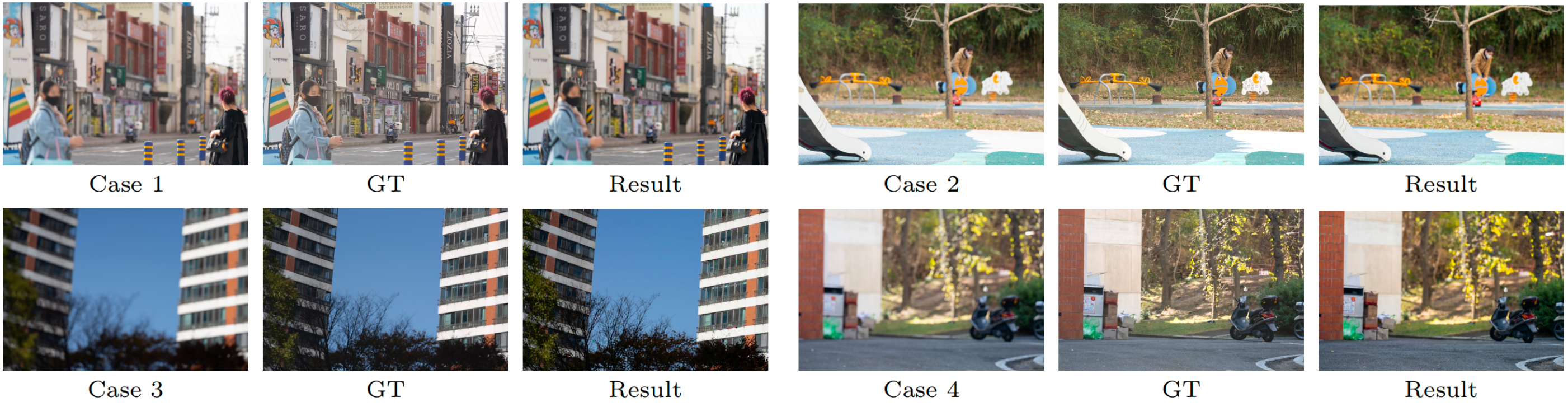} 
    \caption{Some representative qualitative results of Nano Banana Pro on the RealDOF dataset.}
    \label{fig:realdof}
\end{figure}

The limitations of NB Pro are even more pronounced in the RealDOF dataset evaluations, where the model demonstrates a negligible deblurring effect across multiple test cases. As seen in Fig.~\ref{fig:realdof}, in scenes with spatially varying blur, such as the mid-range focus in  Case 1 and the foreground focus in Case 2, the model fails to reverse the defocus entirely. The output images are characterized solely by a slight boost in contrast, leaving the blurred regions perceptually identical to the input. While the model achieves a degree of sharpness recovery in the fully blurred scenario of Case 3, this comes at the cost of introducing high-frequency artifacts, specifically salt-and-pepper noise visible on the building structures. Moreover, the restoration in Case 4 is depth-limited. The model successfully restores the ground texture closest to the lens but fails to extend the depth of field to the background, which remains in a state of defocus with only a minor reduction in the circle of confusion. Collectively, these qualitative results suggest that the Nano Banana Pro lacks the robustness required for consistent defocus deblurring, often failing to produce a discernible improvement in image sharpness.

\subsection{Analysis}
The disparity between the quantitative metrics and the qualitative visual outputs reveals the inherent instability of applying a general-purpose generative model to the specific physical constraints of defocus deblurring. Our analysis suggests that the Nano Banana Pro does not perform a mathematical inversion of the optical point spread function. Instead, it relies on semantic-aware generative priors to synthesize sharp details. This results in a bimodal behavior where the model oscillates between superficial contrast adjustment and aggressive, perceptually driven reconstruction depending on the scene's semantic recognisability.

In complex scenes with varying depth and high-frequency clutter, such as those in the DPDD dataset, the model's generative mechanism often struggles to identify coherent structural cues. Consequently, the model defaults to a global contrast maximization approach. This explains the consistently low PSNR values observed in Tab.~\ref{tab:defocusdeblur}. While increasing local contrast can improve perceptual punch in slightly out-of-focus regions, it fails to mathematically invert the point spread function. Consequently, the model exhibits inconsistent restoration behaviors, resorting to superficial contrast adjustment when semantic cues are ambiguous, while attempting aggressive reconstruction in scenes with recognizable structures.

However, in scenarios with regular, recognizable structures such as the building facade in Fig.~\ref{fig:realdof} Case 3, the model successfully engages its learned priors to "re-paint" the geometry, effectively removing the blur. Yet, this reconstruction is perceptually driven rather than physically constrained, leading to high-frequency artifacts. The salt-and-pepper noise observed along the window frames is likely a byproduct of the generative process (e.g., instability in the diffusion sampling) attempting to force high-frequency gradients into latent features that do not perfectly align with the degraded input.

Conversely, when the defocus aligns with typical photographic aesthetics, the model exhibits passivity. This is clearly observed in Fig.~\ref{fig:realdof} Case 4, where the background exhibits only mild defocus and remains semantically distinguishable, yet the model sharpens only the foreground ground texture while leaving the background blur intact. This divergence strongly suggests that the model’s priors interpret the slight background defocus as an intended aesthetic attribute, specifically, as a natural depth-of-field, rather than a degradation requiring correction. Unlike dedicated deblurring networks that aim to minimize the circle of confusion globally, NB Pro appears to prioritize perceptual naturalness, effectively treating the background blur as context to be preserved rather than an error to be inverted.

Finally, the lack of fidelity constraints in this zero-shot setting leads to significant structural deviations. Since the model prioritizes perceptual plausibility over pixel-wise accuracy, it introduces semantic errors when input ambiguity is high (such as hallucinating "OE CANADA" instead of "GE CANADA" in Fig.~\ref{fig:realdof} and altering the geometric scale of the vehicle in Fig.~\ref{fig:dpdd}). These behaviors confirm that the Nano Banana Pro operates as an image re-synthesis engine rather than a dedicated restoration tool, resulting in low quantitative scores (PSNR/SSIM) despite occasional visual successes.

\section{Denoising}
\label{sec:denoising}

\subsection{Background}
Recent advancements in vision-language models~\cite{liu2023visual,bai2025qwen2,lu2025hyper} mark a significant paradigm shift toward unified architectures capable of integrating diverse modalities and tasks within a single framework. Notably, models such as Nano Banana Pro have demonstrated that the synergistic training of understanding and generation objectives can unlock emergent capabilities and enhance cross-task generalization.

Despite the architectural elegance and representation efficiency offered by unified models~\cite{lu2025hyper}, a critical question persists: Can these generalist systems rival the precision of dedicated restoration networks~\cite{murali2020image} in specialized low-level vision tasks? Image denoising, specifically, stands as a rigorous litmus test. It evaluates a model's capacity to preserve fine-grained details, textures, and structural fidelity, which are intrinsic not only to restoration but also to high-fidelity image generation and editing.

In this technical report, we conduct a systematic evaluation of Nano Banana Pro’s denoising performance across five established benchmark suites: McMaster~\cite{mcmaster} for natural image statistics, Kodak24~\cite{Kodak} for photographic quality assessment, Urban100~\cite{urban} for challenging high-frequency texture reconstruction, and PolyU~\cite{polyu} and SIDD-small~\cite{sidd} for real-world sensor noise suppression. This study serves a dual purpose: first, to determine whether Nano Banana Pro’s unified training regime yields competitive low-level restoration quality; and second, to elucidate the interplay between generative capabilities and fine-grained reconstruction, providing insights to guide the design of future unified architectures.

\subsection{Experimental Setup}
NanoBanana is a closed-source unified multimodal model accessed through its official API. As a model capable of image understanding, generation, and text-driven editing, we evaluate its denoising capability by providing noisy images alongside a natural language instruction. The prompt used throughout our experiments is: ``This is a noisy image, please remove the noise in this image while keep other elements in this image unchanged.''

\textbf{Datasets.} We evaluate Nano Banana Pro on five widely-used image denoising benchmarks spanning both synthetic and real-world noise scenarios. For synthetic noise datasets like McMaster~\cite{mcmaster}, Kodak24~\cite{Kodak}, and Urban100~\cite{urban}, we corrupt clean images with additive white Gaussian noise at a fixed noise level of $\sigma$ = 50, representing a challenging high-noise regime. McMaster contains 18 high-resolution images with rich color and texture, Kodak24 comprises 24 classic photographic images, and Urban100 includes 100 images with complex urban structures and repetitive patterns that stress high-frequency reconstruction. For real-world noise datasets like PolyU~\cite{polyu} and SIDD-small~\cite{sidd}, we use the standard noisy/clean image pairs provided by each benchmark without additional synthetic corruption. These datasets capture realistic sensor noise from various camera devices under diverse lighting conditions, presenting a more practical evaluation scenario.

\textbf{Resolution and Failure Case Handling.} Nano Banana Pro outputs images at approximately 1K resolution, though the exact dimensions vary across samples (e.g., 1024×1024, 1200×896, 720×1456). We resize the output images to match the resolution of corresponding ground truth images using bilinear interpolation. All metrics are then computed between the resized outputs and the ground truths. In addition, during evaluation, we observed that Nano Banana Pro occasionally produces outputs that are either semantically irrelevant to the input image or fail to remove noise effectively. In such cases, we regenerate the output by resubmitting the same input and prompt to the API until a valid denoised result is obtained. This protocol ensures that our quantitative metrics reflect the model's denoising capability under successful generation, while the occurrence of such failures is noted as a limitation of applying unified generative models to restoration tasks.

\textbf{Evaluation Metrics.}
We adopt two complementary metrics to assess denoising quality: PSNR (Peak Signal-to-Noise Ratio), which measures pixel-level fidelity. SSIM (Structural Similarity Index): Evaluates perceptual structural similarity. Both are computed on RGB channels

\subsection{Quantitative and Qualitative Results}
To systematically evaluate the image denoising capabilities of Nano Banana Pro, we invoked the model via its official API and compared its performance against five representative task-specific baselines (DnCNN~\cite{dncnn}, Restormer~\cite{zamir2022restormer}, MaskDenoising~\cite{maskdenoising}, HAT~\cite{HAT}, and DIL~\cite{DIL}). The evaluation spans two distinct regimes. First, we employed three synthetic benchmarks—McMaster~\cite{mcmaster}, Kodak24~\cite{Kodak}, and Urban100~\cite{urban}—corrupted with additive Gaussian noise ($\sigma=50$) to test reconstruction across varying complexities. Specifically, McMaster assesses basic noise removal in smooth textures; Kodak24 covers diverse natural scenes to balance texture and color fidelity; and Urban100 challenges the model's ability to preserve high-frequency details within complex architectural structures. Complementing these synthetic tests, we assessed real-world blind denoising performance using SIDD Val~\cite{sidd} and PolyU~\cite{polyu}, where no prior noise information is provided. SIDD Val serves as a core benchmark for handling authentic sensor noise captured under varying lighting and device conditions. Furthermore, PolyU is utilized to stress-test the model’s generalization capabilities on irregular noise distributions characteristic of low-light and complex environments.

\begin{table}[h]
\caption{Quantitative results of performance comparison on synthetic and natural noise datasets. The metrics are PSNR and SSIM, where higher values indicate better performance.}
\label{tab:denoising}
\centering
\resizebox{\linewidth}{!}{%
\begin{tabular}{l l c c c c c c}
\toprule
Noise Types & Datasets & DnCNN~\cite{dncnn}  & Restormer~\cite{zamir2022restormer}  & MaskDenoising~\cite{maskdenoising}  & HAT~\cite{HAT}  & DIL~\cite{DIL}  & NB pro \\
\midrule
\multirow{3}{*}{\makecell{Gauss \\ $\sigma=50$}} 
& McMaster~\cite{mcmaster} & 20.18/0.312 & 20.47/0.312 & 20.63/0.379 & 20.79/0.364 & 26.61/0.669 & 21.57/0.594 \\
& Kodak24~\cite{Kodak}  & 19.78/0.301 & 20.12/0.321 & 20.72/0.368 & 21.04/0.390 & 27.46/0.736 & 20.04/0.517 \\
& Urban100~\cite{urban} & 19.62/0.420 & 19.36/0.437 & 20.51/0.485 & 20.80/0.492 & 25.89/0.768 & 19.22/0.607 \\
\midrule
\multirow{2}{*}{\makecell{Natural}} 
&SIDD Val~\cite{sidd}  &-/- &-/-  & 33.14/0.913 & 28.58/0.570 & 34.76/0.848 & 26.76/0.681\\
&PolyU~\cite{polyu}  &-/- &-/-& 24.78/0.812 & 37.25/0.948 & 37.65/0.950 & 22.82/0.806\\
\bottomrule
\end{tabular}%
}
\end{table}

As shown in Tab.~\ref{tab:denoising}, Nano Banana Pro exhibits a substantial performance deficit compared to all task-specific baselines. On synthetic datasets, it lags significantly behind the state-of-the-art DIL, with PSNR gaps ranging from 5.04 dB to 7.42 dB and SSIM reductions between 0.075 and 0.219. Notably, this disparity persists regardless of texture complexity (from McMaster to Urban100), indicating a fundamental lack of competitiveness in Gaussian noise removal. This limitation is further exacerbated in real-world blind denoising tasks. On SIDD Val~\cite{sidd}, Nano Banana Pro trails DIL by 8.00 dB in PSNR. The gap widens drastically on PolyU~\cite{polyu}, where it underperforms DIL by 14.83 dB and even falls behind the basic MaskDenoising model. These results underscore an inherent inability of Nano Banana Pro to effectively model and remove complex, realistic noise compared to specialized restoration models.

\begin{figure}[htbp]
    \centering 
    \includegraphics[width=0.9\textwidth]{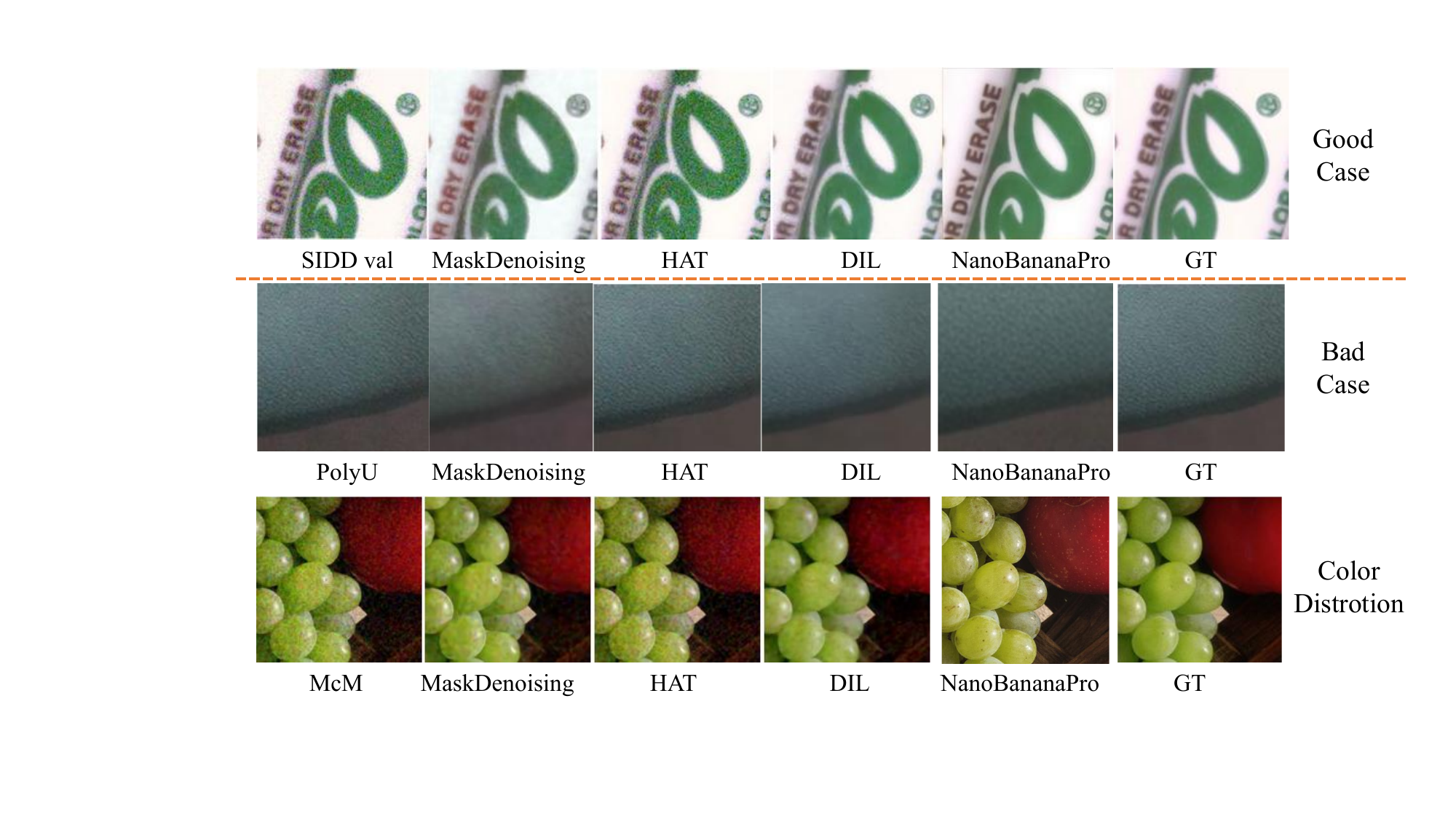} 
    \caption{Visual comparison of denoising results. The first row shows a successful case, where Nano Banana Pro produces text with sharper edges than the Ground Truth, partly due to its generative priors. The last two rows, however, reveal limitations in fidelity: in the second row, surface texture is not preserved, losing high‑frequency details; in the third row, noticeable color distortion shifts the hue of the grapes away from the Ground Truth.}
    \label{fig:denoising} 
    \vspace{-3mm}
\end{figure}

Fig.~\ref{fig:denoising} visually compares Nano Banana Pro against state-of-the-art baselines. The results reveal a distinct characteristic of the generative approach: a trade-off between perceptual clarity and pixel-level fidelity. As shown in the first row, Nano Banana Pro exhibits exceptional perceptual quality on text-rich images. Leveraging its generative priors, it reconstructs the characters with remarkable sharpness. Notably, the output appears even clearer and more legible than the Ground Truth, effectively performing text enhancement alongside denoising. Conversely, the model struggles with consistency in texture and color, as seen in the subsequent rows: In the second row, the model fails to recover the subtle grain of the surface. Instead of preserving the original high-frequency details, it produces an over-smoothed or hallucinated texture that deviates significantly from the Ground Truth. In the third row, the model introduces chromatic deviations. While the noise is removed, the color of the grapes shifts noticeably (appearing brighter and yellower). These cases underscore that while Nano Banana Pro can generate visually pleasing results, it lacks the strict fidelity required for high-precision restoration tasks.

\subsection{Discussion}
Based on the experimental results and architectural characteristics, the suboptimal denoising performance of Nano Banana Pro is attributed to two primary factors:

\noindent \textbf{Misalignment of Task Objectives}: Nano Banana Pro is a general-purpose model optimized for high-level multimodal understanding and generation, rather than low-level pixel-wise restoration. It lacks the specialized architectural biases and targeted loss functions that enable baseline models to effectively balance noise removal with detail preservation.

\noindent \textbf{Trade-off Between Generative Prior and Pixel Fidelity}: As a unified model, Nano Banana Pro prioritizes semantic plausibility and visual coherence over strict pixel-level accuracy. This generative nature often leads to the over-smoothing of high-frequency details in pursuit of ``reasonable'' content, resulting in inferior quantitative metrics compared to task-specific models trained via strict supervision.

In summary, this study evaluated the unified generative model Nano Banana Pro against state-of-the-art specialized models on both synthetic Gaussian noise and real-world blind denoising datasets. Nano Banana Pro significantly underperforms task-specific baselines across all benchmarks, indicating limited competitiveness in direct denoising applications. Direct application of Nano Banana Pro for denoising is not recommended without modification. Enhancing its utility requires targeted adaptations such as prompt engineering, parameter fine-tuning, or integration with post-processing modules. Future research should explore methodologies to align general-purpose generative priors with low-level processing demands.
\section{Reflection Removal}
    \subsection{Background}
In fields such as computer vision, clear and interference-free image data is a fundamental foundation for subsequent analytical tasks including object detection and semantic segmentation. However, reflective surfaces like glass, water, and metal easily reflect ambient light into images, forming an interfering reflection layer over the real scene. This causes blurred details and obscured target information, directly compromising the reliability and accuracy of subsequent tasks. Single-Image Reflection Removal (SIRR), as a core technical solution, aims to accurately separate the transmission layer (real scene) from the reflection layer (interfering component) in a single mixed image to restore the true scene. It holds irreplaceable practical value in autonomous driving, security monitoring, consumer electronics, and other areas.

SIRR is inherently a typical ill-posed inverse problem—without additional constraints, mixed image decomposition has infinitely many solutions. Early traditional methods relied on manually designed prior knowledge (sparsity, smoothness, and other assumptions) and linear modeling (such as $I=T+R$) to simplify the problem, but real-world reflections exhibit complex nonlinear characteristics due to light intensity, shooting angle, surface material, and other factors, leading to limited generalization of these methods. In recent years, deep learning has become the mainstream in SIRR research, forming three core architectures: single-stage approaches that directly output the target layer via a single network \cite{weiSingleImageReflection2019, zhangSingleImageReflection2018}, two-stage approaches that perform intermediate feature estimation followed by refinement \cite{liTwostageSingleImage2023, dongLocationawareSingleImage2021}, and multi-stage approaches that achieve reflection removal through recurrent cascaded iterative optimization \cite{liSingleImageReflection2020, yangSeeingDeeplyBidirectionally2018}. Notably, the rapid development of generative artificial intelligence has injected new vitality into the field, with methods based on diffusion models and Transformers demonstrating potential to break through traditional limitations \cite{hongLDiffERSingleImage2025, zakarinReflectionRemovalEfficient2025}. Nevertheless, existing approaches face significant bottlenecks: the scarcity of high-quality annotated datasets restricts model generalization, and issues such as scene information loss from strong reflections and overlapping appearance distributions between transmission and reflection layers make it hard to balance thorough reflection removal and detail preservation.

Existing research covers traditional methods, deep learning architectures, and generative paradigms, but there remains substantial room for improvement in robustness and detail fidelity under complex real-world scenarios. On one hand, while generative models show promise, their hallucination suppression capabilities and adaptability to complex reflection mechanisms in high-fidelity tasks like SIRR have not been fully verified. On the other hand, efficient solutions for diverse reflection scenarios (diverse material surfaces, extreme lighting, and other scenarios) are lacking, demanding more generalizable generative models. Based on this, this report focuses on the latest generative model Nano Banana Pro. By systematically comparing it with existing baselines using quantitative and qualitative metrics, we investigate its detail restoration, anti-interference performance, and generalization in real reflection removal scenarios. The goal is to reveal its core advantages and limitations, providing practical references for technical optimization and model design in the SIRR field.

\subsection{Quantitative Results}

To evaluate the performance of Google's Nano Banana Pro model on the SIRR task, we conducted experiments using the model as an off-the-shelf solution via API calls. Given the closed-source nature of the model which precludes task-specific fine-tuning, we adopted a direct inference strategy. Only the raw reflection-contaminated images and task-specific prompts were provided as input, without introducing any additional priors or auxiliary guidance. It is worth noting that the resolution of images generated by Nano Banana Pro is fixed at a scale of approximately 1024 pixels, which differs from the original input dimensions. To ensure the fairness and accuracy of the quantitative evaluation, all output images were resized to match the original resolution of the corresponding ground-truth images before metric calculation.

For a comprehensive assessment, we adopted three mainstream datasets in the SIRR domain as our evaluation benchmark. Specifically, we utilized Real20~\cite{zhangSingleImageReflection2018}, which contains 20 images from real-world glass reflection scenes; Nature~\cite{liSingleImageReflection2020}, consisting of 20 samples focusing on outdoor natural landscapes; and SIR$^2$~\cite{wanBenchmarkingSingleImageReflection2017}, where we evaluated on its Objects, Postcard, and Wild subsets. We compared Nano Banana Pro against 15 state-of-the-art baseline models, including ERRNet~\cite{weiSingleImageReflection2019}, IBCLN~\cite{liSingleImageReflection2020}, YTMT~\cite{huTrashTreasureInteractive2021}, Dong et al.~\cite{dongLocationawareSingleImage2021}, DSRNet~\cite{huSingleImageReflection2023}, RAGNet~\cite{liTwostageSingleImage2023}, RRW~\cite{zhuRevisitingSingleImage2024}, DSIT~\cite{guoSingleImageReflection2024}, RDNet~\cite{zhaoReversibleDecouplingNetwork2025}, F2T2-HiT~\cite{caiF2T2HiTUShapedFFT2025}, Huang et al.~\cite{huangSingleImageReflection2025}, L-DiffER~\cite{hongLDiffERSingleImage2025}, DAI~\cite{huDereflectionAnyImage2025}, Lu et al.~\cite{luSingleimageReflectionRemoval2025}, and WindowSeat~\cite{zakarinReflectionRemovalEfficient2025}.

The evaluation metrics assess both basic image quality and perceptual quality. For pixel-level fidelity, we employed Peak Signal-to-Noise Ratio (PSNR) and Structural Similarity Index (SSIM), with comprehensive comparison results summarized in Tab.~\ref{tab:sirr_quantitative_simplified}. To better assess visual perception, we further incorporated Multi-Scale Structural Similarity (MS-SSIM) and Learned Perceptual Image Patch Similarity (LPIPS). Note that lower LPIPS values indicate better perceptual quality. For these perceptual metrics, we selected recent SOTA methods for comparison (with baseline data sourced from WindowSeat~\cite{zakarinReflectionRemovalEfficient2025}), as presented in Tab.~\ref{tab:sirr_perceptual}. All metrics were calculated on the RGB channels between the resized output and the ground truth.

\begin{table*}[ht]
  \centering
  \caption{Quantitative Comparison of Single-Image Reflection Removal Methods. The best and second-best results are highlighted by \textbf{black bold} and \underline{underline}, respectively. $\dag$: Training data includes Nature dataset; $\star$: Generative AI-based method. Note: $\uparrow$ indicates higher is better. SIR² dataset is divided into Objects, Postcard and Wild subsets; SIR$^2$(454) denotes the commonly used subset, while SIR$^2$(500) denotes the full public dataset.}
  \normalsize
  \setlength{\heavyrulewidth}{1.2pt}
  \setlength{\lightrulewidth}{1pt}  
  \renewcommand{\arraystretch}{1.2}
  \resizebox{\textwidth}{!}{
  \begin{tabular}{lccccccccccccccc}
    \toprule
    \multirow{2}{*}{\textbf{Method}} & \multirow{2}{*}{\textbf{Year}} & 
    \multicolumn{2}{c}{\textbf{Real20 (20)}} & \multicolumn{2}{c}{\textbf{Nature (20)}} & \multicolumn{2}{c}{\textbf{Objects (200)}} & 
    \multicolumn{2}{c}{\textbf{Postcard (199)}} & \multicolumn{2}{c}{\textbf{Wild (55)}} & \multicolumn{2}{c}{\textbf{SIR$^2$ (454)}} & \multicolumn{2}{c}{\textbf{SIR$^2$ (500)}}\\
    \cmidrule(lr){3-4} \cmidrule(lr){5-6} \cmidrule(lr){7-8} \cmidrule(lr){9-10} \cmidrule(lr){11-12} \cmidrule(lr){13-14} \cmidrule(lr){15-16}
    & & PSNR$\uparrow$ & SSIM$\uparrow$ & PSNR$\uparrow$ & SSIM$\uparrow$ & PSNR$\uparrow$ & SSIM$\uparrow$ & PSNR$\uparrow$ & SSIM$\uparrow$ & PSNR$\uparrow$ & SSIM$\uparrow$ & PSNR$\uparrow$ & SSIM$\uparrow$ & PSNR$\uparrow$ & SSIM$\uparrow$ \\
    \midrule
    ERRNet~\cite{weiSingleImageReflection2019} & 2019 & 22.89 & 0.803 & - & - & 24.87 & 0.896 & 22.04 & 0.876 & 24.25 & 0.853 & 23.55 & 0.882 & - & - \\
    IBCLN~\cite{liSingleImageReflection2020} & 2020 & 21.86 & 0.762 & 23.57 & 0.783 & 24.87 & 0.893 & 23.39 & 0.875 & 24.71 & 0.886 & 24.20 & 0.884 & - & - \\
    YTMT~\cite{huTrashTreasureInteractive2021} & 2021 & 23.26 & 0.806 & - & - & 24.87 & 0.896 & 22.91 & 0.884 & 25.48 & 0.890 & 24.08 & 0.890 & - & - \\
    Dong et al.$^\dag$~\cite{dongLocationawareSingleImage2021} & 2021 & 23.34 & 0.812 & 23.45 & 0.808 & 24.36 & 0.898 & 23.72 & 0.903 & 25.73 & 0.902 & 24.25 & 0.901 & - & - \\
    DSRNet (w/o extra)~\cite{huSingleImageReflection2023} & 2023 & 24.23 & 0.820 & - & - & 26.28 & 0.914 & 24.56 & 0.908 & 25.68 & 0.896 & 25.45 & 0.909 & - & - \\
    DSRNet (with extra)~\cite{huSingleImageReflection2023} & 2023 & 23.91 & 0.818 & - & - & 26.74 & 0.920 & 24.83 & 0.911 & 26.11 & 0.906 & 25.83 & 0.914 & - & - \\
    RAGNet~\cite{liTwostageSingleImage2023} & 2023 & 22.95 & 0.793 & - & - & 26.15 & 0.903 & 23.67 & 0.879 & 25.52 & 0.880 & 24.99 & 0.890 & - & - \\
    RRW~\cite{zhuRevisitingSingleImage2024} & 2024 & 23.82 & 0.817 & 25.96 & 0.843 & - & - & - & - & - & - & 25.45 & 0.910 & - & - \\
    DSIT (data I)~\cite{guoSingleImageReflection2024} & 2024 & 25.06 & 0.836 & - & - & 26.81 & 0.919 & 25.63 & 0.924 & 27.06 & 0.910 & 26.32 & 0.920 & - & - \\
    DSIT (data II)~\cite{guoSingleImageReflection2024} & 2024 & 25.22 & 0.836 & - & - & 27.27 & \textbf{0.932} & 25.58 & 0.922 & 27.40 & 0.918 & 26.54 & 0.926 & - & - \\
    RDNet (w/o nature)~\cite{zhaoReversibleDecouplingNetwork2025} & 2025 & 24.43 & 0.835 & - & - & 25.76 & 0.905 & 25.95 & 0.920 & 27.20 & 0.910 & 26.02 & 0.912 & - & - \\
    RDNet (w nature)~\cite{zhaoReversibleDecouplingNetwork2025} & 2025 & 25.58 & 0.846 & - & - & 26.78 & 0.921 & 26.33 & 0.922 & 27.70 & 0.915 & 26.69 & 0.921 & - & - \\
    F2T2-HiT~\cite{caiF2T2HiTUShapedFFT2025} & 2025 & 21.64 & 0.766 & 26.08 & 0.837 & - & - & - & - & - & - & 25.72 & 0.903 & - & - \\
    Huang et al.~\cite{huangSingleImageReflection2025} & 2025 & 25.12 & 0.828 & 27.03 & 0.853 & 27.07 & 0.930 & 26.43 & 0.931 & 27.96 & \textbf{0.922} & 26.90 & 0.929 & - & - \\
    L-DiffER$^\star$~\cite{hongLDiffERSingleImage2025} & 2025 & 23.77 & 0.821 & 23.95 & 0.831 & - & - & - & - & - & - & - & - & 25.18 & 0.911 \\
    DAI$^\star$~\cite{huDereflectionAnyImage2025} & 2025 & 25.24 & 0.840 & 27.05 & 0.846 & - & - & - & - & - & - & - & - & 27.32 & 0.931 \\
    Lu et al.$^\star$~\cite{luSingleimageReflectionRemoval2025} & 2025 & - & - & - & - & - & - & - & - & - & - & - & - & 28.41 & 0.912 \\
    WindowSeat$^\star$~\cite{zakarinReflectionRemovalEfficient2025} & 2025 & 26.28 & 0.856 & 27.12 & 0.849 & 28.81 & 0.944 & 29.17 & 0.934 & 28.97 & 0.935 & 28.99 & 0.939 & 28.75 & \textbf{0.940} \\
    WindowSeat$^\star$ (Qwen-IE)~\cite{zakarinReflectionRemovalEfficient2025} & 2025 & \textbf{26.60} & \textbf{0.864} & \textbf{27.57} & \textbf{0.855} & \textbf{28.85} & \underline{0.938} & \underline{28.70} & \underline{0.933} & \textbf{29.44} & \textbf{0.936} & \underline{28.84} & \underline{0.936} & \underline{28.60} & \underline{0.937} \\
    \rowcolor{lightgray!50}
\textbf{\textcolor{orange}{Nano Banana Pro}} & \textbf{\textcolor{orange}{2025}} & \textbf{\textcolor{orange}{20.26}} & \textbf{\textcolor{orange}{0.655}} & \textbf{\textcolor{orange}{21.48}} & \textbf{\textcolor{orange}{0.723}} & \textbf{\textcolor{orange}{21.95}} & \textbf{\textcolor{orange}{0.751}} & \textbf{\textcolor{orange}{19.29 }}& \textbf{\textcolor{orange}{0.675 }}& \textbf{\textcolor{orange}{23.58 }}& \textbf{\textcolor{orange}{0.798 }}& \textbf{\textcolor{orange}{20.98 }}& \textbf{\textcolor{orange}{0.724 }}& \textbf{\textcolor{orange}{21.11 }}& \textbf{\textcolor{orange}{0.730 }}\\
    \bottomrule
  \end{tabular}
  }
  \label{tab:sirr_quantitative_simplified}
\end{table*}

As shown in Tab.~\ref{tab:sirr_quantitative_simplified}, Nano Banana Pro exhibits a notable performance gap compared to state-of-the-art specialist methods. Quantitatively, it lags behind across all datasets in pixel-wise metrics (PSNR/SSIM). This disparity largely stems from the fundamental difference in optimization objectives: regression-based SOTA methods are supervised to minimize pixel-level reconstruction error, ensuring precise alignment. In contrast, the generative approach of Nano Banana Pro prioritizes semantic coherence over structural fidelity, often resulting in global intensity scaling and spatial shifts that heavily penalize PSNR, even if the image content is semantically correct.

\begin{table*}[ht]
  \centering
  \caption{Perceptual Quality Comparison (MS-SSIM and LPIPS) on Mainstream Datasets. $\uparrow$ indicates higher is better, while $\downarrow$ indicates lower is better. The best and second-best results are highlighted in \textbf{black bold} and \underline{underline}. Baseline results are sourced from WindowSeat~\cite{zakarinReflectionRemovalEfficient2025}.}
  \normalsize
  \setlength{\heavyrulewidth}{1.2pt}
  \setlength{\lightrulewidth}{1pt}  
  \renewcommand{\arraystretch}{1.2}
  \resizebox{\textwidth}{!}{
  \begin{tabular}{lcccccccccc}
    \toprule
    \multirow{2}{*}{\textbf{Method}} & 
    \multicolumn{2}{c}{\textbf{Real20 (20)}} & \multicolumn{2}{c}{\textbf{Nature (20)}} & \multicolumn{2}{c}{\textbf{Objects (200)}} & 
    \multicolumn{2}{c}{\textbf{Postcard (199)}} & \multicolumn{2}{c}{\textbf{Wild (55)}} \\
    \cmidrule(lr){2-3} \cmidrule(lr){4-5} \cmidrule(lr){6-7} \cmidrule(lr){8-9} \cmidrule(lr){10-11}
    & MS-SSIM$\uparrow$ & LPIPS$\downarrow$ & MS-SSIM$\uparrow$ & LPIPS$\downarrow$ & MS-SSIM$\uparrow$ & LPIPS$\downarrow$ & MS-SSIM$\uparrow$ & LPIPS$\downarrow$ & MS-SSIM$\uparrow$ & LPIPS$\downarrow$ \\
    \midrule
    DSRNet~\cite{huSingleImageReflection2023} & 0.8737 & 0.1831 & 0.9144 & 0.1478 & 0.9564 & 0.0847 & 0.9263 & 0.1260 & 0.9338 & 0.1096 \\
    DAI~\cite{huDereflectionAnyImage2025} & 0.9045 & 0.1790 & 0.9309 & 0.2161 & 0.9638 & 0.0689 & 0.9567 & 0.1029 & 0.9423 & 0.0941 \\
    RDNet~\cite{zhaoReversibleDecouplingNetwork2025} & 0.9081 & 0.1442 & 0.9231 & \underline{0.1361} & 0.9609 & 0.0836 & 0.9361 & 0.1121 & 0.9406 & 0.0992 \\
    DSIT~\cite{guoSingleImageReflection2024} & 0.8934 & 0.1618 & 0.9223 & 0.1598 & 0.9586 & 0.0939 & 0.9441 & 0.1242 & 0.9447 & 0.0967 \\
    WindowSeat~\cite{zakarinReflectionRemovalEfficient2025} & \underline{0.9296} & \underline{0.1131} & \underline{0.9435} & 0.1368 & \textbf{0.9759} & \textbf{0.0470} & \textbf{0.9693} & \textbf{0.0504} & \underline{0.9625} & \textbf{0.0632} \\
    WindowSeat (Qwen-IE)~\cite{zakarinReflectionRemovalEfficient2025} & \textbf{0.9396} & \textbf{0.1074} & \textbf{0.9494} & \textbf{0.1355} & \underline{0.9661} & \underline{0.0550} & \underline{0.9664} & \underline{0.0549} & \textbf{0.9655} & \underline{0.0682} \\
    \rowcolor{lightgray!50}
    \textbf{\textcolor{orange}{Nano Banana Pro}} & \textbf{\textcolor{orange}{0.8013}} & \textbf{\textcolor{orange}{0.2411}} & \textbf{\textcolor{orange}{0.8580}} & \textbf{\textcolor{orange}{0.1851}} & \textbf{\textcolor{orange}{0.8861}} & \textbf{\textcolor{orange}{0.1552}} & \textbf{\textcolor{orange}{0.8373}} & \textbf{\textcolor{orange}{0.2513}} & \textbf{\textcolor{orange}{0.8874}} & \textbf{\textcolor{orange}{0.1578}} \\
    \bottomrule
  \end{tabular}
  }
  \label{tab:sirr_perceptual}
\end{table*}

Tab.~\ref{tab:sirr_perceptual} further illustrates the performance in terms of perceptual quality metrics. Despite the generative nature of Nano Banana Pro, which typically favors perceptual scores, it still exhibits high LPIPS values (e.g., 0.2513 on Postcard vs. 0.0549 for SOTA). Unlike PSNR, which penalizes misalignment, the poor LPIPS performance points to a deeper issue: \textit{semantic and stylistic deviation}. The model tends to perform aggressive "image-to-image translation" rather than faithful restoration, altering fundamental scene characteristics—such as modifying illumination, hallucinating textures, or shifting the color domain—thereby drifting away from the ground truth's perceptual manifold.

From an interpretive perspective, the elevated LPIPS and sub-optimal MS-SSIM scores highlight the limitations of general-purpose generative models in high-fidelity restoration tasks. Although the model possesses strong generative capabilities, it lacks a precise mechanism for decoupling reflection layers from the background. The high LPIPS values suggest a substantial deviation in color distribution, texture details, and high-level semantic features relative to the ground truth. This deviation likely stems from the model partially merging residual reflections into the background or altering the original color and complex textures during the regeneration process. Consequently, while the generated images may appear visually natural, they fail to meet the strict fidelity requirements essential for reflection removal tasks.

\subsection{Qualitative Results}

\begin{figure*}[ht!]  
  \centering  
  \includegraphics[width=\textwidth]{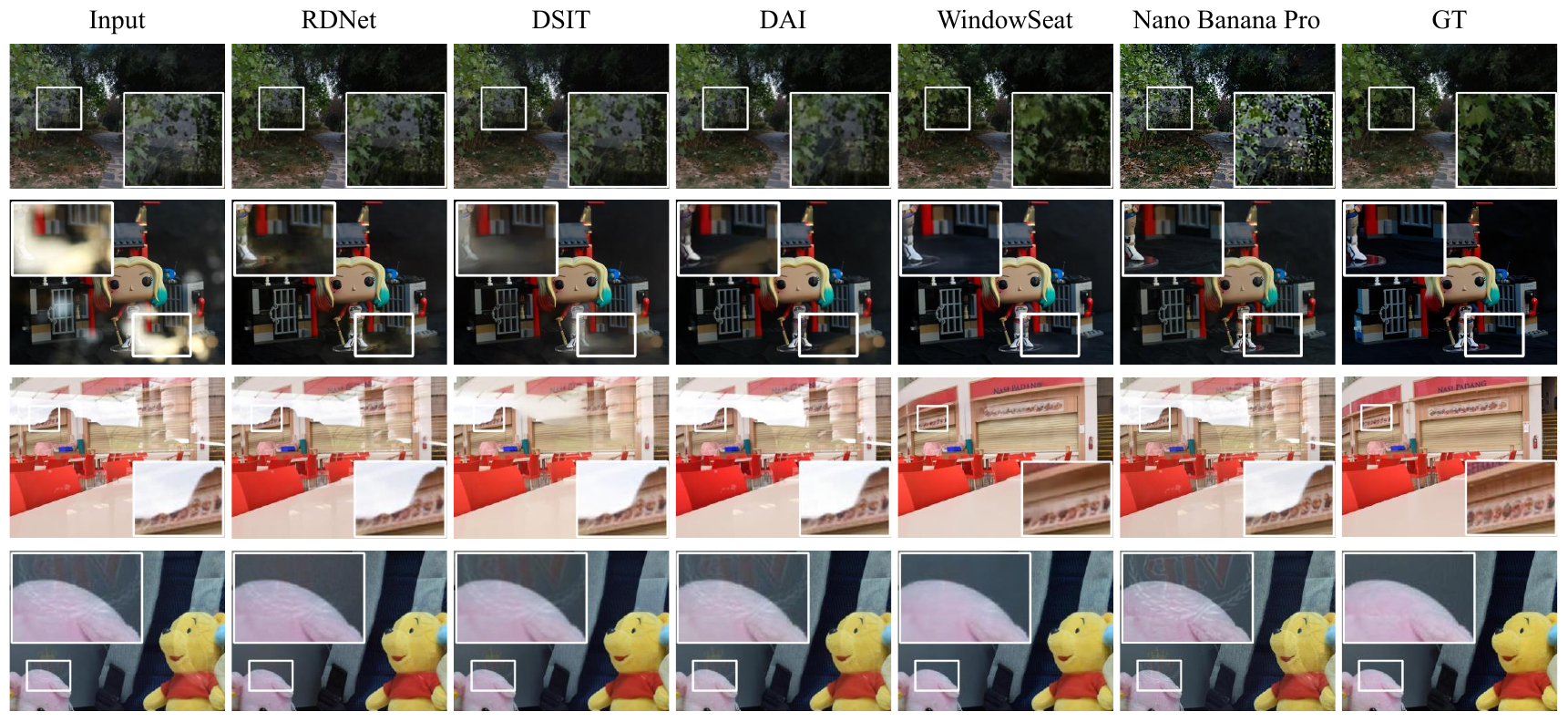}  
\caption[\captionfont Qualitative comparison of reflection removal results]{Qualitative comparison of reflection removal results. Each row shows the input image, predictions from state-of-the-art methods, Nano Banana Pro, and the ground truth (GT) for a single sample. Except for the results of Nano Banana Pro, all other method results are sourced from WindowSeat~\cite{zakarinReflectionRemovalEfficient2025}.}
\label{fig:qualitative_comparison}
\end{figure*}

\begin{figure*}[ht!]  
  \centering  
  \includegraphics[width=\textwidth]{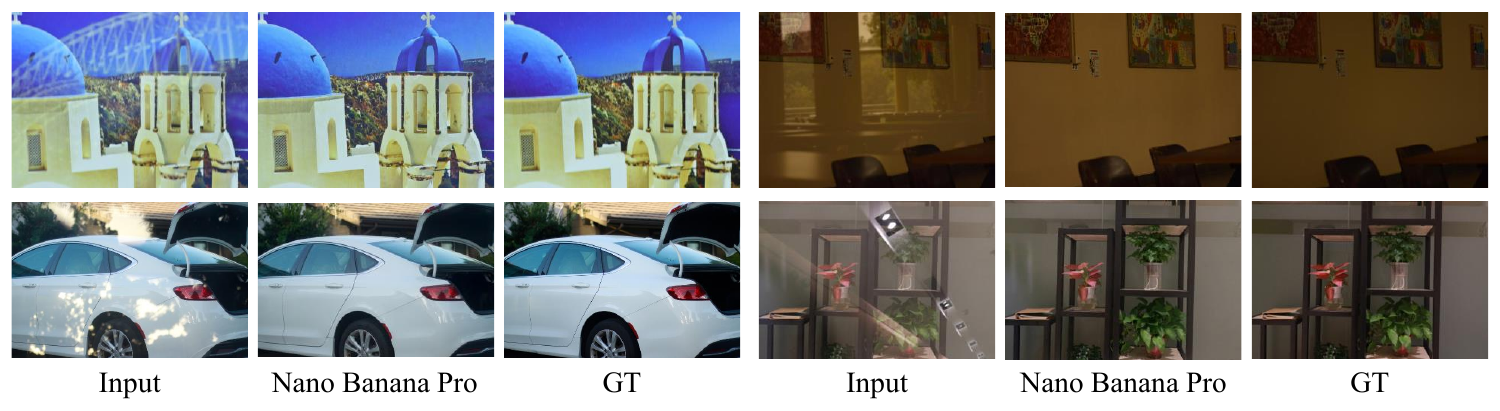}  
\caption{Visual comparison of selected samples. We present the input images, the restoration results generated by Nano Banana Pro, and the corresponding ground truth. These examples illustrate the visual performance of the method in recovering background content across different scenes.}
\label{fig:high-score-results}
\end{figure*}

\begin{figure*}[ht!]  
  \centering  
  \includegraphics[width=\textwidth]{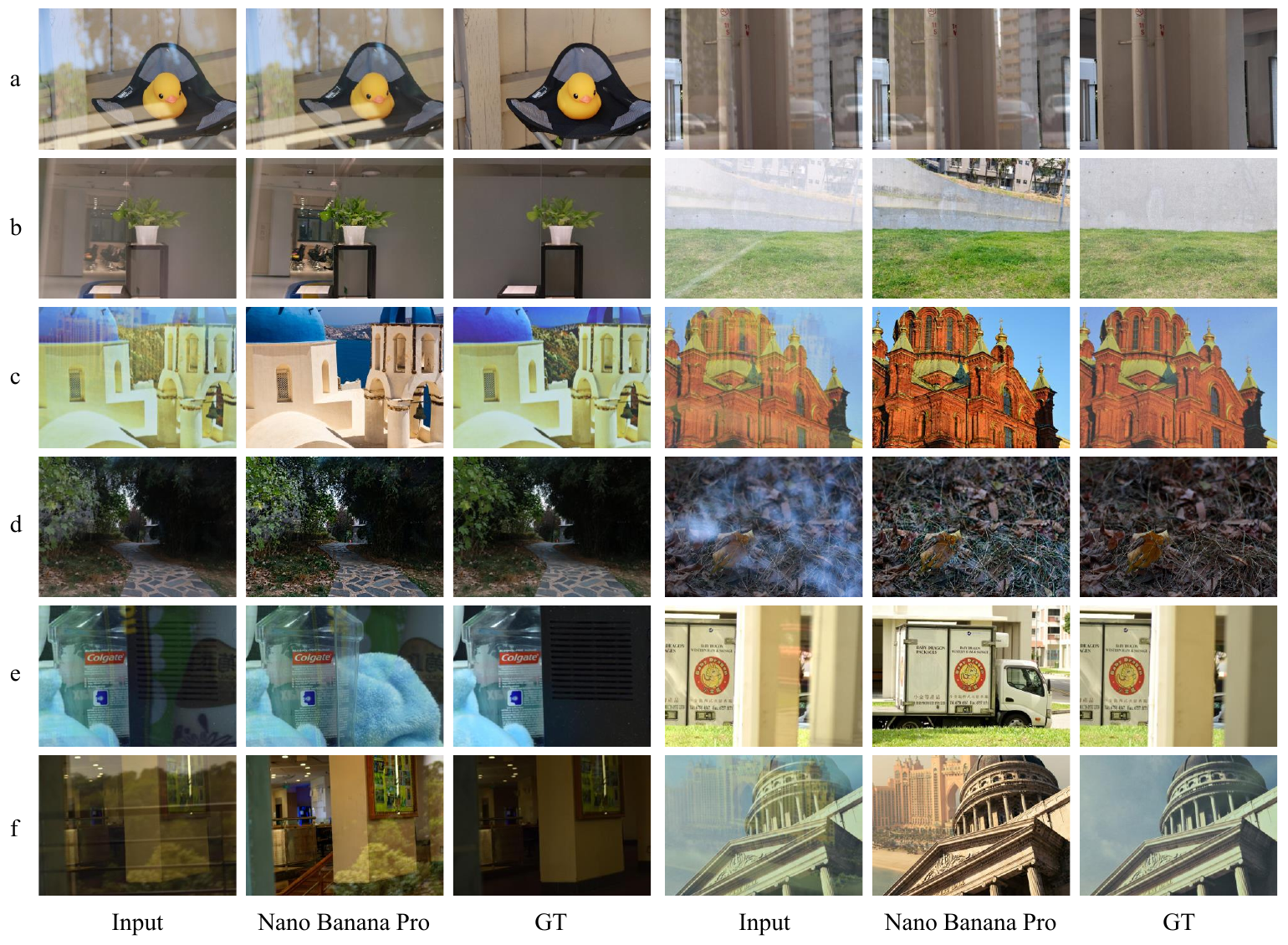}
\caption{Qualitative analysis of limitations and failure cases. We present typical examples where the proposed method yields suboptimal results, categorized by specific degradation types: (a) incomplete reflection removal due to strong intensity; (b) erroneous enhancement of reflection artifacts mistaken for background details; (c) unintended color deviation in the transmission layer; (d) significant texture distortion; (e) structural deformation compared to the ground truth; and (f) complex scenarios exhibiting compound artifacts involving multiple aforementioned issues.}
\label{fig:low-score-results}
\end{figure*}

In this section, we conduct a comprehensive qualitative evaluation of the proposed Nano Banana Pro. To establish a comparative baseline, we first benchmark our visual results against existing state-of-the-art methods in Fig.~\ref{fig:qualitative_comparison}. Subsequently, we examine the specific restoration capabilities of our model through selected samples in Fig.~\ref{fig:high-score-results}. Finally, to ensure a balanced assessment and facilitate future improvements, we provide an analysis of the model's limitations by categorizing typical failure cases and degradation patterns in Fig.~\ref{fig:low-score-results}.

As illustrated in Fig.~\ref{fig:qualitative_comparison}, the proposed Nano Banana Pro exhibits a high variance in performance compared to state-of-the-art methods. In specific instances, our method outperforms existing approaches, yielding results that are visually comparable from the ground truth. However, generally, it lacks the stability of specialized regression-based models. Notably, the model struggles with preserving high-frequency details, leading to the loss of complex textures (e.g., row 1), or suffers from semantic ambiguity where reflection artifacts are erroneously interpreted as background elements and subsequently enhanced (e.g., row 4). These limitations contribute to a lower average performance despite the high perceptual quality in successful cases.

Fig.~\ref{fig:high-score-results} showcases selected samples where Nano Banana Pro demonstrates superior restoration capabilities. It is observed that when there is a significant semantic or visual distinction between the reflection and transmission layers, the model effectively suppresses the reflection while preserving background integrity. The results indicate that the model possesses a high performance upper bound, occasionally achieving reconstruction quality nearly identical to the ground truth. We attribute this potential to the robust generative priors acquired from large-scale pre-training. However, the absence of domain-specific supervision for reflection separation implies a trade-off: without explicit guidance, the model may misapply these priors, failing to disentangle the layers or introducing generative hallucinations and noise into the transmission layer. Experimental results suggest that such misuse of priors accounts for a considerable portion of the suboptimal outputs.

Given the quantitative gap observed in the previous section, explicitly analyzing the failure modes provides critical insights into the misbehavior of generative priors. Based on the distinct characteristics of the introduced artifacts and degradation mechanisms, the suboptimal performance of Nano Banana Pro can be systematically categorized into six types, as visualized in Fig.~\ref{fig:low-score-results}.

\textbf{(a) Incomplete Reflection Removal.} In these instances, the model fails to effectively decouple the reflection layer from the transmission layer, resulting in significant residual artifacts. This behavior likely stems from a conservative inference strategy induced by prompts emphasizing background preservation. When the reflection intensity is high or statistically similar to the background, the model tends to classify the reflection as intrinsic scene content to avoid over-erasing potential background details.

\textbf{(b) Erroneous Enhancement due to Semantic Ambiguity.} Despite explicit instructions to suppress reflections, the model occasionally misinterprets reflection artifacts as valid background elements and erroneously enhances them. This phenomenon highlights a limitation in current generative priors: the model is driven by semantic plausibility rather than physical layer separation. When a reflection (e.g., a light source or architectural reflection) aligns semantically with the background scene, the model prioritizes generating a "coherent" image without contradictions, thereby integrating the artifact as a strengthened feature.

\textbf{(c) Unintended Chromatic and Domain Shift.} This error is predominantly observed in the \textit{Postcard} dataset, which features images of paintings or prints (e.g., urban or humanist subjects). The model struggles with domain ambiguity, failing to distinguish between a "picture of a scene" and a "real-world scene." Consequently, it attempts to "restore" the printed content as a realistic photograph, aggressively altering saturation, removing characteristic grain, or modifying illumination. This results in severe color deviations and stylistic inconsistencies compared to the ground truth.

\textbf{(d) Texture Fidelity Loss.} In scenarios containing high-frequency details, such as natural foliage or fabric textures (e.g., towels), the generative process often fails to maintain the original texture distribution. The output tends to exhibit either unnatural over-smoothing (loss of fine grain) or artificial sharpening (introduction of high-frequency noise), indicating a lack of fine-grained control in the texture reconstruction module.

\textbf{(e) Structural Hallucination and Deformation.} In rare but severe cases, the model breaks structural consistency, generating outputs that deviate geometrically from the input. This includes the hallucination of non-existent background structures or the removal of actual objects mistaken for reflections. Such failures represent a collapse of the conditioning mechanism, where the strong generative prior overrides the spatial constraints provided by the input image.

\textbf{(f) Compound Degradation.} A significant portion of low-scoring results exhibits a hybrid of the aforementioned failure modes. For example, an image may suffer from incomplete reflection removal while simultaneously undergoing a global color shift, or experience structural deformation alongside texture smoothing. These complex scenarios represent the most challenging cases for the current architecture.
\section{Flare Removal}

\subsection{Background}
Lens flare constitutes a fundamental optical phenomenon wherein intense incident light undergoes scattering and reflection within a camera's lens system, resulting in parasitic artifacts that degrade image quality. These artifacts manifest predominantly as two distinct categories: scattering flares and reflective flares. Beyond aesthetic degradation, these artifacts critically impair downstream computer vision applications, including stereo matching misestimation, optical flow corruption, and semantic segmentation misclassification, thereby posing substantial risks to safety-critical systems such as autonomous driving and aerial object tracking~\cite{Hullin2011FlareRendering, Li2021CARMC, Wu2021NN4FR}.

Contemporary flare removal methodologies have evolved from traditional detection-based approaches to sophisticated deep learning frameworks enabled by large-scale datasets. The Flare7K++~\cite{Dai2024Flare7Kpp} dataset represents a pivotal advancement, providing 7,000 synthetic flares with 25 scattering and 10 reflective patterns, supplemented by 962 real-captured flare images (Flare-R) that capture complex degradation effects unattainable through simulation alone. Recent state-of-the-art approaches demonstrate remarkable progress: DeflareMamba~\cite{Huang2025DeflareMamba} introduces the first State Space Model (SSM)-based architecture, employing a hierarchical U-shaped framework with local-enhanced selective scan mechanisms to maintain contextual consistency across global flare patterns and local scene details. Meanwhile, the MiAlgo AI team achieved top performance in the MIPI 2024 Nighttime Flare Removal Challenge through a Progressive Perception Diffusion Network (PPDN)~\cite{Dai2024MIPI}, combining an IR-SDE diffusion module for comprehensive flare elimination with an AOT Block enhancement stage for detail recovery, employing a two-stage progressive strategy to improve visual quality.

Performance assessment is conducted on two complementary benchmark suites. The Flare7K++~\cite{Dai2024Flare7Kpp} test set comprises 100 meticulously aligned 512×512-resolution real-world flare-corrupted/flare-free image pairs, with manual annotations delineating glare, streak, and light source regions to enable component-specific evaluation via G-PSNR and S-PSNR metrics. Additionally, the MIPI 2024 Challenge introduces FlareReal600~\cite{Dai2024MIPI}, a high-resolution dataset featuring 600 aligned training images, with validation and test sets each containing 50 pairs available in both 2K (1440×1920) and 4K (1774×3840) resolutions to facilitate comprehensive evaluation across different spatial scales. In this subsection, we will systematically evaluate the flare removal capability of the Nano Banana Pro model on these benchmarks. We will examine its effectiveness in eliminating various nighttime flare artifacts while maintaining the scene's semantic and photometric integrity across different image resolutions, thereby providing a reference for the community.

\begin{table*}[ht]
  \centering
  \caption{Quantitative comparisons of Nano Banana Pro and representative specialists on the Flare7K++ dataset.}
  \normalsize
  \setlength{\heavyrulewidth}{1.2pt}
  \setlength{\lightrulewidth}{1pt}  
  \renewcommand{\arraystretch}{1.15}
  \begin{tabular}{l|cccccc}
    \toprule
    \textbf{Metric} & Input & Restormer~\cite{zamir2022restormer} & Uformer~\cite{wang2022uformer} & Flare-level~\cite{deng2024towards} & DeflareMamba~\cite{Huang2025DeflareMamba} & NB Pro \\
    \midrule
    PSNR\textuparrow & 22.56 & 27.60 & 27.63 & 27.05  & 26.06 & 24.92 \\
    SSIM\textuparrow & 0.857 & 0.897 & 0.894 & 0.901 & 0.898 & 0.844 \\
    \bottomrule
  \end{tabular}
  \label{flare_table1}
\end{table*}

\begin{table*}[ht]
  \centering
  \caption{Quantitative comparisons of Nano Banana Pro and representative specialists on the FlareReal600 dataset.}
  \normalsize
  \setlength{\heavyrulewidth}{1.2pt}
  \setlength{\lightrulewidth}{1pt}  
  \renewcommand{\arraystretch}{1.15}
  \begin{tabular}{l|ccc}
    \toprule
    \textbf{Metric} & PPDN~\cite{Dai2024MIPI} & NB Pro(2K) & NB Pro(4K) \\
    \midrule
    LPIPS~\cite{zhang2018unreasonable}\textdownarrow  & 0.143 & 0.287 & 0.361 \\
    PSNR\textuparrow  & 22.15 & 19.07 & 18.32 \\
    SSIM~\cite{wang2004image}\textuparrow  & 0.708 & 0.496 & 0.424 \\
    \bottomrule
  \end{tabular}
  \label{flare_table2}
\end{table*}

\subsection{Qualitative and Quantitative Results}

To comprehensively assess the capabilities of Nano Banana Pro, we organized our experiments into quantitative evaluation and qualitative analysis.

\textbf{Quantitative Evaluation.}
We first evaluated the model on the Flare7K++ dataset configured to an output resolution of 1K. Performance was measured using Peak Signal-to-Noise Ratio (PSNR) and Structural Similarity Index Measure (SSIM)~\cite{wang2004image}. As shown in Tab.~\ref{flare_table1}, we compared Nano Banana Pro against state-of-the-art methods trained on the same dataset, including Restormer~\cite{zamir2022restormer}, Uformer~\cite{wang2022uformer}, and DeflareMamba~\cite{Huang2025DeflareMamba}. Subsequently, we extended our evaluation to the FlareReal600 dataset, processing images at their native resolutions to output 2K and 4K results. In addition to PSNR and SSIM, Learned Perceptual Image Patch Similarity (LPIPS)~\cite{zhang2018unreasonable} was included to assess perceptual quality. Tab.~\ref{flare_table2} benchmarks our results against the MIPI 2024 Challenge champion, MiAlgo AI. Note that while the challenge metrics were derived from an unpublished test set, our evaluation utilized the publicly available validation set. We observed two notable quantitative trends: 1) Resolution Impact: Performance metrics generally decline as output resolution increases. 2) Brightness Sensitivity: On the high-resolution FlareReal600 dataset, higher image brightness results in degraded metrics. However, this trend is not evident in the lower-resolution Flare7K++ dataset.

\begin{figure}[ht]  
  \centering  
  \includegraphics[width=0.8\textwidth]{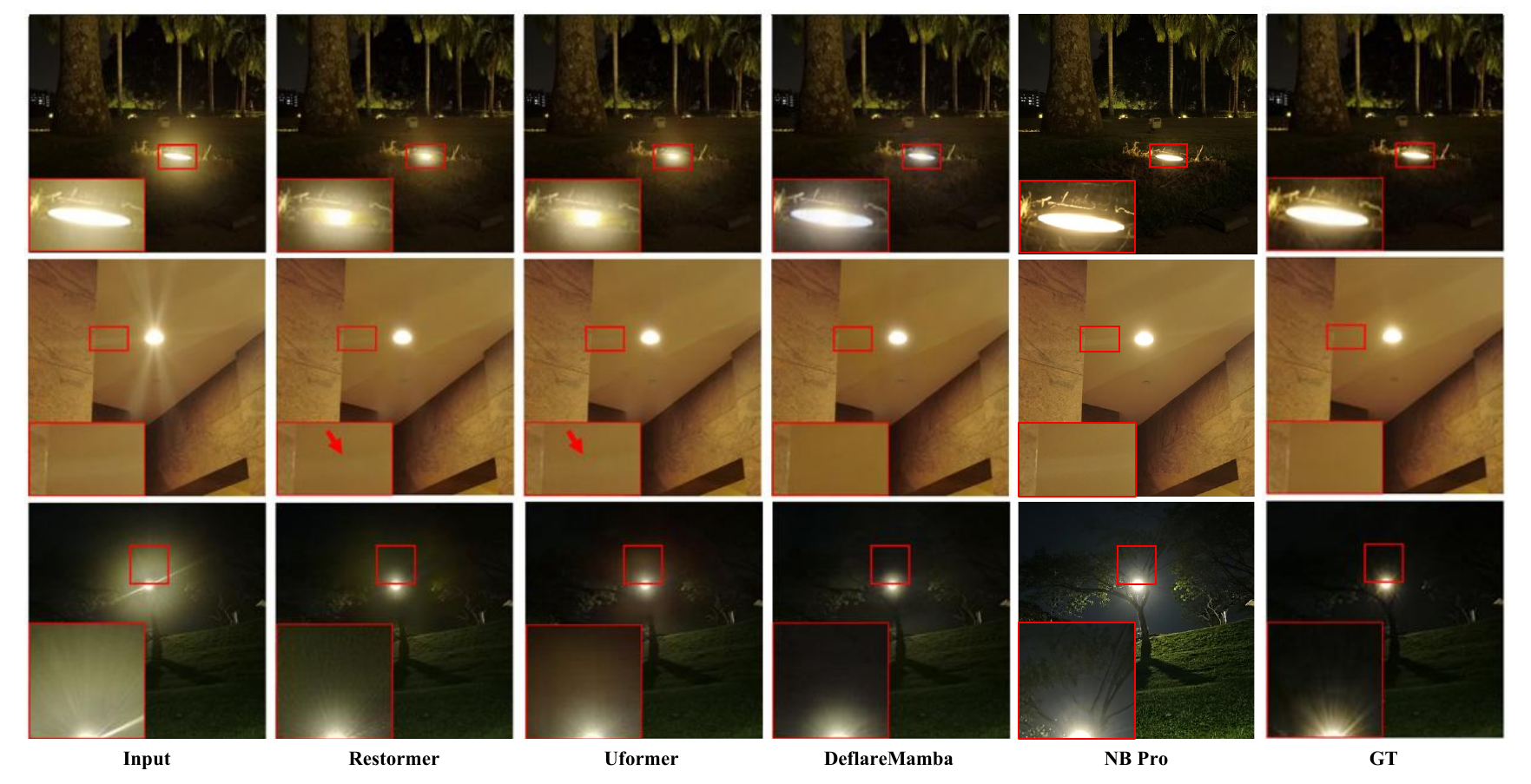}  
  \caption{Qualitative comparison of flare removal results on the Flare7K++ dataset. Nana Banana Pro can preserve image details near light sources and achieves clean removal of streak artifacts. However, it may introduce some brightness changes, as shown in the third row.} 
  \vspace{-5pt}
  \label{flare_figure1}  
\end{figure}

\textbf{Qualitative Analysis.}
Visual comparisons in Fig.~\ref{flare_figure1} reveal a distinct dichotomy in the model's performance: 

\textbf{1. Visual Superiority vs. Stochastic Instability}: On optimal inputs, Nano Banana Pro demonstrates exceptional deflaring capabilities, often surpassing SOTA methods in detail restoration. However, this advantage is compromised by the inherent stochasticity of diffusion models. The model exhibits significant variance and is prone to semantic hallucinations—such as generating unrelated content, suppressing valid light sources, or erroneously illuminating inactive bulbs. While prompt engineering offers partial mitigation, it fails to guarantee the deterministic reliability required for industrial deployment. 

\begin{figure}[H]  
  \centering  
  \includegraphics[width=0.8\textwidth]{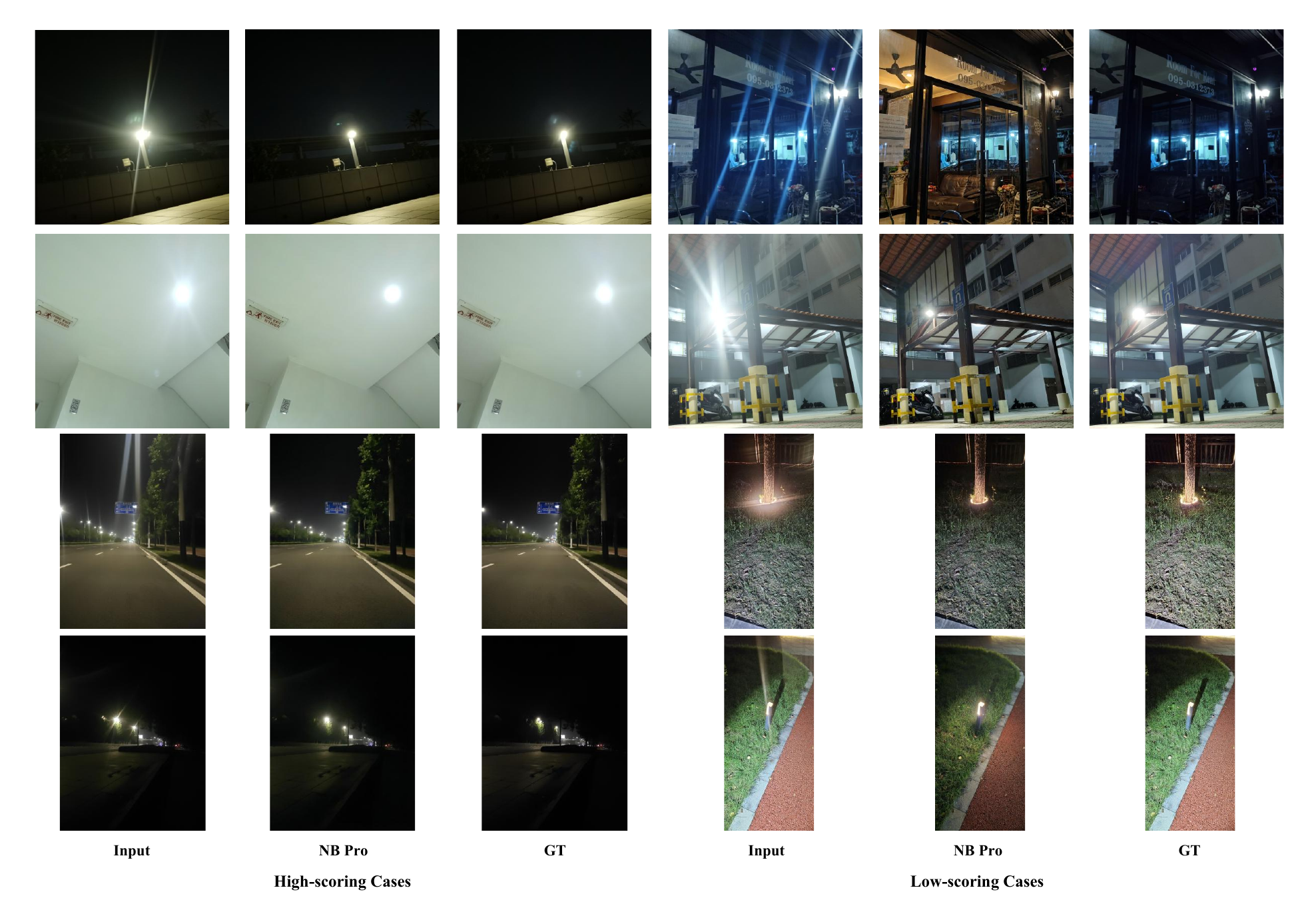}  
  \caption{Examples of some high-scoring and low-scoring samples from Nano Banana Pro on the Flare7K++ dataset. It can be observed that low-scoring samples sometimes appear visually satisfactory, yet their quantitative metric scores may be lower due to certain discrepancies in brightness or color compared to the ground truth.} 
  \vspace{-5pt}
  \label{flare_figure2}  
\end{figure}
\textbf{2. Perceptual-Metric Misalignment}: As illustrated in Fig.~\ref{flare_figure2}, we observe a notable divergence between quantitative metrics and perceptual quality. Instances with low scores sometimes retain high visual fidelity, suggesting that pixel-level metrics may not fully capture the perceptual advantages of generative reconstruction.

In conclusion, Nano Banana Pro exhibits a ``high ceiling, low floor'' characteristic. While it possesses the generative potential to outperform traditional regression-based methods in perceptual quality, it currently sacrifices the stability and consistency essential for robust image restoration.

\newpage

\newpage
\textbf{\LARGE{Image Enhancement}}
\section{Low Light Image Enhancement}

\subsection{Background}
Low-light image enhancement~\cite{lolv1,lolv2,sice} aims to recover visually pleasing images from photographs captured under insufficient illumination. This task presents significant challenges, including brightness adjustment~\cite{retinexformer}, noise suppression~\cite{GSAD}, and color restoration, all while preserving structural details and semantic content. Traditional approaches to this problem have relied on handcrafted priors or supervised deep learning methods trained explicitly on paired low-light and normal-light images.

Recent advances in unified multimodal models~\cite{lu2025hyper}, which jointly handle image understanding and generation within a single framework, raise an intriguing question: can such models perform low-light enhancement in a zero-shot manner, leveraging their broad visual and semantic knowledge without task-specific training? In this chapter, we evaluate Nano Banana Pro, a unified generation and understanding multimodal model, on the low-light image enhancement task. Our evaluation spans three widely-used benchmarks, LOLv1~\cite{lolv1}, LOLv2-real~\cite{lolv2}, and SICE~\cite{sice}, then we compare Nano Banana Pro's zero-shot performance against state-of-the-art supervised and unsupervised methods.

\subsection{Experiment Setup}
\noindent\textbf{Datasets.} We conduct experiments on three established low-light enhancement benchmarks. LOLv1~\cite{lolv1} contains 485 training pairs and 15 testing pairs of real-world low-light and normal-light images captured by adjusting camera exposure time and ISO. LOLv2-real~\cite{lolv2} extends this with 689 training pairs and 100 testing pairs, featuring more diverse indoor and outdoor scenes. SICE~\cite{sice} is a larger-scale dataset containing multi-exposure sequences, from which low-light and reference pairs are constructed; its test set comprises a more diverse range of scenes and illumination conditions. 

\noindent\textbf{Evaluation Metrics.} we adopt two standard full-reference image quality metrics: Peak Signal-to-Noise Ratio (PSNR) and Structural Similarity Index (SSIM). Higher values indicate better reconstruction quality relative to the ground-truth normal-light images.

\noindent\textbf{Comparison Methods.} We compare Nano Banana Pro against several representative low-light enhancement methods spanning different paradigms: ZeroDCE~\cite{zerodce} (zero-reference learning), RUAS~\cite{ruas} (architecture search-based), LLFlow~\cite{llflow}(normalizing flow-based), LLFormer~\cite{llformer} (transformer-based), GSAD~\cite{GSAD} (diffusion-based), and Quadprior~\cite{quadprior} (diffusion prior-based). These methods represent the current state of the art in supervised and unsupervised low-light enhancement.

\noindent\textbf{Nano Banana Pro Configuration.} Nano Banana Pro is evaluated in a zero-shot setting without any fine-tuning on low-light enhancement data~\cite{lolv1,lolv2,sice}. We provide the model with the following natural language instruction: ``This is a low-light image, please turn this image into a normal image while keeping other elements unchanged.'' No additional inference-time configurations or post-processing steps are applied.

\subsection{Qualitative and Quantitative Results}
Tab.~\ref{tab:lol_quantitative} presents the quantitative comparison across all three benchmarks. On LOLv1~\cite{lolv1} and LOLv2-real~\cite{lolv2}, Nano Banana Pro's zero-shot performance falls considerably short of the state-of-the-art supervised methods. The gap is particularly pronounced on LOLv2-real~\cite{lolv2}, where the PSNR of 15.661 dB and SSIM of 0.537 lag behind leading methods by a substantial margin. This suggests that without task-specific training, the model struggles to consistently produce enhancements that align with the ground-truth references in these benchmarks.
Interestingly, on the SICE~\cite{sice} dataset, Nano Banana Pro achieves slightly higher metrics than several comparison methods, demonstrating competitive zero-shot performance on this more challenging and diverse benchmark.

\begin{figure}[htbp]
    \centering 
    \includegraphics[width=\textwidth]{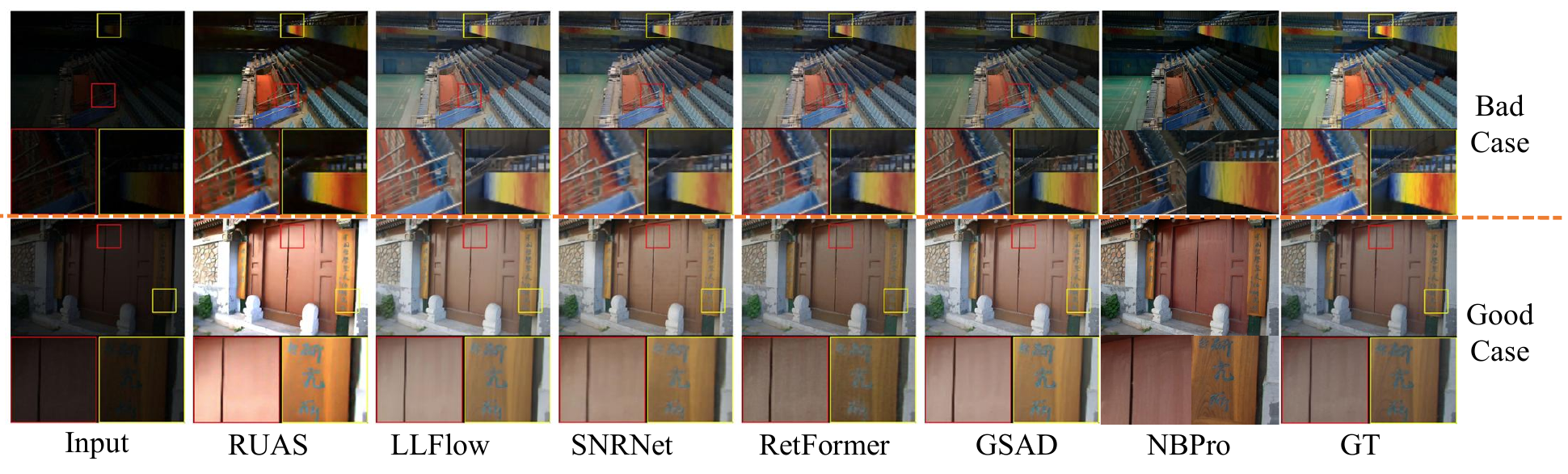} 
    \caption{Visual comparison examples of Nano Banana Pro and several representative specialists. The first row shows its shortcomings in brightness consistency, while the second row shows its superiority in detail preservation.}
    \label{fig:psnr_comparison}
    \vspace{-5mm}
\end{figure}

\begin{table*}[htbp]
  \centering
  \caption{Quantitative comparisons on the LOL and SICE datasets. The best results are highlighted by \textbf{black bold.}}
  \label{tab:lol_quantitative}
  \resizebox{\linewidth}{!}{
  \begin{tabular}{l l c c c c c c c c c}
    \toprule
    \multirow{2}{*}{Methods} & \multirow{2}{*}{Color Model}  & \multicolumn{3}{c}{LOLv1~\cite{lolv1}} & \multicolumn{3}{c}{LOLv2-Real~\cite{lolv2}} & \multicolumn{3}{c}{SICE~\cite{sice}} \\
     \cmidrule(lr){3-5} \cmidrule(lr){6-8} \cmidrule(lr){9-11}
    &  & PSNR$\uparrow$ & SSIM$\uparrow$ & LPIPS$\downarrow$ & PSNR$\uparrow$ & SSIM$\uparrow$ & LPIPS$\downarrow$ & PSNR$\uparrow$ & SSIM$\uparrow$ & LPIPS$\downarrow$ \\
    \midrule
    RetinexNet~\cite{lolv1}     & Retinex         & 18.915 & 0.427 & 0.470 & 16.097 & 0.401 & 0.543 & 12.424 & 0.613 & - \\
    KinD~\cite{kind}           & Retinex          & 23.018 & 0.843 & 0.156 & 17.544 & 0.669 & 0.375 & - & - & - \\
    ZeroDCE ~\cite{zerodce}       & RGB               & 21.880 & 0.640 & 0.335 & 16.059 & 0.580 & 0.313 & 12.452 & \textbf{0.639} & - \\
    RUAS ~\cite{ruas}          & Retinex           & 18.654 & 0.518 & 0.270 & 15.326 & 0.488 & 0.310 & 8.656 & 0.494 & - \\
    LLFlow ~\cite{llflow}        & RGB              & 24.998 & 0.871 & 0.117 & 17.433 & 0.831 & 0.176 & 12.737 & 0.617 & - \\
    EnlightenGAN ~\cite{enlightengan}  & RGB              & 20.003 & 0.691 & 0.317 & 18.230 & 0.617 & 0.309 & - & - & - \\
    SNR-AW ~\cite{snraware}        & SNR+RGB          & {26.716} & 0.851 & 0.152 & 21.480 & \textbf{0.849} & 0.163 & - & - & - \\
    Bread ~\cite{bread}          & YCbCr            & 25.299 & 0.847 & 0.155 & 20.830 & 0.847 & 0.174 & - & - & - \\
    PairLIE ~\cite{pairlie}       & Retinex          & 23.526 & 0.755 & 0.248 & 19.085 & 0.778 & 0.317 & - & - & - \\
    LLFormer ~\cite{llformer}      & RGB              & 25.278 & 0.823 & 0.167 & 20.856 & 0.792 & 0.211 & - & - & - \\
    RetinexFormer ~\cite{retinexformer}  & Retinex         & 27.140 & 0.850 & 0.129 & \textbf{22.794} & 0.840 & 0.171 & - & - & - \\
    GSAD ~\cite{GSAD}          & Retinex         & \textbf{27.605} & \textbf{0.876} & \textbf{0.092} & 20.153 & 0.846 & \textbf{0.113} & - & - & - \\
    QuadPrior ~\cite{quadprior}     & Kubelka-Munk   & 22.849 & 0.800 & 0.201 & 20.592 & 0.811 & 0.202 & - & - & - \\
    \rowcolor{lightgray!50}
    \textbf{\textcolor{orange}{Nano Banana Pro}}       & -               & \textbf{\textcolor{orange}{18.496}} & \textbf{\textcolor{orange}{0.684}} & \textbf{\textcolor{orange}{0.481}} & \textbf{\textcolor{orange}{15.661}} & \textbf{\textcolor{orange}{0.537}} & \textbf{\textcolor{orange}{0.465}} & \textbf{{14.081}} & \textbf{\textcolor{orange}{0.493}} & - \\
    \bottomrule
  \end{tabular}%
  }
\end{table*}
Fig.~\ref{fig:psnr_comparison} presents representative visual comparisons across these three datasets~\cite{lolv1,lolv2,sice}.
Nano Banana Pro produces visually reasonable enhancements in many cases, successfully brightening dark regions and revealing scene content. However, the model exhibits inconsistent brightness control: in the first row, it tends to overexpose bright regions, while in others, it insufficiently enhances dark areas, leaving the output still underexposed. This inconsistency likely stems from the model's reliance on general visual priors rather than explicit illumination modeling.
Notably, Nano Banana Pro does not introduce visible artifacts such as color distortion, halo effects, or structural corruption, which is a common failure mode of some enhancement methods. Texture preservation remains comparable to other approaches, with fine details in enhanced regions generally retained. The absence of artifacts suggests that the model's generative capabilities are well-regularized, even when applied to out-of-distribution tasks like low-light enhancement.

\subsection{Analysis}
The evaluation results reveal a nuanced picture of Nano Banana Pro's zero-shot capabilities for low-light image enhancement. In this section, we provide an in-depth analysis of the observed performance patterns, examine potential underlying causes, and discuss broader implications for applying unified multimodal models to image restoration tasks. A notable positive finding is that Nano Banana Pro does not introduce visible artifacts such as color distortion, halo effects around high-contrast edges, amplified noise, or structural corruption. This is significant because artifact introduction is a common failure mode of enhancement methods, particularly those based on generative models. The absence of artifacts may partially explain the moderate PSNR/SSIM scores. More aggressive enhancement methods might achieve higher metrics on average by pushing brightness and contrast more strongly, but at the cost of occasional artifacts. Nano Banana Pro's conservative approach avoids such failures but may sacrifice peak performance. For practical applications where artifact-free outputs are critical, this trade-off may be acceptable. However, the performance gap on standard benchmarks indicates that zero-shot application is not yet competitive with task-specific methods. The lack of explicit illumination modeling, sensitivity to prompt formulation, and inability to match benchmark-specific ground-truth definitions all limit current performance. Several avenues could potentially improve performance: 1) prompt engineering to provide more specific enhancement guidance; 2) few-shot learning with example image pairs to calibrate the model's enhancement behavior; 3) lightweight fine-tuning or adapter-based adaptation to inject task-specific knowledge while preserving general capabilities; and 4) hybrid approaches that combine unified models' semantic understanding with task-specific enhancement modules.
\section{Underwater Image Enhancement}
\subsection{Background}
Underwater image degradation is an inherent visual quality loss issue in marine environments. As light propagates through water, it undergoes selective absorption by water molecules, multiple scattering by suspended particles such as plankton and sediments, and complex illumination fluctuations, resulting in a series of characteristic image defects. These defects manifest primarily in three core types: color distortion, where red light rapidly attenuates in shallow waters, resulting in a blue-green color bias; contrast reduction stemming from haze-like obscuration caused by scattered light, significantly lowering the image signal-to-noise ratio; and texture blurring, a direct consequence of high-frequency edge information being obscured by scattered particles. Beyond compromising the visual presentation of underwater scenes, these degradations severely undermine the reliability of downstream computer vision tasks. This includes missed detections in underwater object recognition, biased seabed topography reconstruction, and misinterpreted biological behavior analysis. Consequently, it poses significant risks to safety-critical applications such as marine resource exploration, underwater cultural heritage archaeology, unmanned submersible navigation, and seabed infrastructure maintenance.

Traditional underwater image enhancement methods primarily rely on passive restoration, typically employing pixel-domain processing \cite{iqbal2010enhancing,ghani2015underwater,ancuti2018color,gao2019underwater,ancuti2012enhancing} or physical modeling \cite{he2010single,galdran2015automatic,drews2016underwater, li2016single,li2016underwater,zhang2022underwater} to optimize images. However, these methods rely on manually designed rules and exhibit weak generalization capabilities in complex underwater scenes. Deep learning approaches for underwater image enhancement adopt a data-driven strategy, leveraging neural network architectures to learn end-to-end mapping relationships between degraded and clear images. They significantly outperform traditional methods in color correction and detail restoration. Diffusion model approaches for underwater image enhancement \cite{tang2023transformerdiffusion, zhao2024wfdiff} leverage progressive noise injection and reverse denoising mechanisms to effectively balance global tonal restoration with local texture preservation, emerging as one of the leading techniques in recent years. The UIEB \cite{li2019underwater} dataset serves as the benchmark for underwater image enhancement. This section utilizes its released 89-image challenge test set, which contains degraded real images and corresponding high-quality reference images to evaluate models' adaptability to complex scenarios. The LSUI \cite{peng2023ushapetransformer} dataset is a large-scale real underwater image dataset. This section adopts the test set partitioning from WF-Diff \cite{zhao2024wfdiff}, utilizing 400 images, covering diverse water bodies and target categories, making it suitable for evaluating model generalization. The U45 \cite{Li2019Fusion} dataset contains 45 reference-free real underwater images categorized into green cast, blue cast, and haze degradation types, simulating practical application scenarios requiring evaluation without reference images.

Performance evaluation is based on the above three datasets. In this subsection, we systematically assess the underwater image enhancement capabilities of the Nano Banana Pro model, focusing on its effectiveness in eliminating blue/green color casts and restoring blurred textures while preserving scene semantic integrity—such as biological morphology and artifact structure—and maintaining luminance consistency, including natural transitions between light and dark areas.

\begin{table*}[ht]
  \centering
  \vspace{-5pt}
  \caption{Quantitative comparisons on UIEB, LUSI and U45 datasets. The best results are highlighted by \textbf{black bold}.}
  \normalsize
  \setlength{\heavyrulewidth}{1.2pt}
  \setlength{\lightrulewidth}{1pt}  
  \renewcommand{\arraystretch}{1.15}
  \resizebox{0.7\textwidth}{!}{
  \begin{tabular}{l|cc|cc|cc}
    \toprule
    \multirow{2}{*}{\textbf{Method}} & \multicolumn{2}{c|}{\textbf{UIEB}} & \multicolumn{2}{c|}{\textbf{LUSI}} & \multicolumn{2}{c}{\textbf{U45}} \\
    \cmidrule(lr){2-3} \cmidrule(lr){4-5} \cmidrule(lr){6-7}
     & $UIQM$$\uparrow$ & $UCIQE$$\uparrow$ & $UIQM$$\uparrow$ & $UCIQE$$\uparrow$ & $UIQM$$\uparrow$ & $UCIQE$$\uparrow$ \\
    \midrule
    UWCNN~\cite{li2019underwater} & {3.8325} & 0.5552 & {4.1699} & 0.5453 & 4.387 & 0.5622 \\
    UIEC²-Net~\cite{wang2021uiec2net} & 3.327 & 0.609 & 3.9833 & {0.5888} & \textbf{4.4293} & {0.6104} \\
    U-Shape~\cite{peng2023ushapetransformer} & 3.332 & 0.5751 & 4.0334 & 0.574 & 4.3524 & 0.5856 \\
    PUGAN~\cite{cong2023pugan} & 3.2163 & \textbf{0.6176} & 4.0417 & 0.5834 & 4.3377 & \textbf{0.6117} \\
    DM-Water~\cite{tang2023transformerdiffusion} & \textbf{3.8925} & {0.5994} & 4.0595 & 0.5883 & 4.1986 & 0.586 \\
    WF-Diff~\cite{zhao2024wfdiff} & 3.7388 & 0.5867 & 4.0308 & 0.5688 & 4.2193 & 0.5813 \\
    \rowcolor{lightgray!50}
    \textbf{\textcolor{orange}{NB Pro}} & \textbf{\textcolor{orange}{3.634}} & \textbf{\textcolor{orange}{0.5899}} & \textbf{4.2993} & \textbf{0.5961} & \textbf{\textcolor{orange}{4.3907}} & \textbf{\textcolor{orange}{0.5899}} \\
    \bottomrule
  \end{tabular}
  }
  \label{uie_table1}
\end{table*}
\subsection{Quantitative Results}
We conducted experiments using the Nano Banana Pro model configured to output 1K resolution across three datasets, quantitatively evaluating underwater image enhancement performance through reference-free metrics. The Underwater Image Quality Measure (UIQM)~\cite{panetta2015hvsunderwater} comprehensively quantifies underwater image quality by integrating color richness, sharpness, and contrast. Underwater Color Image Quality Evaluation (UCIQE)~\cite{yang2015underwatercolor} addresses non-uniform color shifts and low contrast in underwater scenes by evaluating quality across standard deviation, luminance contrast, and saturation mean dimensions. Both metrics adapt to real-world unreferenced scenarios without requiring reference images. Results were compared against UWCNN~\cite{li2019underwater}, UIEC²-Net~\cite{wang2021uiec2net}, U-Shape~\cite{peng2023ushapetransformer}, PUGAN~\cite{cong2023pugan}, DM-water~\cite {tang2023transformerdiffusion}, and WF-Diff~\cite {zhao2024wfdiff}. Relevant details are summarized in Tab.~\ref{uie_table1}.

NB Pro demonstrates competitiveness on underwater image reference-free evaluation metrics UIQM and UCIQE that rivals existing mainstream UIE methods. On the UIEB dataset, NB Pro's gap with optimal results in reference-free metrics is relatively small. On the larger-scale LSUI dataset with more complex degradation scenarios, NB Pro achieves top performance in both UIQM and UCIQE metrics, fully validating its robustness in challenging environments. On the reference-free dataset U45, NB Pro achieves the best UIQM score and ranks third in UCIQE, demonstrating its practical value in real-world reference-free scenarios.
\begin{figure}[ht]  
  \centering  
  \includegraphics[width=1.0\textwidth]{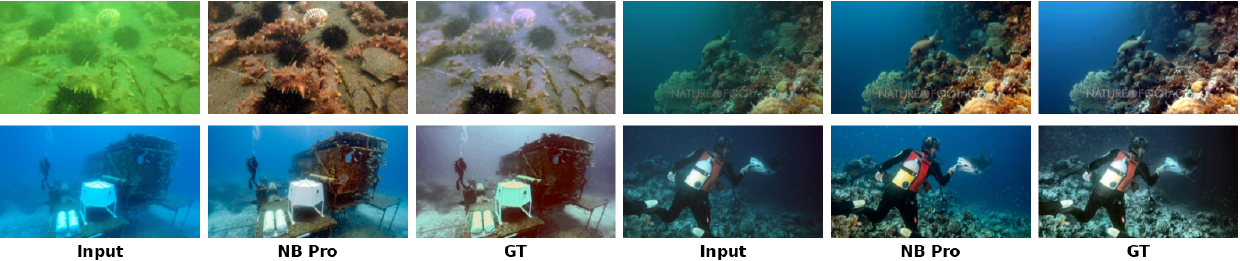}  
  \caption{Visualization examples for underwater image enhancement using NB Pro on the UIEB and LSUI datasets.} 
  \vspace{-5mm}
  \label{uie1}  
\end{figure}

\subsection{Qualitative Results}
\begin{figure}[ht]  
  \centering  
  \includegraphics[width=1.0\textwidth]{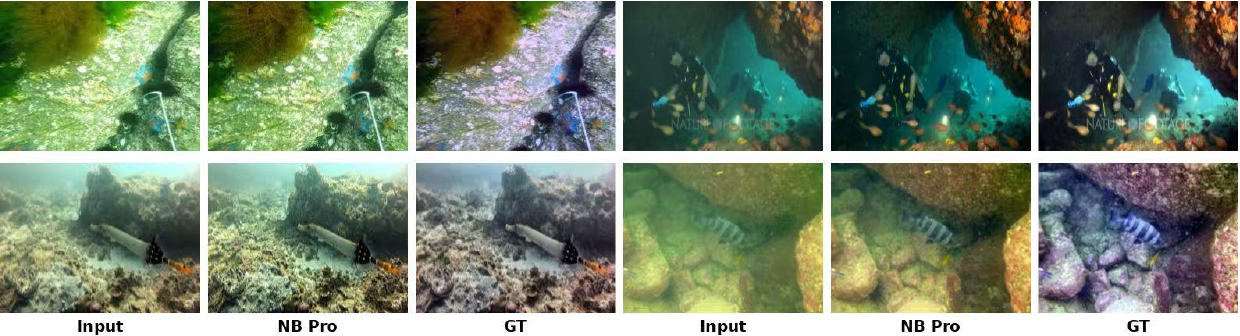}  
  \caption{Visualization of failed cases in underwater image enhancement for NB Pro on the UIEB and LSUI datasets.} 
  \vspace{-5pt}
  \label{uie2}  
\end{figure}

\begin{figure}[H]  
  \centering  
  \includegraphics[width=1.0\textwidth]{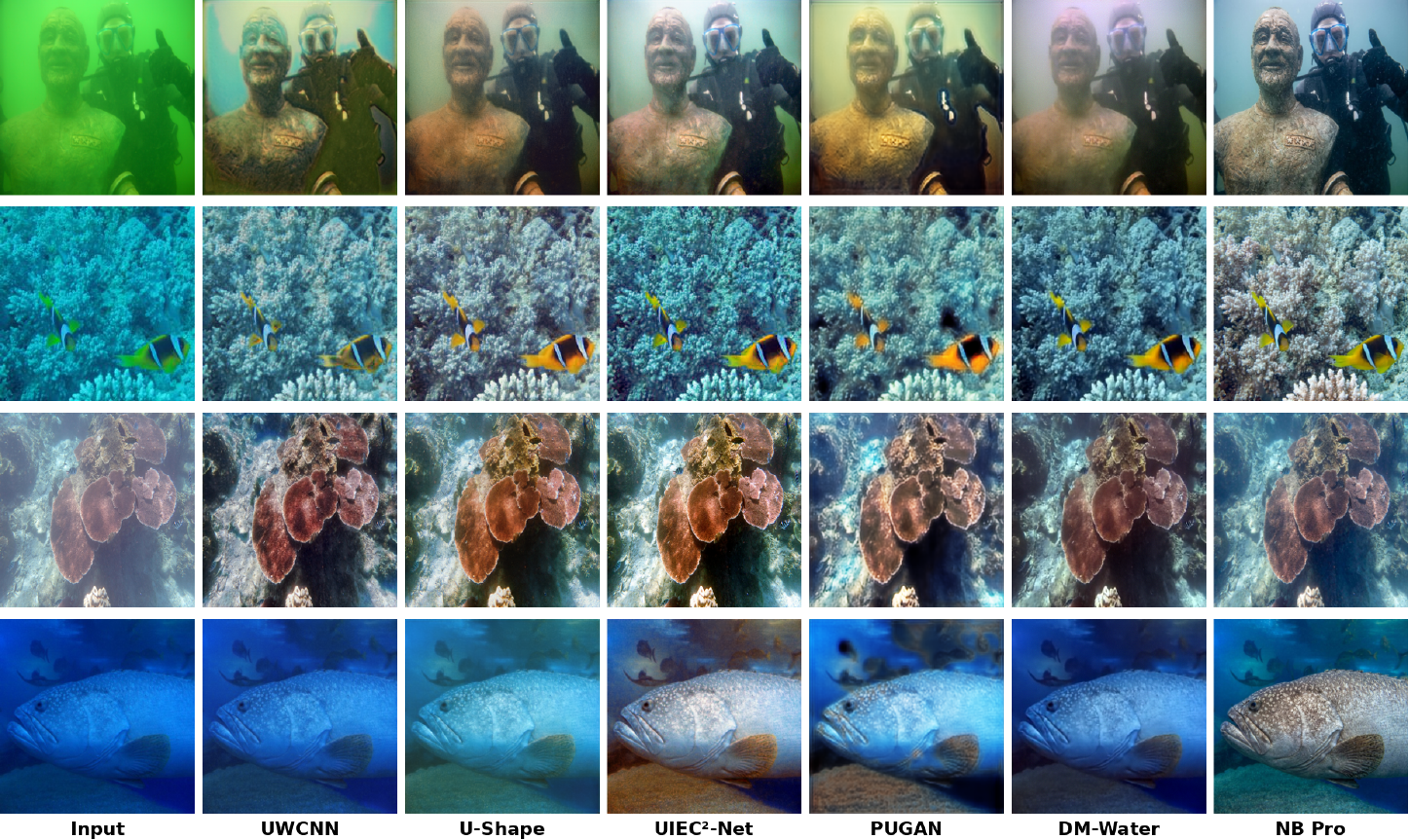}  
  \caption{Visual comparison of processing results between NB Pro and other underwater image enhancement methods on the U45 dataset without reference.} 
  \vspace{-5pt}
  \label{u45}  
\end{figure}

To intuitively present the underwater image enhancement outcomes of the Nano Banana (NB) Pro model, we provide visualizations of its processing results across the UIEB, LSUI, and U45 datasets, with comparisons to state-of-the-art baseline methods.
Fig.~\ref{uie1} displays the visualization of exemplary cases for NB Pro in underwater image enhancement tasks on the UIEB and LSUI datasets.
Fig.~\ref{uie2} presents the visualization of failed cases for NB Pro in underwater image enhancement tasks on the UIEB and LSUI datasets.
Fig.~\ref{u45} shows the visual comparison of processing results between NB Pro and other mainstream underwater image enhancement methods on the reference-free U45 dataset.

Qualitative experimental results demonstrate that for extreme degradation scenarios such as severe green color bias, severe blue color bias, insufficient illumination, and high turbidity, NB Pro can generate high-quality enhanced results without relying on GT images from paired training data. Even in scenarios with multiple severe degradations overlapping, its output images exhibit superior visual quality compared to GT images (as shown in Fig.~\ref{uie1}). However, in mildly degraded scenarios with weaker color shifts, NB Pro struggles to effectively identify degradation features. The generated enhanced images exhibit minimal differences from the input images, resulting in visual quality inferior to GT images and failing to fully meet the core objective of underwater image enhancement (as shown in Fig.~\ref {uie2}). This conclusion is further validated by visualization results from the reference-free dataset U45 (Fig.~\ref {u45}): In the first row depicting a diver scene with high turbidity and severe green color cast degradation, NB Pro's enhancement significantly outperforms other comparison methods, completely eliminating the color cast while removing foggy blur; In the second row of underwater scenes with blue color cast, NB Pro also performed comparably to existing mainstream methods; however, in the third row of slightly degraded underwater plant scenes, NB Pro's restoration results were relatively poor among all comparison methods, failing to effectively optimize image details and contrast.

\subsection{Analyses}
This study conducted a comprehensive evaluation across multiple benchmark datasets representing both synthetic and real underwater environments. Results demonstrate that NB Pro offers a unique paradigm for underwater image enhancement. Its core characteristic is a distinct trade-off between robust perceptual recovery in complex environments and precise pixel-level fidelity in benign conditions.

Quantitative analysis demonstrates that NB Pro excels in the absence of reference images, achieving state-of-the-art results on large-scale complex datasets like LSUI and real-world datasets such as U45. This indicates that when confronted with severely degraded images, the model efficiently generates images with perceptual clarity, rich color saturation, and optimal contrast. Existing underwater ground truth images often exhibit residual degradation or artifacts. NB Pro's generative freedom enables it to surpass the visual quality of these reference images, consistent with its performance in the no-reference evaluation.

From a qualitative perspective, NB Pro demonstrates unique strengths in handling extremely degraded scenarios. Visual examples prove that NB Pro can successfully reconstruct scenes severely affected by green/blue color casts, low light, and high turbidity. It can simultaneously synthesize clear details and correct severe color shifts, sometimes producing results that visually outperform the ground truth images. This indicates NB Pro possesses a strong understanding of underwater degradation and the potential to reverse such degradation.

Despite these strengths in high-intensity restoration tasks, the model exhibits instability in mildly degraded scenarios. For scenes with only slight color shifts or minor turbidity, its enhancement effects are suboptimal, reflecting NB Pro's sensitivity limitations. When degradation signals are weak, the model struggles to identify or localize degraded features, failing to trigger necessary optimizations for fine details and contrast. Future research should focus on enhancing NB Pro's adaptability across the full spectrum of degradation levels. This will ensure it achieves both refined enhancement in mildly blurred environments and thorough reconstruction in severely degraded scenarios, delivering outstanding results in both cases.
\section{HDR Imaging}
\subsection{Background}
High Dynamic Range reconstruction technology aims to address the core limitations of traditional imaging systems. Conventional imaging systems fail to capture the complete spectrum of light intensity information in real-world scenes leading to irreversible loss of highlight details namely overexposed regions and shadow information namely underexposed regions in low dynamic range images. This inherent limitation not only degrades visual quality but also severely impairs the performance of downstream computer vision tasks such as misaligned object detection blurred scene parsing and inaccurate depth estimation posing substantial risks to safety-critical applications including intelligent surveillance environmental perception for autonomous driving and aerial scene analysis.

HDR reconstruction methods have evolved steadily forming a sophisticated technical framework anchored in large-scale datasets and driven by deep learning. Notably the MIT FiveK\cite{Bychkovsky2011FiveK} dataset serves as a classic benchmark for tone mapping and image enhancement offering large-scale professional-grade HDR-LDR image pairs. It includes 5,000 image pairs 4,500 for model training augmented via random cropping flipping rotation and other techniques and 500 for validation and testing. The dataset covers diverse real-world scenarios from indoor and outdoor environments to low-light and high-contrast scenes and supplies professional-grade HDR reference images with exposure parameters and semantic annotations. This enables the reproduction of real-world dynamic range variations and meets evaluation requirements for algorithms focusing on global tone adjustment and local detail preservation. The HDR+\cite{Vinker2021HDR+} dataset is a large-scale benchmark for real-world HDR imaging focusing on the actual shooting conditions of consumer cameras. It contains over 100,000 RAW image sequences captured by consumer cameras each paired with aligned HDR ground truth images. Its advantages include capturing natural noise patterns sensor characteristics and complex lighting variations such as backlighting and high contrast simulating real-world tone mapping demands while retaining resolution suitable for practical use.

Performance evaluation leverages two benchmark datasets. The FiveK dataset includes 500 aligned LDR and HDR image pairs and the HDR+ benchmark emphasizing real-world applicability includes 250 test image pairs. Evaluations use quantitative metrics including PSNR SSIM and LPIPS complemented by subjective assessments of naturalness and detail fidelity to comprehensively measure model performance. In this section we systematically evaluate the Nano Banana Pro model's HDR reconstruction performance on 480p images using the aforementioned benchmarks. The assessment prioritizes verifying the model's capacity to recover highlight and shadow details across diverse scenes and spatial scales while ensuring semantic consistency and photometric integrity. These results offer valuable reference data for the research community.
\subsection{Quantitative Results}
\begin{table*}[ht]
  \centering
  \vspace{-5pt}
  \caption{Quantitative comparison on HDR+ and MIT-FiveK datasets. The best results are highlighted by \textbf{black bold}.}
  \normalsize
  \setlength{\heavyrulewidth}{1.2pt}
  \setlength{\lightrulewidth}{1pt}  
  \renewcommand{\arraystretch}{1.15}
  \resizebox{0.7\textwidth}{!}{
  \begin{tabular}{l|cccc|cccc}
    \toprule
    \multirow{2}{*}{\textbf{Method}} & \multicolumn{4}{c|}{\textbf{HDR+ (480p)}} & \multicolumn{4}{c}{\textbf{MIT-FiveK (480p)}} \\
    \cmidrule(lr){2-5} \cmidrule(lr){6-9}
     & PSNR$\uparrow$ & SSIM$\uparrow$ & LPIPS$\downarrow$ & $\triangle E \downarrow$ & PSNR$\uparrow$ & SSIM$\uparrow$ & LPIPS$\downarrow$ & $\triangle E \downarrow$ \\
    \midrule
    UPE\cite{wang2019underexposed}      & 23.33 & 0.852 & 0.150 & 7.68 & 21.82 & 0.839 & 0.136 & 9.16 \\
    HDRNet\cite{gharbi2017deep}   & 24.15 & 0.845 & 0.110 & 7.15 & 23.31 & 0.881 & 0.075 & 7.73 \\
    CSRNet\cite{he2020conditional}   & 23.72 & 0.862 & 0.104 & 6.67 & {25.31} & {0.909} & \textbf{0.052} & {6.17} \\
    DeepLPF\cite{moran2020deeplpf}  & 25.73 & {0.904} & {0.073} & 6.05 & 24.97 & 0.897 & 0.061 & 6.22 \\
    LUT\cite{zeng2020learning}      & 23.29 & 0.855 & 0.117 & 7.16 & {25.10} & {0.902} & 0.059 & {6.10} \\
    sLUT\cite{wang2021realtime}     & 26.13 & {0.901} & {0.069} & {5.34} & 24.67 & 0.896 & 0.059 & 6.39 \\
    CLUT\cite{zhang2022clut}     & {26.05} & 0.892 & 0.088 & {5.57} & 24.94 & 0.898 & {0.058} & 6.71 \\
    LLF-LUT++\cite{zhang2025high} & \textbf{28.43} & \textbf{0.924} & \textbf{0.056} & \textbf{4.54} & \textbf{26.06} & \textbf{0.912} & 0.054 & \textbf{4.93} \\
    \rowcolor{lightgray!50}
    \textbf{\textcolor{orange}{NB Pro}}  & \textbf{\textcolor{orange}{14.24}} & \textbf{\textcolor{orange}{0.467}} & \textbf{\textcolor{orange}{0.221}} & \textbf{\textcolor{orange}{19.82}} & \textbf{\textcolor{orange}{19.20}} & \textbf{\textcolor{orange}{0.639}} & \textbf{\textcolor{orange}{0.133}} & \textbf{\textcolor{orange}{11.14}}\\
    \bottomrule
  \end{tabular}
  }
  \label{tab_hdr_compare}
\end{table*}
To comprehensively evaluate Nano Banana Pro's performance in HDR tasks, we quantitatively compared it against a range of advanced traditional and deep learning-based image enhancement methods. To ensure fair comparison, all images were downsampled to 480p resolution for evaluation. We employed four standard metrics: PSNR and SSIM to evaluate perceptual similarity, LPIPS to assess visual similarity, and $\triangle E$ to quantify color differences. Results are shown in Tab. \ref{tab_hdr_compare}. NB Pro significantly underperformed against the comparison methods. On the HDR+ dataset, NB Pro achieved lower PSNR and SSIM than the optimal method, while also exhibiting poorer LPIPS and $\triangle E$ values. On the MIT-FiveK dataset, although its PSNR and SSIM improved, they still lagged significantly behind the optimal method. This result clearly indicates that under the standard full-reference evaluation framework, which prioritizes pixel-level accurate reconstruction and color fidelity, NB Pro's generated results exhibit systematic deviations from professionally enhanced or color-graded reference images.

NB Pro fundamentally differs from traditional HDR/enhancement models optimized for specific imaging scenarios. The latter typically undergo end-to-end training directly on paired LDR-reference images, targeting minimization of pixel-level loss, thus inherently excelling in metrics like PSNR and SSIM. In contrast, NB Pro's generation process prioritizes semantic coherence and overall visual appeal. Its outputs can be viewed as reconstructions of the input image rather than strict pixel-to-pixel mappings. Consequently, generated images may exhibit deviations in luminance distribution, local contrast, and even color style compared to reference images, leading to comprehensive score reductions across full-reference metrics. Notably, on the LPIPS metric, NB Pro's performance on the MIT-FiveK dataset remains behind but shows a narrowed gap compared to pixel-level metrics. This suggests its outputs may retain some similarity to reference images at higher-level semantic features, while low-level pixel arrangements have been significantly altered.
\subsection{Qualitative Results}
\begin{figure*}[ht]  
  \centering  
  \includegraphics[width=1.0\textwidth]{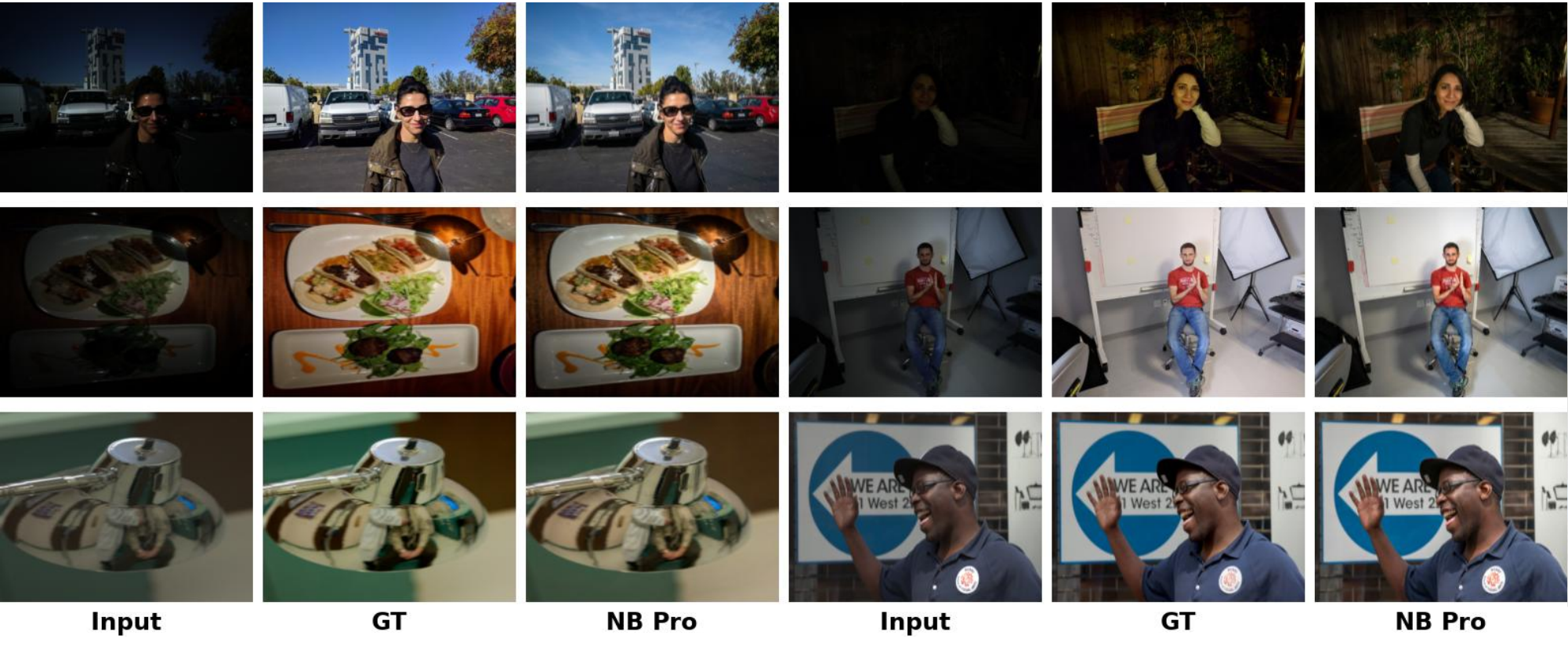}  
  \caption{Visualization of satisfactory exemplary cases for the HDR imaging task using NB Pro.} 
  \vspace{-5pt}
  \label{hdr1}  
\end{figure*}

\begin{figure*}[ht]  
  \centering  
  \includegraphics[width=1.0\textwidth]{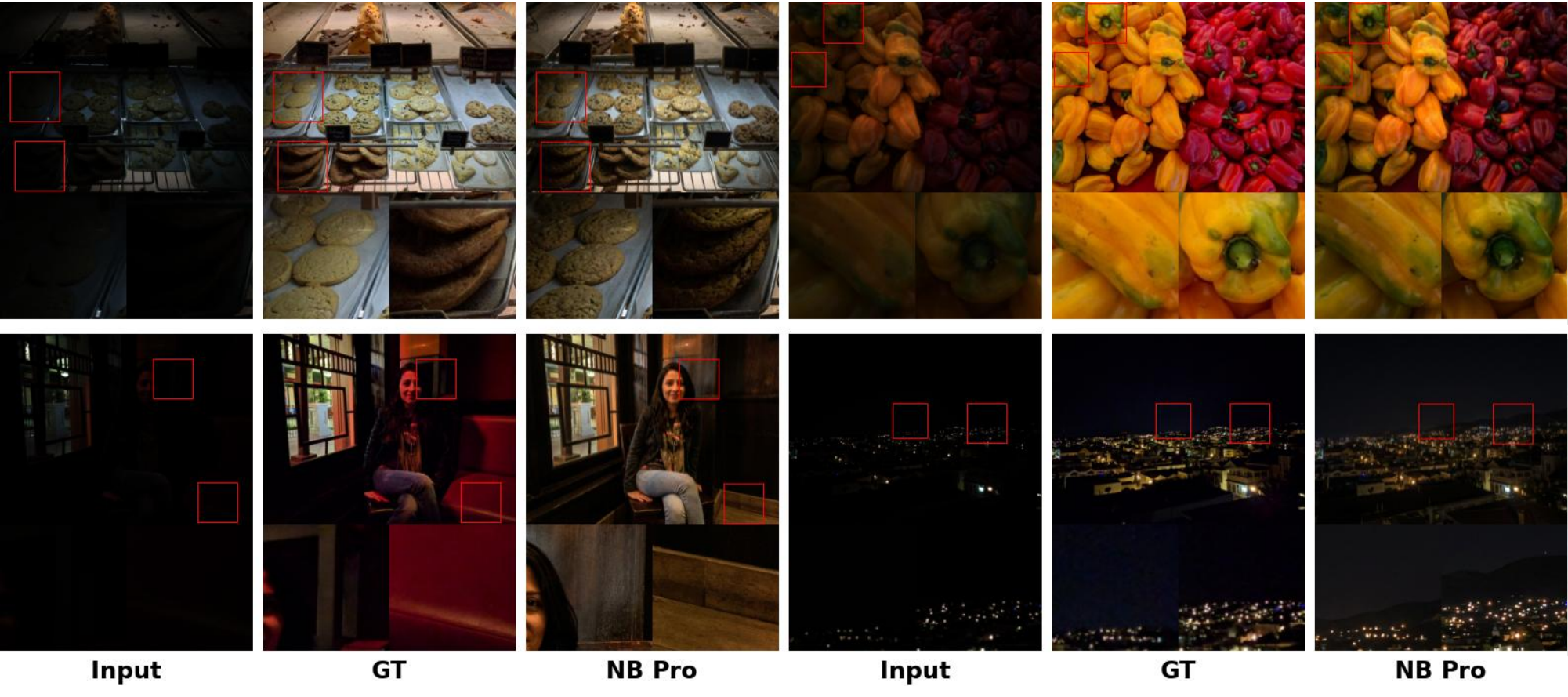}  
  \caption{ Visualization of texture detail loss and artificial addition cases in HDR Imaging using NB Pro.} 
  \vspace{-5pt}
  \label{hdr2}  
\end{figure*}

\begin{figure*}[ht]  
  \centering  
  \includegraphics[width=1.0\textwidth]{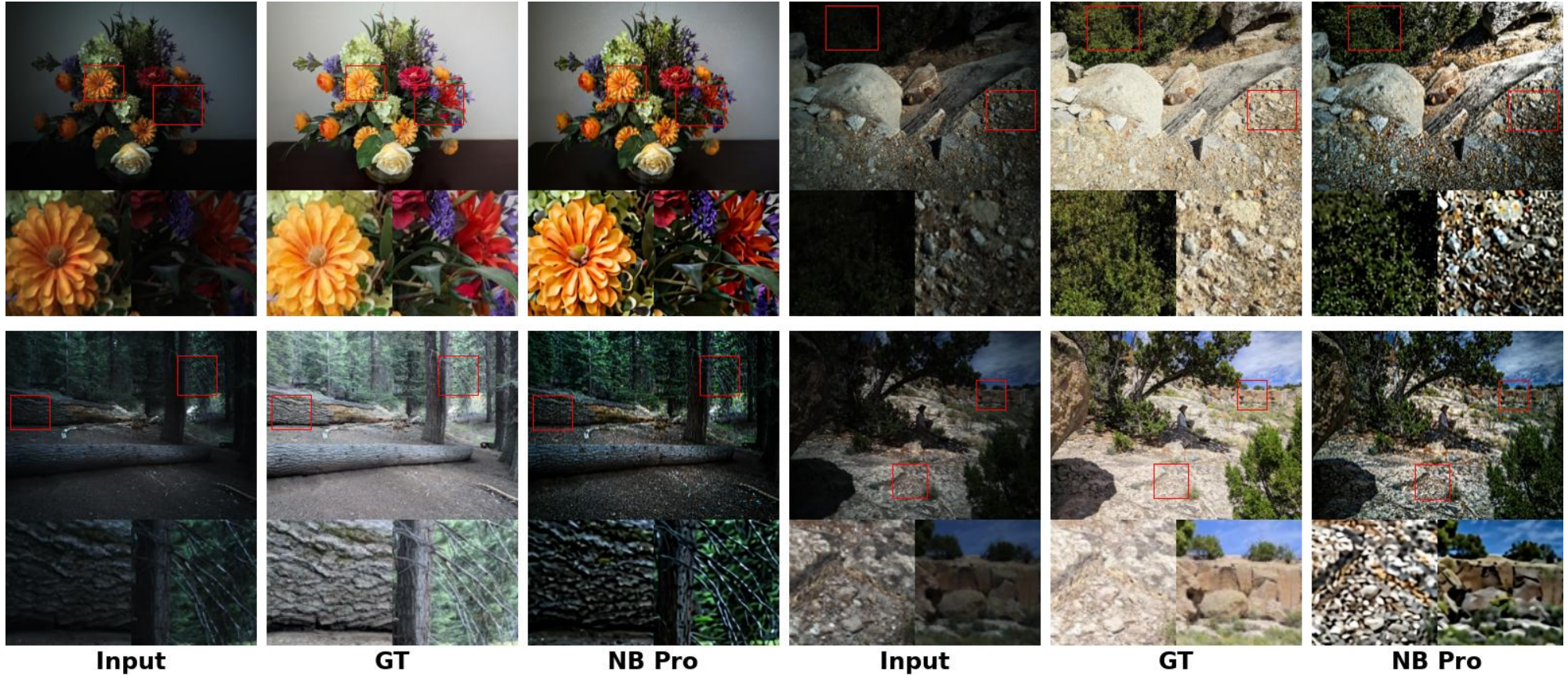}  
  \caption{Visualization of oversharpening cases in the HDR imaging task using NB Pro.} 
  \vspace{-5pt}
  \label{hdr3}  
\end{figure*}
To visually evaluate the performance and potential limitations of NB Pro in HDR imaging tasks, this study conducted qualitative visualization experiments using representative scenes from the HDR+ and MIT-FiveK datasets. The experiments cover three core scenarios: conventional lighting, low-light with minimal detail, and complex dense textures. The corresponding results are presented in Fig. \ref{hdr1}, Fig. \ref{hdr2}, and Fig. \ref{hdr3}, respectively.

Fig. \ref{hdr1} presents an illustrative case of NB Pro in HDR imaging tasks. Comparing the low-dynamic-range input image, the true high-dynamic-range reference image, and the NB Pro generated result reveals that in conventionally lit scenes, such as moderately bright indoor settings or outdoor natural light environments, NB Pro effectively restores the scene's dynamic range and color gradation. The overall visual quality of the generated result approaches or even rivals the reference image, fully validating the model's effectiveness in fundamental HDR imaging tasks.

Fig. \ref{hdr2} further highlights texture anomalies in low-light, low-detail scenarios. When input low-dynamic-range images contain severe shadow areas with inherently weak texture or detail information, NB Pro tends to exhibit two typical defects. One type involves detail loss, such as in the third case where the model's output lacks the wall tile texture and sofa outline in the shadow regions of the input image. The second involves artificially redundant additions, such as the extra cookie surface texture generated in the first case that did not exist in the original scene. The second case failed to restore the vegetables' original colors and applied redundant diffuse rendering to the green spots. The fourth case generated mountain elements out of thin air in the house background. These issues stem from insufficient detail features in low-light scenes, causing the model's texture perception and generation logic to deviate, ultimately reducing detail restoration accuracy.

Fig. \ref{hdr3} shows evaluation results for complex, densely textured scenes. When input images contain fine, dense textures—such as fabric patterns, intricate vegetation textures, or architectural wall textures—NB Pro struggles to precisely balance texture preservation and enhancement scales. Generated results commonly exhibit visual artifacts from excessive sharpening, overly pronounced texture edges, localized artifacts, and even masking of the original texture's natural gradations. This phenomenon reflects the model's ongoing limitations in perceiving fine textures and controlling enhancement.
In summary, qualitative experiments indicate that NB Pro delivers satisfactory visual effects in standard HDR imaging scenarios. However, in low-light, low-detail environments, the model tends to suffer from texture loss or redundant additions. In complex, densely textured scenes, the model exhibits a tendency toward excessive sharpening.

\subsection{Analyses}
The quantitative and qualitative evaluation results systematically reveal the comprehensive performance characteristics of NB Pro in HDR imaging tasks. Quantitatively, the model significantly lags behind mainstream HDR reconstruction methods in all reference evaluation metrics during 480p resolution tests on the HDR+ and MIT-FiveK datasets. However, the gap narrows for the LPIPS metric on the MIT-FiveK dataset, indicating that while its generated results exhibit significant pixel-level deviations, they maintain a degree of semantic consistency with reference images. Qualitatively, NB Pro achieves acceptable visual results in standard-illumination HDR scenes. However, it exhibits pronounced limitations in two typical scenarios: low-light environments with sparse details and complex, densely textured scenes. Specific issues include texture information loss, artificial redundant detail generation, color distortion, and over-sharpening artifacts.

The evaluation results clearly reveal three core deficiencies of NB Pro when applied to HDR imaging tasks: First, in low-light, low-detail scenes, the model exhibits weaknesses in perceiving and accurately restoring subtle texture features in shadow areas. The weak feature signals in these input shadows struggle to support precise reconstruction, leading to either the omission of shadow texture information or the generation of artificial redundant detail fill. Second, for fine-grained texture scenarios like fabric weaves and dense vegetation patterns, the model lacks adaptive control over texture preservation and enhancement scales. Over-sharpening not only disrupts natural texture gradation but also readily induces edge artifacts. Third, the model's color rendering mechanism lacks strong constraints on reference color distributions. In complex color scenes, it prioritizes visual harmony over precise target color reproduction, ultimately causing color distortion. Comprehensive experimental results indicate that NB Pro is only suitable for non-critical HDR imaging scenarios where pixel-level precision is less demanding and visual experience is paramount. It is unsuitable for safety-critical or professional-grade applications requiring stringent detail and color restoration accuracy. Issues such as texture loss, artificial redundant details, and color distortion may cause scene analysis bias or decision misjudgment, failing to meet the core requirements of such scenarios.

\newpage

\newpage
\textbf{\LARGE{Image Fusion}}
\section{Multi-Focus Image Fusion}

\subsection{Introduction}
In the computer vision and digital photography, acquiring fully clear images is crucial for subsequent analysis and processing.
However, due to the limited depth of field (DoF) of camera lenses, a single shot cannot concurrently focus on all the objects at varying depths. 
Multi-focus image fusion (MFIF) provides an effective solution to this challenge, aiming to generate an AIF output from multiple images of the same scene with different focal points. This technique has found significant applications in various fields such as medical diagnosis and consumer electronics.

In recent years, deep learning has been widely applied to MFIF, typically falling into two categories: decision-based and reconstruction-based approaches. 
Decision-based methods learn a decision map by classifying pixels in the source images as either in-focus or out-of-focus, and then select the appropriate pixel to assemble the final composite~\cite{liu2017multi, li2020drpl, xiao2021dtmnet}.
Reconstruction-based methods employ end-to-end networks to directly extract features from source images and reconstruct the all-in-focus output~\cite{zhang2020ifcnn, ma2022swinfusion, li2024fusiondiff}.
However, the former methods normally struggle with focused-defocused boundaries whose focus attributes are ambiguous to distinguish, while the letter ones often suffer from detail loss in other regions away from the boundaries.

More recently, the rapid advancement of Generative Artificial Intelligence has opened new avenues for MFIF. 
Generative models learn the distribution from massive datasets, enabling them to understand and generate complex visual content and structures. 
Despite the potential for creative fusion, their application in pixel-precise tasks like MFIF remains under-explored. 
There is a critical need to verify whether a model designed for creativity can adhere to the strict fidelity requirements of fusion tasks without introducing hallucinations or losing spectral information.
This report evaluates the performance of the newest generative model Nano Banana Pro in the task of MFIF, comparing its fusion quality with existing baseline methods using various metrics. The results aim to provide insights into its strengths and limitations in real-world applications.

\subsection{Quantitative Results}
\begin{table*}[ht]
  \centering
  \caption{Quantitative comparison on the Lytro and MFFW datasets. The best results are highlighted by \textbf{black bold}.}
  \normalsize
  \setlength{\heavyrulewidth}{1.2pt}
  \setlength{\lightrulewidth}{1pt}  
  \renewcommand{\arraystretch}{1.15}
  \resizebox{\textwidth}{!}{
  \begin{tabular}{l|cccccc|cccccc}
    \toprule
    \multirow{2}{*}{\textbf{Method}} & \multicolumn{6}{c|}{\textbf{Lytro}} & \multicolumn{6}{c}{\textbf{MFFW}}\\
    \cmidrule(lr){2-7} \cmidrule(lr){8-13}
     & $NMI$ & $EN$ & $AG$ & $SF$ & $Q_{Y}$ & $Q_{CB}$ & $NMI$ & $EN$ & $AG$ & $SF$ & $Q_{Y}$ & $Q_{CB}$ \\
    \midrule
    IFCNN~\cite{zhang2020ifcnn} & 0.9388 & 7.5318 & 8.1473 & 19.3992 & 0.9518 & 0.7294 & 0.8206 & 7.1688 & 9.6849 & 22.0173 & 0.8715 & 0.6423 \\
    SESF~\cite{ma2021sesf} & 1.6808 & 7.5322 & 8.1530 & 19.4251 & {0.9879} & 0.8064 & 1.0876 & 7.1850 & 9.8678 & 22.9291 & {0.9588} & {0.7418} \\
    GACN~\cite{ma2022end} & 1.1723 & 7.5311 & 8.1245 & 19.3247 & 0.9878 & 0.8062 & 1.0820 & 7.1923 & 9.7328 & 22.2842 & 0.9273 & 0.7192 \\
    GRFusion~\cite{li2023generation} & 1.1879 & 7.5329 & 8.1823 & 19.4539 & 0.9863 & {0.8071} & 1.1426 & 7.1711 & 9.8826 & 22.6520 & 0.9334 & 0.7223 \\
    ZMFF\cite{hu2023zmff} & 0.8795 & 7.5223 & 7.7771 & 18.8184 & 0.9321 & 0.6731 & 0.7711 & 7.1546 & 9.2144 & 21.3405 & 0.8474 & 0.6722 \\
    MUFusion~\cite{cheng2023mufusion} & 0.8088 & \textbf{7.6093} & 8.1578 & 18.9240 & 0.8997 & 0.6819 & 0.7551 & {7.2026} & 8.9247 & 19.9974 & 0.8167 & 0.6205 \\
    DB-MFIF~\cite{zhang2024exploit} & 1.0573 & 7.5386 & 8.2415 & 19.5290 & 0.9637 & 0.7770 & 0.8699 & 7.1935 & 10.1346 & 23.0305 & 0.8663 & 0.6647\\
    MFFT~\cite{zhai2024multi} & 1.1533 & {7.5620} & \textbf{8.6089} & \textbf{21.3759} & 0.9523 & 0.7511 & 1.1310 & \textbf{7.2281} & {10.1971} & \textbf{24.5709} & 0.9336 & 0.6865 \\
    DMANet~\cite{quan2025multi} & {1.1897} & 7.5319 & 8.2059 & 19.5129 & 0.9853 & 0.8054 & {1.1513} & 7.1846 & 10.0948 & {23.2370} & 0.9506 & 0.7276\\
    MCCSR~\cite{zheng2025unfolding} & \textbf{1.1920} & 7.5329 & 8.1757 & 19.4392 & \textbf{0.9890} & \textbf{0.8084} & \textbf{1.1800} & 7.1688 & 9.7407 & 22.2989 & \textbf{0.9813} & \textbf{0.7517}\\
    \rowcolor{lightgray!50}
    \textbf{\textcolor{orange}{NB Pro}} & \textbf{\textcolor{orange}{0.7476}} & \textbf{\textcolor{orange}{7.5267}} & \textbf{\textcolor{orange}{8.2977}} & \textbf{\textcolor{orange}{20.1044}} & \textbf{\textcolor{orange}{0.7755}} & \textbf{\textcolor{orange}{0.6603}} & \textbf{\textcolor{orange}{0.6319}} & \textbf{\textcolor{orange}{7.1823}} & \textbf{10.2879} & \textbf{\textcolor{orange}{23.0223}} & \textbf{\textcolor{orange}{0.5537}} & \textbf{\textcolor{orange}{0.5638}}\\
    \bottomrule
  \end{tabular}
  }
  \label{mfif_table1}
\end{table*}
We evaluate the performance of MFIF on four benchmark: Lytro~\cite{nejati2015multi}, MFFW~\cite{xu2020mffw}, MFI-WHU~\cite{zhang2021mff} and SIMIF~\cite{chun2025multi}. 
The Lytro dataset contains 20 pairs of multi-focus images captured by a light field camera. 
The MFFW dataset includes 13 real image pairs with strong Defocus Spread Effect (DSE). 
The MFI-WHU dataset is constructed using Gaussian blur and decision maps, consists of a larger scale with 120 pairs. 
The SIMIF dataset is composed of 12 pairs of high-resolution images. Six popular objective metrics are employed for evaluation, including non-reference metrics, $EN$~\cite{jahne2005digital}, $AG$~\cite{cui2015detail}, $SF$~\cite{zheng2007new}, and Source-reference metrics $NMI$~\cite{hossny2008comments}, $Q_{Y}$~\cite{yang2008novel},  $Q_{CB}$~\cite{chen2009new}. These datasets and metrics provide a comprehensive assessment from multiple perspectives.

\begin{table*}[ht]
  \centering
  \caption{Quantitative comparison on the MFI-WHU and SIMIF datasets. The best results are in \textbf{black bold}.}
  \normalsize
  \setlength{\heavyrulewidth}{1.2pt}
  \setlength{\lightrulewidth}{1pt}  
  \renewcommand{\arraystretch}{1.2}
  \resizebox{\textwidth}{!}{
  \begin{tabular}{l|cccccc|cccccc}
    \toprule
        \multirow{2}{*}{\textbf{Method}} & \multicolumn{6}{c|}{\textbf{MFI-WHU}} & \multicolumn{6}{c}{\textbf{SIMIF}} \\
    \cmidrule(lr){2-7} \cmidrule(lr){8-13}
     & $NMI$ & $EN$ & $AG$ & $SF$ & $Q_{Y}$ & $Q_{CB}$ & $NMI$ & $EN$ & $AG$ & $SF$ & $Q_{Y}$ & $Q_{CB}$ \\
    \midrule
    IFCNN~\cite{zhang2020ifcnn} & 0.8993 & 7.3285 & 11.3564 & 26.1798 & 0.9404 & 0.7367 & 1.0505 & 7.3897 & 6.9289 & 19.1344 & 0.9211 & 0.7364 \\
    SESF~\cite{ma2021sesf} & 1.1878 & 7.3183 & 11.5399 & 26.5894 & 0.9855 & 0.8166 & 1.2995 & 7.3868 & 6.9931 & 19.4672 & 0.9807 & {0.8306} \\
    GACN~\cite{ma2022end} & 1.2084 & 7.3127 & 11.4633 & 26.5089 & {0.9889} & \textbf{0.8241} & 1.3068 & 7.3802 & 6.9750 & 19.4241 & \textbf{0.9845} & \textbf{0.8328} \\
    ZMFF\cite{hu2023zmff} & 0.7028 & 7.2763 & 10.6727 & 24.7632 & 0.8387 & 0.6512 & 0.8302 & 7.3942 & 7.0007 & 19.2939 & 0.8325 & 0.6259 \\
    GRFusion~\cite{li2023generation} & 1.2134 & 7.3216 & 11.6609 & {26.8362} & 0.9848 & 0.8212 & 1.2898 & 7.3917 & 7.0215 & 19.4954 & 0.9771 & 0.8267 \\
    MUFusion~\cite{cheng2023mufusion} & 0.7449 & \textbf{7.3744} & 9.6015 & 21.4447 & 0.8608 & 0.6480 & 0.8876 & 7.3566 & 6.0633 & 16.3885 & 0.8387 & 0.6321 \\
    DB-MFIF~\cite{zhang2024exploit} & 1.0693 & 7.3250 & {11.6956} & 26.8336 & 0.9466 & 0.7818 & 1.1510 & {7.4073} & {7.1849} & {19.7543} & 0.9157 & 0.7612 \\
    MFFT~\cite{zhai2024multi} & 1.1974 & {7.3358} & \textbf{11.7636} & \textbf{27.5736} & 0.9614 & 0.7648 & 1.2751 & 7.3866 & 7.1754 & \textbf{20.5918} & 0.9492 & 0.7724 \\
    DMANet~\cite{quan2025multi} & {1.2220} & 7.3158 & 11.6003 & 26.8034 & 0.9878 & 0.8222 & {1.3130} & 7.3834 & 7.0634 & 19.5735 & 0.9780 & 0.8270 \\
    MCCSR~\cite{zheng2025unfolding} & \textbf{1.2246} & 7.3120 & 11.4162 & 26.4025 & \textbf{0.9891} & {0.8227} & \textbf{1.3151} & 7.3782 & 6.9240 & 19.3575 & {0.9835} & 0.8298 \\
    \rowcolor{lightgray!50}
    \textbf{\textcolor{orange}{NB Pro}} & \textbf{\textcolor{orange}{0.5923}} & \textbf{\textcolor{orange}{7.2986}} & \textbf{\textcolor{orange}{10.5142}} & \textbf{\textcolor{orange}{24.1274}} & \textbf{\textcolor{orange}{0.5610}} & \textbf{\textcolor{orange}{0.5745}} & \textbf{\textcolor{orange}{0.8042}} & 
    \textbf{7.4925} &
    \textbf{7.3561} & 
    \textbf{\textcolor{orange}{19.5545}} & \textbf{\textcolor{orange}{0.5112}} & \textbf{\textcolor{orange}{0.5775}} \\
    \bottomrule
  \end{tabular}
  }
  \vspace{-10pt}
  \label{mfif_table2}
\end{table*}

\begin{figure*}[ht]  
  \centering  
  \includegraphics[width=0.99\textwidth]{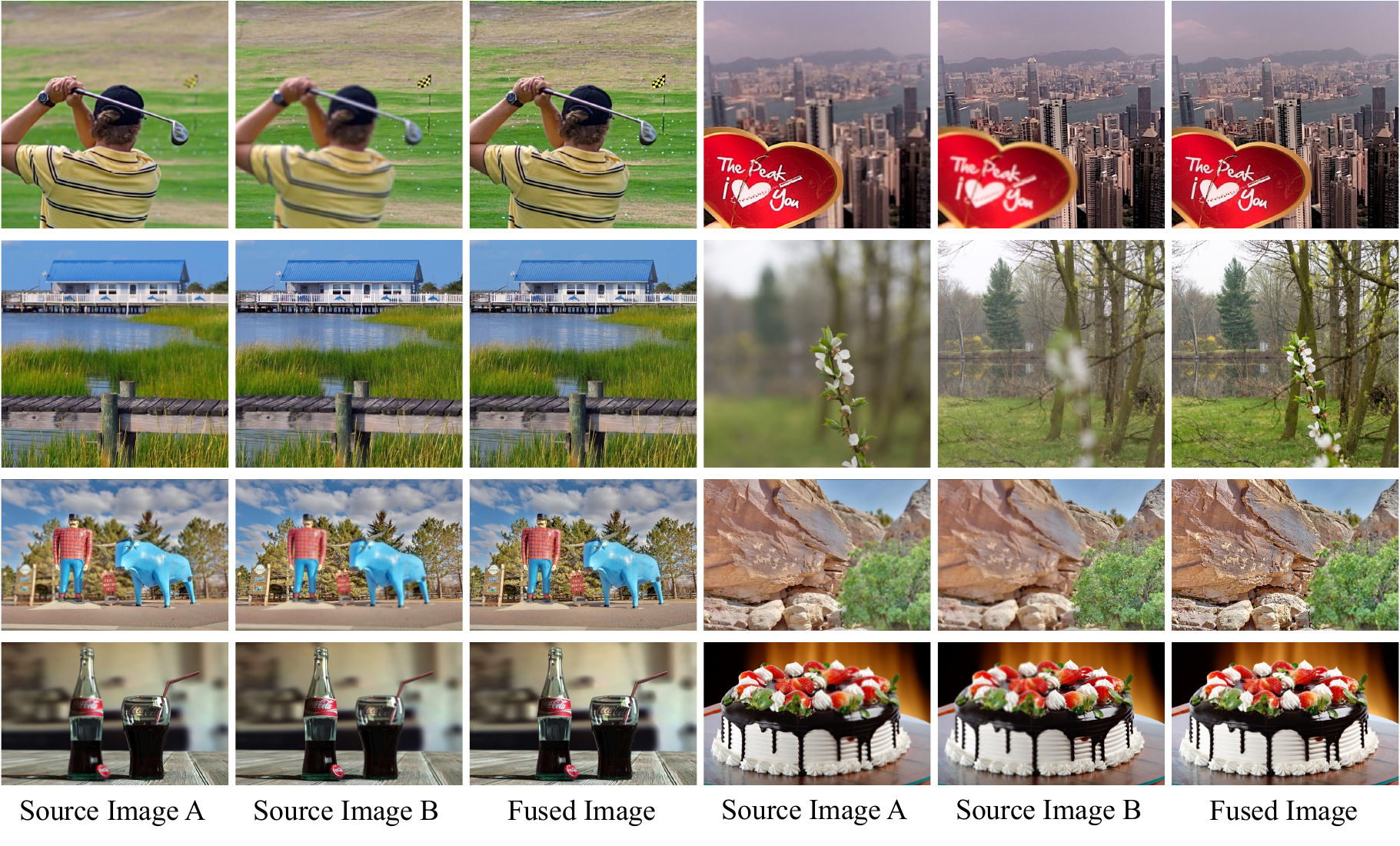}  
  \caption{Visualization of the fusion images generated by the Nano Banana Pro. The rows from top to bottom correspond to samples from the Lytro, MFFW, MFI-WHI and SIMIF datasets.} 
  \vspace{-5pt}
  \label{mfif_figure1}  
\end{figure*}

\begin{figure*}[ht]  
  \centering  
  \includegraphics[width=1.0\textwidth]{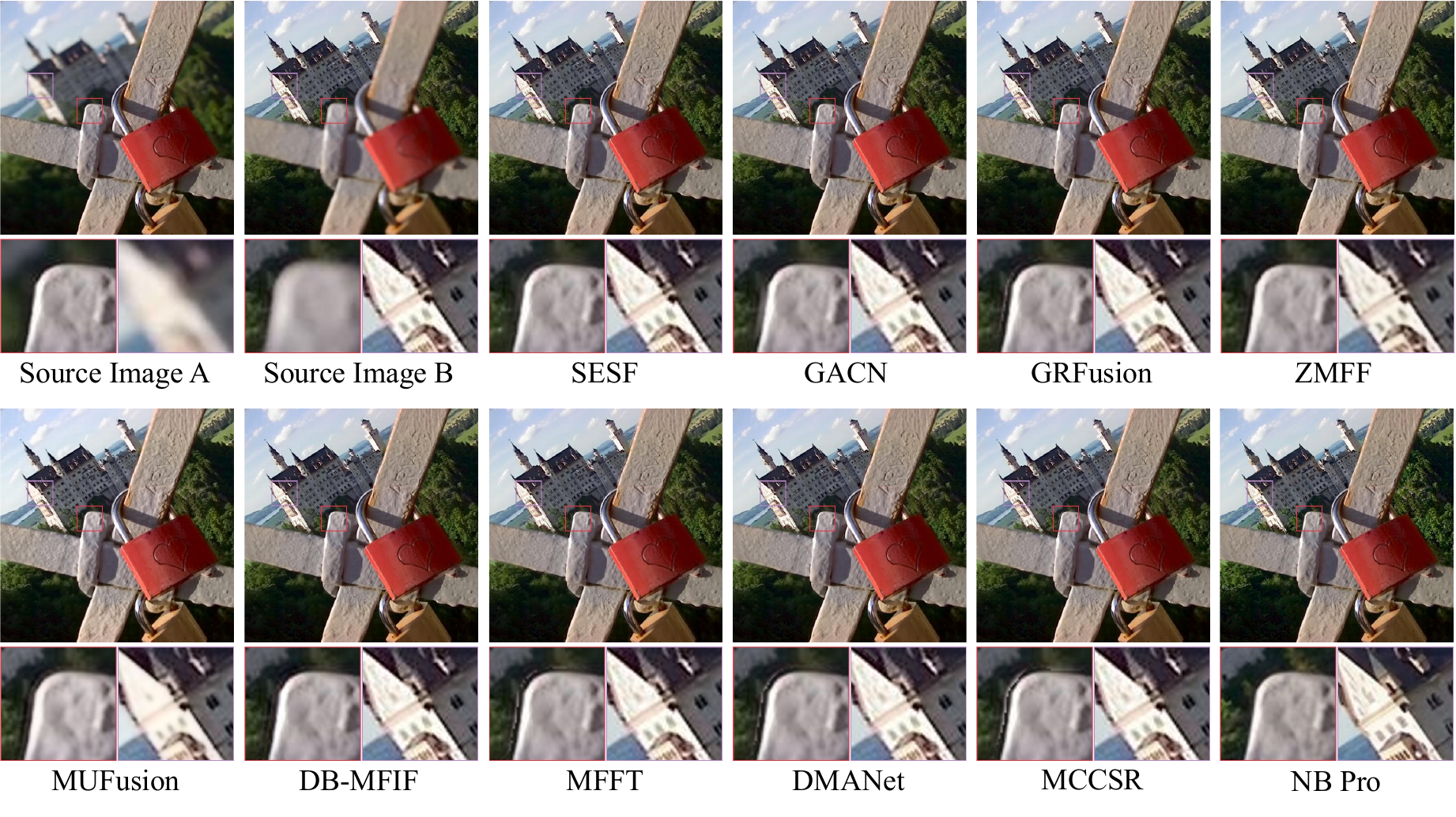}  
  \caption{Visualization of the fusion images of `lock' sample from the Lytro dataset. Two enlarged views are shown to reveal critical details.} 
  \vspace{-5pt}
  \label{mfif_figure2}  
\end{figure*}

We compare the Nano Banana Pro (NB Pro) with 10 other state-of-the-art and representative MFIF methods, where ZMFF is a Zero-shot method, IFCNN and MUFusion are unsupervised methods, and the rest are supervised methods. 
According to the comparison results shown in the Tab.~\ref{mfif_figure1} and Tab.~\ref{mfif_figure2}, NB Pro performs exceptionally well on non-reference metrics, achieving results that are close to or even surpassing the current state-of-the-art, indicating the high quality of the generated images themselves.
Conversely, on source-reference metrics, NB Pro shows poorer performance, meeting the similar dilemma faced by previous Zero-shot and unsupervised methods.
This indicates that during the fusion process, the model failed to adequately preserve consistency between the generated image and the source images in terms of aspects like gradients and structure; it exhibits excessive creativity at the expense of fidelity.

\subsection{Qualitative Results}
Fig.~\ref{mfif_figure1} illustrates a visualization of the fusion results generated by the Nano Banana Pro.
The fusion performance varies across different samples, with some being successful and others subpar. 
For instance, the first example in the first row showcases a favorable fusion result, where the fused image not only preserves the focused foreground person and the background golf course from the source images but also achieves a seamless transition between different regions, effectively avoiding artifacts. 
Notably, it even recovers richer details in the lawn area—which originally had limited clarity—thereby enhancing the overall visual quality.
Conversely, limitations are observed in some other cases. 
Specifically, in the second sample of the second row, the white petals at the base of the foreground plant remain blurred. 
Similarly, in the second sample of the third row, the grass in the bottom-right corner is still defocused. 
Given that sharp counterparts for these regions exist in the source images, this indicates that the model failed to correctly identify or localize the optimal in-focus regions during the fusion process.

\begin{figure*}[ht]  
  \centering  
  \includegraphics[width=1.0\textwidth]{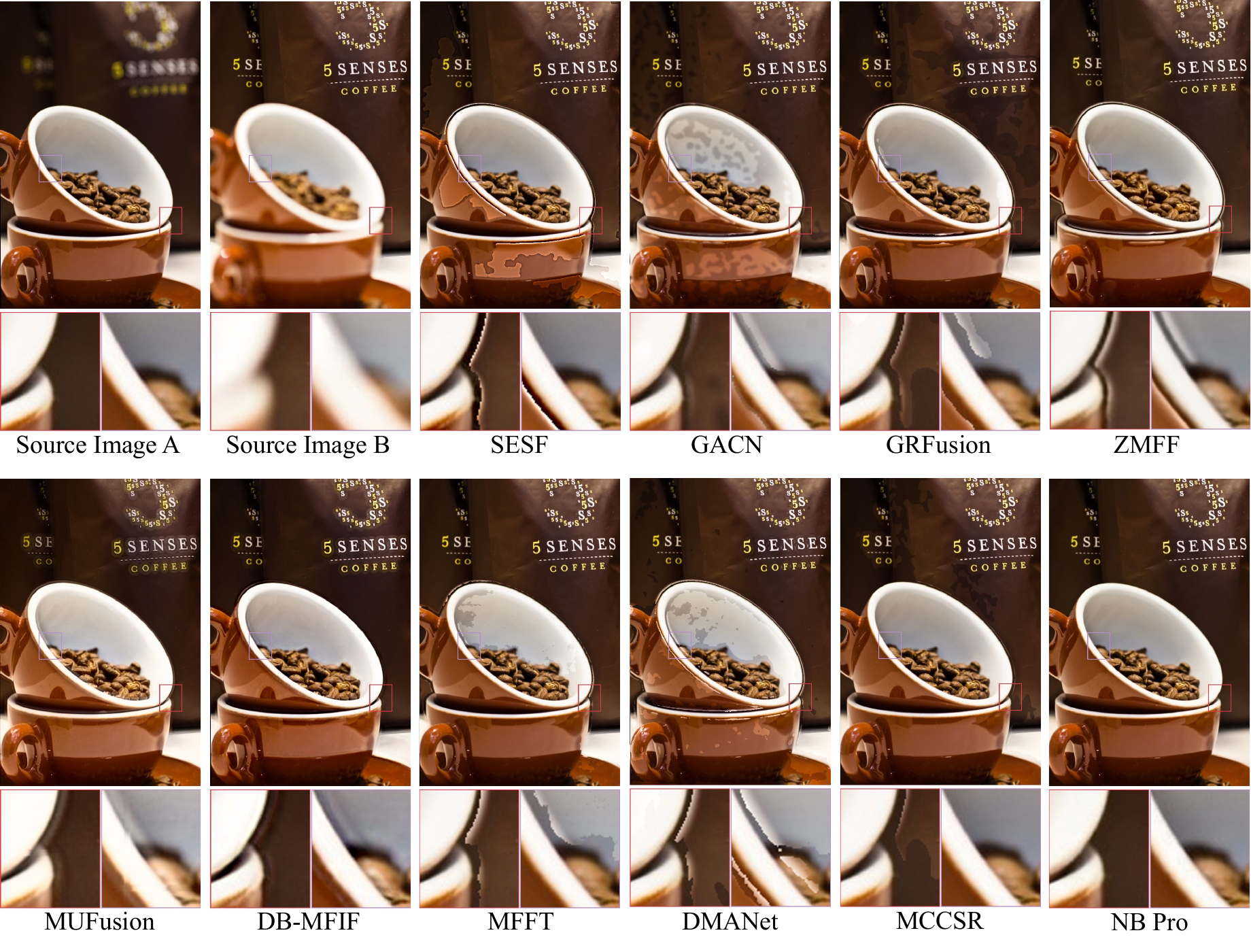}  
  \caption{Visualization of the fusion images of the `coffee cup' sample from the MFFW dataset. Two enlarged views are shown to reveal critical details.} 
  \vspace{-5pt}
  \label{mfif_figure3}  
\end{figure*}

To more intuitively verify the capability of NB Pro, Fig.~\ref{mfif_figure2} and Fig.~\ref{mfif_figure3} compares the fusion results against state-of-the-art MFIF approaches. 
In the `lock' example, NB Pro achieves remarkably smooth transitions across regions without introducing artifacts near the lock head, while producing even sharper details on the house wall that was already in focus in the source images. 
In the `coffee cup' example, due to the spread effect caused by defocus, some methods generate artifacts at the boundary between the two cup rims, and others produce dark ghosting along the cup wall. In contrast, NB Pro effectively overcomes these issues and delivers a visually pleasing fusion result.

\subsection{Analyses}
This comprehensive evaluation, conducted across four diverse benchmarks and benchmarked against ten state-of-the-art methods, establishes NB Pro as a paradigm shift in Multi-Focus Image Fusion. The results characterize a distinct trade-off between perceptual quality and signal fidelity.

Quantitative analysis reveals a significant performance divergence. NB Pro excels in non-reference metrics, generating images with exceptional clarity, textural richness, and visual appeal. Conversely, its performance on source-reference metrics uncovers a critical limitation inherent to zero-shot generative approaches: the prioritization of generative freedom over strict pixel-level adherence to source inputs. While the model can hallucinate plausible high-frequency details (e.g., enhancing lawn textures), it occasionally alters gradients or structures that require preservation.

This quantitative discrepancy stems from the fundamental conflict between the strict consistency required by traditional fusion tasks and the stochastic nature of generative models. First, despite prompt-based constraints, the model’s high degree of freedom can lead to semantic alterations in regions that should be preserved. Second, source images are rarely perfect; NB Pro often performs generative enhancement (e.g., super-resolution or denoising) to supplement details. However, traditional reference-based metrics penalize these visual improvements as errors because they deviate from the imperfect source.

Qualitatively, NB Pro demonstrates superior capability in handling complex scenarios, particularly those affected by the Defocus Spread Effect. By effectively mitigating boundary artifacts, dark ghosting, and unnatural transitions common in traditional algorithms, the model showcases a superior semantic understanding of scene structure.

Despite these strengths, instability in focus detection remains a challenge; the model occasionally blurs clear regions, suggesting failures in the attention mechanism's pixel localization. Ultimately, the misalignment between visual superiority and metric penalties suggests that current evaluation frameworks are insufficient for Generative AI. Future work must focus on better constraining the generative process and developing novel metrics capable of distinguishing between hallucinatory errors and generative enhancements.

\section{Infrared-Visible Image Fusion}
\subsection{Introduction}
In the field of modern computer vision, multi-modal image fusion technology plays an increasingly critical role. 
Infrared-Visible Image Fusion (IVIF) aims to synergize the thermal radiation information from infrared images with the texture details and color information from visible images. 
Infrared sensors capture thermal signatures of objects and are robust against varying lighting conditions and adverse weather (such as smoke or darkness), effectively highlighting targets. 
Conversely, visible light sensors provide rich high-frequency details and scene descriptions that align with human visual perception. 
By fusing these two complementary modalities, the resulting images not only possess all-weather scene perception capabilities but also significantly enhance the accuracy and robustness of target detection, autonomous driving navigation, and security surveillance systems.

Traditional IVIF methods (such as multi-scale transform~\cite{liu2015general} and sparse representation~\cite{li2019discriminative}) often struggle to balance the saliency of thermal targets with the fidelity of background textures, frequently resulting in artificial artifacts. 
In recent years, deep learning-based methods (such as CNNs~\cite{tang2022learning, ma2020infrared, zhao2021efficient}, GANs~\cite{ma2019fusiongan, ma2020ddcgan, li2020attentionfgan}) and transformers~\cite{wang2022swinfuse, vs2022image, li2022cgtf}  have achieved performance breakthroughs but still face challenges in cross-modal feature alignment and detail preservation. 
With the explosion of generative AI technologies, particularly Diffusion Models, image generation quality has reached unprecedented heights. 
However, existing high-compute models are often bulky and difficult to run in real-time on edge devices. 
Furthermore, balancing generative quality with physical fidelity in fusion tasks under specific physical constraints remains a pressing problem.

Against this backdrop, Google's newly released Nano Banana Pro has garnered widespread attention in the industry. 
Nano Banana Pro employs optimized Latent Diffusion technology and efficient attention mechanisms, specifically designed to handle high dynamic range and multi-modal inputs. 
Its uniqueness lies in its ability to understand semantic context more precisely, suggesting that in image fusion tasks, it may preserve the edge information of infrared thermal sources more effectively than its predecessors while naturally integrating visible textures.

Although Nano Banana Pro has demonstrated impressive performance in general image generation, its effectiveness in the specific scientific task of Infrared-Visible Image Fusion has not yet been systematically verified. 
This report aims to bridge this gap by comprehensively evaluating the actual performance of Nano Banana Pro in IVIF tasks through qualitative analysis (visual effects) and quantitative assessment. 
We will focus on examining its fusion quality, noise control capabilities, and inference efficiency across various lighting scenarios to assess its potential as a foundation model for next-generation image fusion.

\subsection{Quantitative Results}
\begin{table*}[ht]
  \centering
  \caption{Quantitative comparison on the MSRS, RoadScene, and M$^3$FD datasets. The best results are in \textbf{black bold}.}
  \normalsize
  \setlength{\heavyrulewidth}{1.2pt}
  \setlength{\lightrulewidth}{1pt}  
  \renewcommand{\arraystretch}{1.2}
  \resizebox{\textwidth}{!}{
  \begin{tabular}{l|cccccc|cccccc|cccccc}
    \toprule
    \multirow{2}{*}{\textbf{Method}} & \multicolumn{6}{c|}{\textbf{MSRS}} & \multicolumn{6}{c|}{\textbf{RoadScene}} & \multicolumn{6}{c}{\textbf{M$^3$FD}}\\
    \cmidrule(lr){2-7} \cmidrule(lr){8-13} \cmidrule(lr){14-19}
     & $EN$ & $SD$ & $SF$ & $AG$ & $SCD$ & $VIF$ & $EN$ & $SD$ & $SF$ & $AG$ & $SCD$ & $VIF$ & $EN$ & $SD$ & $SF$ & $AG$ & $SCD$ & $VIF$ \\
    \midrule
    SDN~\cite{zhang2021sdnet}  & 5.25 & 17.35 & 8.67 & 2.67 & 0.99 & 0.50 & 7.30 & 44.06 & 14.58 & 5.80 & 1.37 & 0.61 & 6.87 & 36.22 & 15.32 & 5.61 & 1.41 & 0.55 \\
    TarD~\cite{liu2022target} & 5.28 & 25.22 & 5.98 & 1.83 & 0.71 & 0.42 & 7.26 & 47.44 & 11.11 & 4.14 & 1.40 & 0.56 & 6.80 & 41.77 & 8.65 & 3.17 & 1.35 & 0.51 \\
    DeF~\cite{liang2022fusion}  & 6.46 & 37.63 & 8.60 & 2.80 & 1.35 & 0.77 & 7.36 & 47.03 & 10.99 & 4.38 & 1.62 & 0.63 & 6.90 & 36.81 & 9.85 & 3.65 & {1.42} & 0.58 \\
    Meta~\cite{zhao2023metafusion} & 5.65 & 24.97 & 9.99 & 3.40 & 1.14 & 0.31 & 6.88 & 31.97 & 14.38 & 5.57 & 0.92 & 0.55 & 6.73 & 30.56 & 16.48 & {6.02} & 1.31 & 0.65 \\
    CDDF~\cite{zhao2023cddfuse} & 6.70 & 43.38 & {11.56} & 3.73 & {1.62} & \textbf{1.05} & \textbf{7.52} & 54.42 & 14.97 & 5.81 & {1.65} & \textbf{0.66} & {7.04} & 42.02 & {16.56} & 5.84 & 1.41 & 0.65 \\
    LRR~\cite{li2023lrrnet}  & 6.19 & 31.78 & 8.46 & 2.63 & 0.79 & 0.54 & 7.12 & 39.16 & 11.41 & 4.37 & 1.46 & 0.45 & 6.58 & 30.28 & 11.83 & 4.21 & 1.34 & 0.54 \\
    MURF~\cite{xu2023murf} & 5.04 & 16.37 & 8.31 & 2.67 & 0.86 & 0.40 & 6.91 & 33.34 & 13.88 & 5.37 & 1.04 & 0.52 & 6.59 & 28.89 & 11.82 & 4.81 & 1.21 & 0.39 \\
    DDFM~\cite{zhao2023ddfm} & 6.19 & 29.26 & 7.44 & 2.51 & 1.45 & 0.73 & 7.24 & 42.43 & 10.68 & 4.15 & 1.64 & 0.62 & 6.82 & 32.68 & 10.07 & 3.71 & 1.35 & 0.60 \\
    SegM~\cite{liu2023multi} & 5.95 & 37.28 & 11.10 & 3.47 & 1.57 & 0.88 & 7.29 & 46.14 & 14.47 & 5.57 & 1.61 & {0.65} & 6.88 & 36.20 & 16.19 & 5.83 & 1.38 & \textbf{0.75} \\
    EMMA~\cite{zhao2024equivariant} & {6.71} & {44.13} & {11.56} & {3.76} & \textbf{1.63} & {0.97} & \textbf{7.52} & {54.81} & {15.21} & {5.83} & \textbf{1.69} & \textbf{0.66} & \textbf{7.12} & \textbf{44.01} & \textbf{16.92} & \textbf{6.23} & \textbf{1.48} & {0.66} \\
    
    \rowcolor{lightgray!50}
    \textbf{\textcolor{orange}{NB Pro}} & 
    \textbf{6.85} & 
    \textbf{44.95} & 
    \textbf{14.39} & 
    \textbf{4.56} & 
    \textbf{\textcolor{orange}{1.15}} & \textbf{\textcolor{orange}{0.58}} & \textbf{\textcolor{orange}{7.39}} & 
    \textbf{56.98} & 
    \textbf{21.81} & 
    \textbf{7.07} & 
    \textbf{\textcolor{orange}{0.83}} & \textbf{\textcolor{orange}{0.51}} & \textbf{\textcolor{orange}{6.98}} & \textbf{\textcolor{orange}{43.44}} & \textbf{\textcolor{orange}{15.68}} & \textbf{\textcolor{orange}{5.06}} & \textbf{\textcolor{orange}{0.75}} & \textbf{\textcolor{orange}{0.38}} \\
    \bottomrule
  \end{tabular}
  }
  \label{ivif_table1}
\end{table*}
\begin{figure*}[ht]  
  \centering  
  \includegraphics[width=1.0\textwidth]{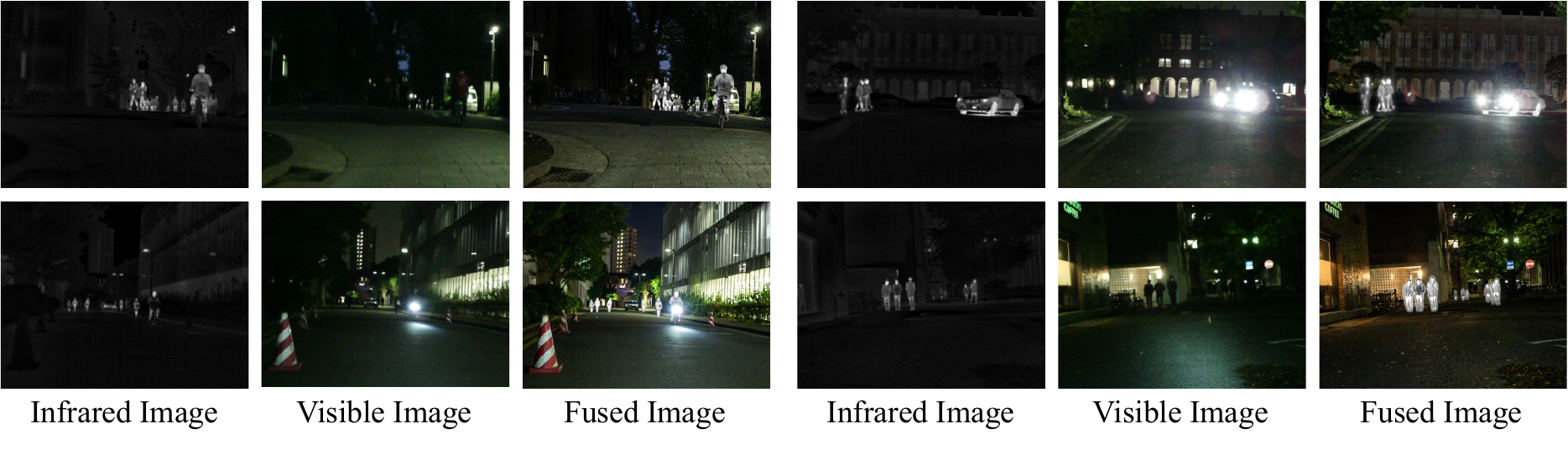}  
  \caption{Visualization examples of image fusion results by Nano Banana Pro in the MSRS dataset.} 
  \label{ivif_figure1}  
\end{figure*}

Following~\cite{zhao2024equivariant}, we conduct experiments on three mainstream benchmarks: MSRS~\cite{tang2022piafusion}, RoadScene~\cite{xu2020fusiondn} and M$^3$FD~\cite{liu2022target} datasets, and leverage six popular metrics for assignment, including non-reference metrics $EN$, $SD$, $SF$, $AG$ and source-reference metrics $SCD$, $VIF$. We compare the fusion results of Nano Banana Pro (NB Pro) with 10 other state-of-the-art and representative IVIF methods, where SDNet and DeFusion are CNN-based methods, TarDAL is a GAN-based method, CDDFuse is a CNN-transfomer hybrid method, DDFM is a diffusion-based and training-free method, and the remaining ones are model or task driven approaches.

As shown in Tab.~\ref{ivif_table1},  the results demonstrate that NB Pro exhibits overwhelming superiority in non-reference metrics representing image information content and texture details, particularly on the MSRS and RoadScene datasets.
On the MSRS dataset, NB Pro secures the top rank in all four of these metrics. Notably, its $EN$ reaches 6.85 and $AG$ hits 4.56, significantly surpassing the runner-up. On the RoadScene dataset, this advantage is even more pronounced. NB Pro achieves an $SF$ score of 21.81, outperforming the nearest competitor (EMMA, 15.21) by nearly 43\%.
This indicates that the fused images generated by NB Pro possess extremely high clarity and contrast. It is capable of mining and reconstructing rich high-frequency edge information from source images, attributing to the acute capability of its powerful generative architecture in capturing latent features.

However, we also observe an intriguing phenomenon: while NB Pro leads by a wide margin in detail metrics, it scores relatively lower in source-reference metrics, specifically $SCD$ and $VIF$. For instance, on the  MSRS dataset, its $VIF$ is only 0.58, and and it drops to 0.38 on M$^3$FD.
This reflects the inherent characteristic of generative models: while the model dramatically enhances texture and human-perceived sharpness, this reconstruction process may introduce stylized features or pixel-level deviations not present in the source images, leading to reduced correlation with the original infrared/visible inputs. 
In contrast, traditional methods, while not as sharp as NB Pro, demonstrate more robustness in maintaining original pixel fidelity.

NB Pro's performance varies across different scenarios. It performs best on MSRS (often containing night-time and complex lighting scenes) and RoadScene, suggesting its proficiency in handling high dynamic range scenes requiring edge enhancement. 
On the M$^3$FD dataset, although it maintains the second-best score in $SD$(43.44), its overall dominance is less distinct compared to the other two datasets. 
This implies there may still be room for parameter fine-tuning in specific types of multi-modal target detection scenarios.
\begin{figure*}[ht]  
  \centering  
  \includegraphics[width=1.0\textwidth]{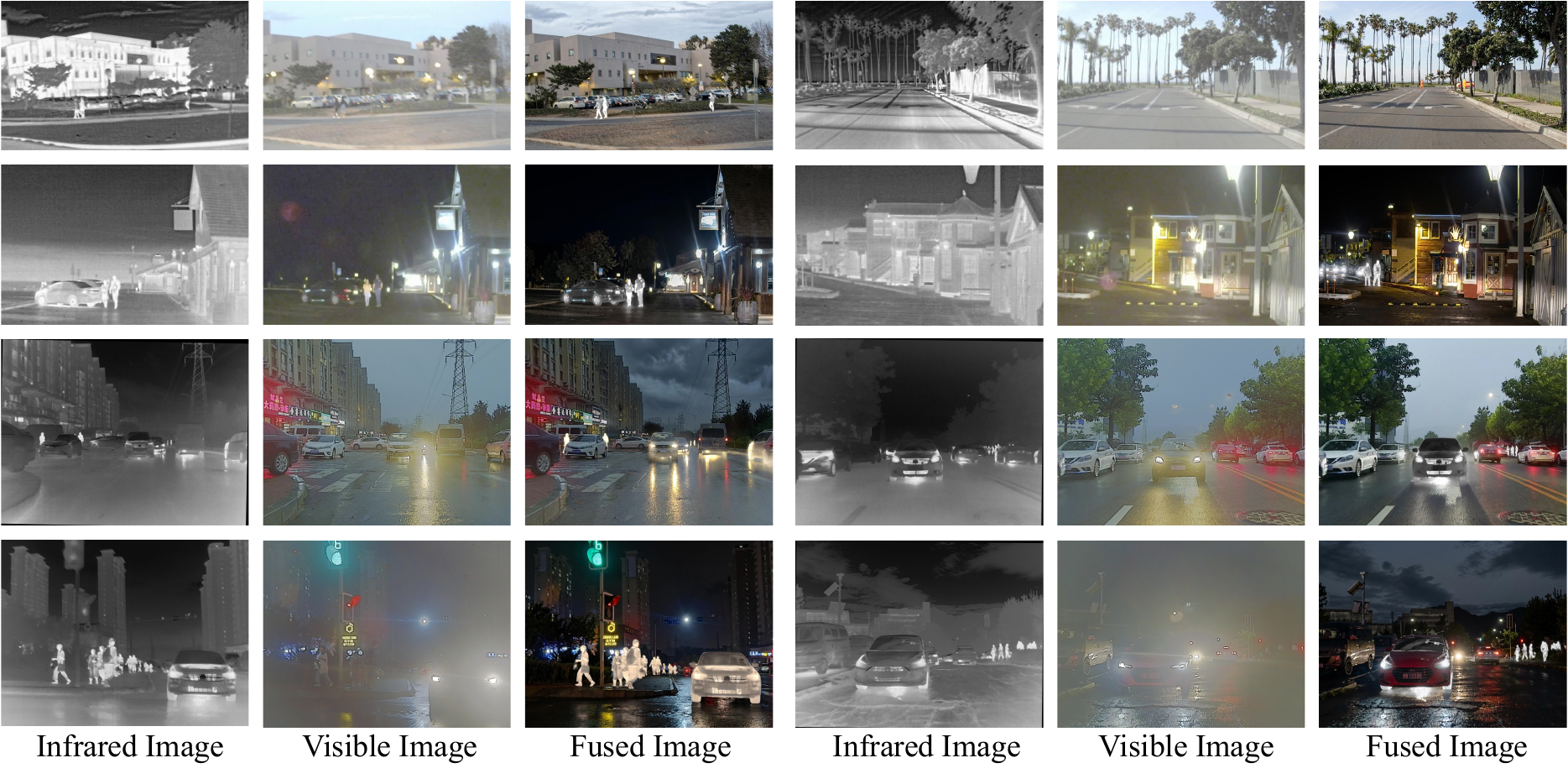}  
  \caption{Visualization examples of image fusion results by Nano Banana Pro in the RoadScene and M$^3$FD datasets.} 
  \vspace{-3pt}
  \label{ivif_figure2}  
\end{figure*}

\subsection{Qualitative Results}
While NB Pro yields impressive fusion results in most scenarios, certain limitations persist. 
To intuitively demonstrate the perceptual quality of the fused images, Fig.~\ref{ivif_figure1} presents qualitative results on the MSRS dataset. 
As illustrated in the first row, the method effectively handles extreme lighting conditions. 
It accurately captures pedestrian targets hidden in low-light regions of the visible spectrum, establishing sharp contours for thermal targets while enhancing overall background illumination. 
Simultaneously, it restores clear details and textures in over-exposed areas, such as those adjacent to vehicle headlights.
However, suboptimal cases are observed in the second row. 
Although the thermal targets remain highlighted, the first sample exhibits excessive sharpening of the building structure, leading to slight over-exposure. 
In the second example, a distinct halo effect emerges around the pedestrian, manifesting as unnatural bright fringes surrounding the thermal target.


Fig.~\ref{ivif_figure2} further visualize the fusion performance on the other two datasets. 
In general, NB Pro demonstrates superior performance in terms of overall visual quality and realism, excelling in infrared target recovery and background texture reconstruction. 
Nevertheless, deficiencies persist in certain fine-grained details. The model may occasionally fail to faithfully utilize and preserve the source information, leading to the introduction of unnatural hallucinations or artifacts.

\subsection{Analyses}

This report presents a comprehensive evaluation of Google's Nano Banana Pro in Infrared-Visible Image Fusion. The experimental outcomes reveal a pronounced performance dichotomy: 

On one hand, leveraging powerful generative priors, NB Pro demonstrates overwhelming superiority in non-reference metrics. It successfully circumvents the bottlenecks of traditional methods regarding night-time enhancement and texture reconstruction, yielding fused images of exceptional contrast and clarity. On the other hand, this aggressive generation strategy incurs a fidelity cost. Lower scores in source consistency metrics, combined with qualitative artifacts such as excessive sharpening and halo effects, indicate that the model sacrifices pixel-level fidelity to the original physical signals in exchange for perceptual appeal.

The performance of NB Pro catalyzes a re-evaluation of the current IVIF landscape:

Traditional methods fundamentally operate as signal processing routines aiming to preserve pixel intensity. However, NB Pro introduces a paradigm of semantic generation. Rather than merely superimposing pixels, it interprets the scene context to re-synthesize the image. This explains its capability to recover astonishing details alongside its propensity for hallucinations. Future research must focus on integrating physical constraints, enforcing strict adherence to thermal distribution laws while exploiting generative capabilities.

While visually striking, NB Pro's outputs raise concerns for safety-critical applications like autonomous driving. Perceptual pleasantness does not equate to operational reliability. Artifacts or over-sharpening can trigger false positives in detection algorithms or obscure small targets. Consequently, evaluation standards must evolve beyond visual quality to include Machine Perception Metrics, directly validating the utility of fused images on downstream tasks.

Our findings highlight the insufficiency of the current evaluation framework. Metrics like SCD and VIF, which penalize pixel-level misalignment, are overly rigid for generative models. As Generative AI becomes prevalent, there is an imperative need for novel No-Reference Image Quality Assessment metrics that prioritize semantic consistency and naturalness over strict pixel-wise alignment.
\newpage

\section{Discussion}


This comprehensive empirical study, through systematic zero-shot evaluation across 14 diverse low-level vision tasks, elucidates the dual nature of Nano Banana Pro as a generalist generative model: \textbf{it excels in perceptual quality but lags significantly in traditional pixel-fidelity metrics.} This core finding not only quantifies the current capabilities of large-scale generative models within the low-level vision domain but also prompts profound reflections on task definitions, evaluation paradigms, and future model evolution.

\subsection{Generative vs. Regression Paradigms}
Our work highlights intrinsic difference between generative and traditional low-level vision models. Traditional methods predominantly follow a regression paradigm. They learn a deterministic mapping from degraded inputs to clean references via pixel-level supervision, aligning their optimization objective directly with metrics like PSNR and SSIM. In contrast, Nano Banana Pro embodies a generative paradigm. Its training core involves learning the joint distribution of large-scale image data and performing conditional synthesis based on semantic priors. Its goal is to produce plausible and visually pleasing images, not to achieve pixel-wise alignment with a specific reference. Consequently, in regions with severe information loss, traditional methods, constrained by input information, often yield blurry or insipid results. The generative model, however, can leverage its robust world knowledge to hallucinate plausible details, leading to subjectively superior outputs that constitute a deviation from the canonical ground truth.

\subsection{The  Potential Misguidance of Traditional Metrics}

Our results strongly challenge the universal applicability of full-reference metrics (e.g., PSNR, SSIM), which are predominant in current low-level vision research. These pixel-difference-based metrics carry a strong implicit assumption: the existence of a single, pixel-perfect ground truth. This assumption is problematic for evaluating generative solutions:

\textbf{Ground truth is not the unique optimum for generative repair.} For regions with catastrophic information loss, multiple visually plausible and contextually correct reconstructions may exist. A generative model provides one such possibility, yet it is penalized by the metric as incorrect.

\textbf{The metrics are misaligned with human perception.} As shown in previous sections, Nano Banana Pro achieves excellent scores on No-Reference perceptual metrics (e.g., NIQE, NIMA), often surpassing specialized models. This indicates its outputs possess superior statistical naturalness and aesthetic appeal. The drop in PSNR can sometimes be attributed solely to the model's reasonable global color adjustment, mild denoising, or detail enhancement, which are improvements that are paradoxically penalized.

\textbf{The quality of dataset ground truth itself.} Ground truth images in many real-world datasets contain residual noise, slight blur, or imperfect color balance. A generative model producing a cleaner version constitutes a perceptual enhancement but is scored as a fidelity loss.

Therefore, \textbf{judging generative low-level vision models solely by traditional metrics may be both unfair and misleading.} This calls for the community to establish a new generation of evaluation frameworks.

\subsection{Operational Scope and Limitations of Nano Banana Pro}

Our evaluation clearly defines the scope of Nano Banana Pro, shaped by a fundamental compromise: \textbf{it favors semantic plausibility and visual appeal over precise pixel-level fidelity.} This positions the model as highly effective for creative and perceptual tasks, such as artistic image enhancement, restoration of severely degraded photos, and scenarios where a visually compelling result is more critical than strict accuracy. Its ability to perform these tasks without specialized training also makes it a practical tool for rapid prototyping.

However, these capabilities come with inherent constraints. The model is not suitable for applications demanding rigorous factual accuracy, including forensic examination, scientific imaging, or any context where the output must correspond exactly to the original scene data. Its generative approach can introduce alterations, such as softened boundaries in super-resolution, altered text in deblurring, or non-physical color shifts, that prioritize visual completeness over authentic reproduction. In essence, \textbf{Nano Banana Pro serves as a powerful semantic reconstructor and enhancer for common visual applications, but it is not designed for high-precision tasks where strict fidelity is paramount.}

\subsection{Future Research Directions}

The findings of this study point to several critical directions for future work:

\textbf{Exploration of Hybrid Architectures.} The future all-rounder may not be a purely generative model but a generative-regression hybrid. For instance, a lightweight regression network could first recover basic structure and color in the front-end, followed by a conditional generative model for detail enhancement and beautification in the back-end. This process should be constrained by physics-informed loss functions to curb arbitrariness.

\textbf{Prompt Engineering and Controllable Generation.} It is important to note that the present evaluation reflects a conservative estimate of the model’s capability, as we did not engage in meticulous prompt tuning or employ multi-round inference to cherry-pick optimal outputs. Our fixed, simple prompts represent a pragmatic but unoptimized use case. Future work should therefore systematically explore how carefully designed textual instructions, visual cues, or interactive refinement can more effectively steer the generative process. Enhancing such controllability will be key to reducing unwanted variability in color, structure, and texture—ultimately improving the reliability and practical utility of generative models in restoration-sensitive applications.

\textbf{Innovation in Evaluation Frameworks for Generative Models.} The rise of generative models calls for a fundamental shift in how we evaluate their output. Traditional metrics, which rely on a single ground truth, fall short when assessing models that can produce multiple plausible reconstructions from a degraded image. We urgently need new benchmarks that reflect this reality, for instance, datasets that include several expert-approved restoration options for a given input. At the same time, evaluation should move beyond pixel-level fidelity alone. Developing unified metrics that capture both perceptual quality and distortion would provide a more nuanced view of a model’s performance across the quality–fidelity spectrum.

\subsection{Conclusion}
The evaluation of Nano Banana Pro signals a paradigm shift: foundational generative models are redefining the boundaries of low-level vision. Functioning less as a traditional restoration tool and more as a semantic reconstruction engine, the model leverages deep generative priors to synthesize visual content rather than merely recovering pixels. This emergence challenges the community to reconsider the fundamental metric of success: is it absolute pixel fidelity, or the maximization of perceptual plausibility?.

Our empirical results confirm that while Nano Banana Pro trails domain-specific experts in zero-shot pixel fidelity, it demonstrates exceptional potential in perceptual quality, particularly when handling extreme degradations and cross-task generalization. Consequently, the trajectory of the field lies not in a binary choice between paradigms, but in strategic integration. The next generation of robust vision systems must bridge the semantic imagination of generative models with the physical constraints and precision of specialized networks.

Ultimately, Nano Banana Pro is a double-edged sword. It has successfully raised the ceiling of perceptual quality for complex visual tasks, yet it has not secured the floor of stability required for forensic precision.

\newpage

\section{Appendix}
\subsection{Prompts for Each Task}
\textbf{Dehazing:} \textit{``This is an image with hazy, which reduces clarity and contrast. Please completely remove this atmospheric haze from the image while meticulously preserving all other elements without any alteration. The post-processing must be strictly limited to haze removal only. Ensure that every other aspect of the image remains entirely unchanged, including but not limited to the original composition, all subjects and objects within the scene. The final output should be a clear, natural-looking version of the original image with its core content perfectly intact.''}

\textbf{Super-Resolution:} \textit{``Please upscale this low-resolution image to a high-quality 1k (1024x1024) resolution. Perform super-resolution processing to significantly enhance clarity, remove pixelation, and sharpen textures, while strictly preserving the original content, composition, and colors unchanged. Do not alter the subject's features or hallucinate new elements.''}

\textbf{Deraining:} \textit{``This is a rainy image. Please remove the rain streaks and raindrops while keeping all other elements, the original color tone, lighting, and atmosphere unchanged.''}

\textbf{Shadow Removal:} \textit{``This is an image with prominent cast shadows. Please remove these shadows from the image while ensuring all other elements in the scene remain completely unchanged and consistent. The core requirement is to eliminate the shadow effects while preserving the inherent properties of all objects and the background. The final output should be a clean, evenly lit version of the original image, without any dark occlusions or shaded areas.''}

\textbf{Motion Deblurring:} \textit{``This is an image with blurring and lack of clarity due to motion. Transform this image into a sharp, static photograph. Please carefully observe this picture to eliminate motion streaks and recover details in any blurred areas—regardless of the motion source—while keeping originally sharp elements consistent. CRITICAL EXPOSURE LOCK: Strictly prohibit any form of exposure correction, HDR tone mapping, or lighting enhancement. Do not attempt to recover details in blown-out highlights or brighten dark/underexposed shadows. You must preserve the exact luminance levels of the original image. Zero changes to brightness, contrast, or exposure are allowed.''}

\textbf{Defocus Deblurring:} \textit{``This is an image with partial blurring and lack of clarity due to camera defocus. Transform this image into an all-in-focus photograph. Please carefully observe this picture to enhance the sharpness of the blurred areas while keeping all other elements consistent. Maintain the original color palette, lighting, and exposure faithfully. No color shifts or brightness changes.''}

\textbf{Denoising:} \textit{``This is a noisy image, please remove the noise in this image while keep other elements in this image unchanged.''}

\textbf{Reflection Removal:} \textit{``Strictly remove only the glass reflections and glare overlaying the scene. Do not alter the underlying scene composition, object details, geometry, or color grading in any way. The objective is to make the glass invisible while maintaining pixel-perfect fidelity to the original background. Zero tolerance for hallucinations, artistic interpretation, or style changes. Retain the exact original perspective and lighting conditions of the scene behind the glass. Treat this as a forensic image restoration.''}

\textbf{Flare Removal:} \textit{``Identify and remove lens flare and glare artifacts generated by multiple distinct light sources throughout the entire image.  For every single light source causing flare, eliminate the optical interference (such as ghosting, halos, and streaks) while strictly maintaining the natural illumination and intensity of each individual light.  Seamlessly restore the obscured background textures behind all flare instances to ensure global consistency.''}

\textbf{Low Light Image Enhancement:} \textit{``This is a low-light image, please turn this image into a normal image while keeping other elements unchanged.''}

\textbf{Underwater Image Enhancement:} \textit{``This is an underwater image with obvious color cast, low contrast, and scattered fog. Please carefully analyze the main scene, eliminate the interference caused by water absorption and suspended particles, and restore a clear, fog-free version with true colors. Ensure that all elements except for the degradation factors are consistent with the real underwater environment, without introducing new artifacts or over-enhancement.CRITICAL ELEMENT LOCK: remove only the color shift and fog; do not add, remove, or alter any original object, edge, texture.''}

\textbf{HDR Imaging:} \textit{``This is an image suffering from limited dynamic range, with lost details in highlights and shadows. Transform this image into a high dynamic range photograph. Please carefully analyze this picture to expand its dynamic range, recover details in both clipped highlights and underexposed shadows, while preserving the integrity of properly exposed areas.CRITICAL CONTENT LOCK:Strictly prohibit any form of scene content alteration, object addition, or deletion. Do not change the shape, position, texture, or color relationships of any original elements. Forbid the introduction of any artistic styles, filter effects, or non-physical halos.CRITICAL PROCESSING PRINCIPLES:1. Highlight Recovery: Intelligently reconstruct plausible texture and detail lost in overexposed areas (e.g., sky, windows, light sources), but do not alter their fundamental form or color.2. Shadow Enhancement: Selectively lift the brightness of shadow areas to reveal concealed details, while must retaining the depth and original distribution of the shadows.3. Midtone Preservation: Maintain the natural contrast and color of midtone areas. Avoid introducing an overall flat/gray look or unnatural local contrast.FINAL OBJECTIVE: Produce a detail-rich, balanced, and natural-looking image that appears as if the same scene was captured in a single shot with professional HDR equipment, not artificially composited.''}

\textbf{Multi-Focus Image Fusion:} \textit{``Act as an advanced Computer Vision expert specialized in Multi-Focus Image Fusion(MFIF). 
Your task is to process a pair of source images with different focal depths (e.g., near-focus and far-focus) and merge them into a single, high-quality 'all-in-focus' image. You must analyze the local sharpness and high-frequency details of each input to accurately identify the clearest regions, constructing a precise decision map that selects the best pixels from the respective sources. Ensure to apply boundary refinement techniques to guarantee smooth transitions between fused areas. 
The final output must be a seamless, fully focused image that strictly preserves the original color and structural fidelity while being completely free of visual artifacts such as ghosting, halos, or unnatural stitching seams.''}

\textbf{Infrared-Visible Image Fusion:} \textit{``Act as an expert in Infrared and Visible Image Fusion (IVIF). Your task is to generate a single high-quality fused image based on the provided Infrared (IR) and Visible (VIS) source images. The core objective is to integrate complementary features: you must preserve the high-intensity thermal saliency (such as pedestrians and vehicles) from the IR image to ensure target detectability, while simultaneously injecting the high-frequency textural details (such as edges, vegetation, and building structures) from the VIS image to ensure background clarity. The fusion process must balance the intensity distribution to look perceptually natural, strictly minimizing artifacts like ghosting, halos, or noise. 
The final output should be a sharp, noise-free image that combines the high contrast of the thermal targets with the rich structural details of the visible scene.''}

\subsection{Contributors}
\textbf{Jialong Zuo:} Project Leader. \textbf{Haoyou Deng:} Document Polish.

\textbf{Hanyu Zhou:} Denoising and Low Light Enhancement. \textbf{Jiaxin Zhu:} Motion Deblurring and Defocus Deblurring.

\textbf{Yicheng Zhang:} Multi-focus Image Fusion and Infrared-Visible Image Fusion. 

\textbf{Yiwei Zhang:} Underwater Image Enhancement and HDR Imaging.

\textbf{Yongxin Yan:} Dehazing and Shadow Removal. \textbf{Kaixing Huang:} Super Resolution.

\textbf{Weisen Chen:} Deraining. \textbf{Yongtai Deng:} Reflection Removal. \textbf{Rui Jin:} Flare Removal.

\textbf{Nong Sang:} Advisor. \textbf{Changxin Gao:} Advisor.

\clearpage
\newpage

\bibliographystyle{plainnat}
\bibliography{main}

@inproceedings{rombach2022high,
    title={High-resolution image synthesis with latent diffusion models},
    author={Rombach, Robin and Blattmann, Andreas and Lorenz, Dominik and Esser, Patrick and Ommer, Bj{\"o}rn},
    booktitle={IEEE/CVF Conference on Computer Vision and Pattern Recognition},
    pages={10684-10695},
    year={2022}
}

@inproceedings{ke2021musiq,
    title = {{MUSIQ}: Multi-scale image quality transformer},
    author = {Ke, Junjie and Wang, Qifei and Wang, Yilin and Milanfar, Peyman and Yang, Feng},
    booktitle = {IEEE/CVF International Conference on Computer Vision},
    pages = {5148-5157},
    year = 2021
}

@article{liu2017multi,
  title={Multi-focus image fusion with a deep convolutional neural network},
  author={Liu, Yu and Chen, Xun and Peng, Hu and Wang, Zengfu},
  journal={Information Fusion},
  volume={36},
  pages={191--207},
  year={2017},
  publisher={Elsevier}
}

@inproceedings{xiao2021dtmnet,
  title={DTMNet: A discrete Tchebichef moments-based deep neural network for multi-focus image fusion},
  author={Xiao, Bin and Wu, Haifeng and Bi, Xiuli},
  booktitle={Proceedings of the IEEE/CVF International Conference on Computer Vision},
  pages={43--51},
  year={2021}
}

@article{zhang2020ifcnn,
  title={IFCNN: A general image fusion framework based on convolutional neural network},
  author={Zhang, Yu and Liu, Yu and Sun, Peng and Yan, Han and Zhao, Xiaolin and Zhang, Li},
  journal={Information Fusion},
  volume={54},
  pages={99--118},
  year={2020},
  publisher={Elsevier}
}

@article{ma2022swinfusion,
  title={SwinFusion: Cross-domain long-range learning for general image fusion via swin transformer},
  author={Ma, Jiayi and Tang, Linfeng and Fan, Fan and Huang, Jun and Mei, Xiaoguang and Ma, Yong},
  journal={IEEE/CAA Journal of Automatica Sinica},
  volume={9},
  number={7},
  pages={1200--1217},
  year={2022},
  publisher={IEEE}
}

@article{li2024fusiondiff,
  title={FusionDiff: Multi-focus image fusion using denoising diffusion probabilistic models},
  author={Li, Mining and Pei, Ronghao and Zheng, Tianyou and Zhang, Yang and Fu, Weiwei},
  journal={Expert Systems with Applications},
  volume={238},
  pages={121664},
  year={2024},
  publisher={Elsevier}
}

@article{li2020drpl,
  title={DRPL: Deep regression pair learning for multi-focus image fusion},
  author={Li, Jinxing and Guo, Xiaobao and Lu, Guangming and Zhang, Bob and Xu, Yong and Wu, Feng and Zhang, David},
  journal={IEEE Transactions on Image Processing},
  volume={29},
  pages={4816--4831},
  year={2020},
  publisher={IEEE}
}

@article{nejati2015multi,
  title={Multi-focus image fusion using dictionary-based sparse representation},
  author={Nejati, Mansour and Samavi, Shadrokh and Shirani, Shahram},
  journal={Information fusion},
  volume={25},
  pages={72--84},
  year={2015},
  publisher={Elsevier}
}

@article{xu2020mffw,
  title={MFFW: A new dataset for multi-focus image fusion},
  author={Xu, Shuang and Wei, Xiaoli and Zhang, Chunxia and Liu, Junmin and Zhang, Jiangshe},
  journal={arXiv preprint arXiv:2002.04780},
  year={2020}
}

@article{zhang2021mff,
  title={MFF-GAN: An unsupervised generative adversarial network with adaptive and gradient joint constraints for multi-focus image fusion},
  author={Zhang, Hao and Le, Zhuliang and Shao, Zhenfeng and Xu, Han and Ma, Jiayi},
  journal={Information Fusion},
  volume={66},
  pages={40--53},
  year={2021},
  publisher={Elsevier}
}

@web{Chun2025multi,
  title={Standard images for multifocus image fusion},
  author={Chun-Chieh Tsai},
  journal={MATLAB Central File Exchange},
  year={2025},
  publisher={MATLAB}
}

@book{jahne2005digital,
  title={Digital image processing},
  author={J{\"a}hne, Bernd},
  year={2005},
  publisher={Springer}
}

@article{cui2015detail,
  title={Detail preserved fusion of visible and infrared images using regional saliency extraction and multi-scale image decomposition},
  author={Cui, Guangmang and Feng, Huajun and Xu, Zhihai and Li, Qi and Chen, Yueting},
  journal={Optics Communications},
  volume={341},
  pages={199--209},
  year={2015},
  publisher={Elsevier}
}

@article{zheng2007new,
  title={A new metric based on extended spatial frequency and its application to DWT based fusion algorithms},
  author={Zheng, Yufeng and Essock, Edward A and Hansen, Bruce C and Haun, Andrew M},
  journal={Information Fusion},
  volume={8},
  number={2},
  pages={177--192},
  year={2007},
  publisher={Elsevier}
}

@article{hossny2008comments,
  title={Comments on ‘Information measure for performance of image fusion’},
  author={Hossny, Mohammed and Nahavandi, Saeid and Creighton, Douglas},
  journal={Electronics letters},
  volume={44},
  number={18},
  pages={1066--1067},
  year={2008},
  publisher={IET}
}

@article{yang2008novel,
  title={A novel similarity based quality metric for image fusion},
  author={Yang, Cui and Zhang, Jian-Qi and Wang, Xiao-Rui and Liu, Xin},
  journal={Information Fusion},
  volume={9},
  number={2},
  pages={156--160},
  year={2008},
  publisher={Elsevier}
}

@article{chen2009new,
  title={A new automated quality assessment algorithm for image fusion},
  author={Chen, Yin and Blum, Rick S},
  journal={Image and vision computing},
  volume={27},
  number={10},
  pages={1421--1432},
  year={2009},
  publisher={Elsevier}
}

@article{ma2021sesf,
  title={Sesf-fuse: An unsupervised deep model for multi-focus image fusion},
  author={Ma, Boyuan and Zhu, Yu and Yin, Xiang and Ban, Xiaojuan and Huang, Haiyou and Mukeshimana, Michele},
  journal={Neural Computing and Applications},
  volume={33},
  number={11},
  pages={5793--5804},
  year={2021},
  publisher={Springer}
}

@article{ma2022end,
  title={End-to-end learning for simultaneously generating decision map and multi-focus image fusion result},
  author={Ma, Boyuan and Yin, Xiang and Wu, Di and Shen, Haokai and Ban, Xiaojuan and Wang, Yu},
  journal={Neurocomputing},
  volume={470},
  pages={204--216},
  year={2022},
  publisher={Elsevier}
}

@article{li2023generation,
  title={Generation and recombination for multifocus image fusion with free number of inputs},
  author={Li, Huafeng and Wang, Dan and Huang, Yuxin and Zhang, Yafei and Yu, Zhengtao},
  journal={IEEE Transactions on Circuits and Systems for Video Technology},
  volume={34},
  number={7},
  pages={6009--6023},
  year={2023},
  publisher={IEEE}
}

@article{cheng2023mufusion,
  title={MUFusion: A general unsupervised image fusion network based on memory unit},
  author={Cheng, Chunyang and Xu, Tianyang and Wu, Xiao-Jun},
  journal={Information Fusion},
  volume={92},
  pages={80--92},
  year={2023},
  publisher={Elsevier}
}

@article{zhai2024multi,
  title={Multi-focus image fusion via interactive transformer and asymmetric soft sharing},
  author={Zhai, Hao and Zheng, Wenyi and Ouyang, Yuncan and Pan, Xin and Zhang, Wanli},
  journal={Engineering Applications of Artificial Intelligence},
  volume={133},
  pages={107967},
  year={2024},
  publisher={Elsevier}
}

@article{zheng2025unfolding,
  title={Unfolding coupled convolutional sparse representation for multi-focus image fusion},
  author={Zheng, Kecheng and Cheng, Juan and Liu, Yu},
  journal={Information Fusion},
  volume={118},
  pages={102974},
  year={2025},
  publisher={Elsevier}
}

@article{zhang2024exploit,
  title={Exploit the best of both end-to-end and map-based methods for multi-focus image fusion},
  author={Zhang, Juncheng and Liao, Qingmin and Ma, Haoyu and Xue, Jing-Hao and Yang, Wenming and Liu, Shaojun},
  journal={IEEE Transactions on Multimedia},
  volume={26},
  pages={6411--6423},
  year={2024},
  publisher={IEEE}
}

@inproceedings{quan2025multi,
  title={Multi-Focus Image Fusion via Explicit Defocus Blur Modelling},
  author={Quan, Yuhui and Wan, Xi and Tang, Zitao and Liang, Jinxiu and Ji, Hui},
  booktitle={Proceedings of the AAAI Conference on Artificial Intelligence},
  volume={39},
  number={6},
  pages={6657--6665},
  year={2025}
}

@article{hu2023zmff,
  title={ZMFF: Zero-shot multi-focus image fusion},
  author={Hu, Xingyu and Jiang, Junjun and Liu, Xianming and Ma, Jiayi},
  journal={Information Fusion},
  volume={92},
  pages={127--138},
  year={2023},
  publisher={Elsevier}
}

@article{liu2015general,
  title={A general framework for image fusion based on multi-scale transform and sparse representation},
  author={Liu, Yu and Liu, Shuping and Wang, Zengfu},
  journal={Information fusion},
  volume={24},
  pages={147--164},
  year={2015},
  publisher={Elsevier}
}

@article{li2019discriminative,
  title={Discriminative dictionary learning-based multiple component decomposition for detail-preserving noisy image fusion},
  author={Li, Huafeng and Wang, Yitang and Yang, Zhao and Wang, Ruxin and Li, Xiang and Tao, Dapeng},
  journal={IEEE Transactions on Instrumentation and Measurement},
  volume={69},
  number={4},
  pages={1082--1102},
  year={2019},
  publisher={IEEE}
}

@article{tang2022learning,
  title={Learning attention-guided pyramidal features for few-shot fine-grained recognition},
  author={Tang, Hao and Yuan, Chengcheng and Li, Zechao and Tang, Jinhui},
  journal={Pattern Recognition},
  volume={130},
  pages={108792},
  year={2022},
  publisher={Elsevier}
}

@article{zhao2021efficient,
  title={Efficient and model-based infrared and visible image fusion via algorithm unrolling},
  author={Zhao, Zixiang and Xu, Shuang and Zhang, Jiangshe and Liang, Chengyang and Zhang, Chunxia and Liu, Junmin},
  journal={IEEE Transactions on Circuits and Systems for Video Technology},
  volume={32},
  number={3},
  pages={1186--1196},
  year={2021},
  publisher={IEEE}
}

@article{zhang2021sdnet,
  title={SDNet: A versatile squeeze-and-decomposition network for real-time image fusion},
  author={Zhang, Hao and Ma, Jiayi},
  journal={International Journal of Computer Vision},
  volume={129},
  number={10},
  pages={2761--2785},
  year={2021},
  publisher={Springer}
}

@article{ma2019fusiongan,
  title={FusionGAN: A generative adversarial network for infrared and visible image fusion},
  author={Ma, Jiayi and Yu, Wei and Liang, Pengwei and Li, Chang and Jiang, Junjun},
  journal={Information fusion},
  volume={48},
  pages={11--26},
  year={2019},
  publisher={Elsevier}
}

@article{ma2020ddcgan,
  title={DDcGAN: A dual-discriminator conditional generative adversarial network for multi-resolution image fusion},
  author={Ma, Jiayi and Xu, Han and Jiang, Junjun and Mei, Xiaoguang and Zhang, Xiao-Ping},
  journal={IEEE Transactions on Image Processing},
  volume={29},
  pages={4980--4995},
  year={2020},
  publisher={IEEE}
}

@article{li2020attentionfgan,
  title={AttentionFGAN: Infrared and visible image fusion using attention-based generative adversarial networks},
  author={Li, Jing and Huo, Hongtao and Li, Chang and Wang, Renhua and Feng, Qi},
  journal={IEEE Transactions on Multimedia},
  volume={23},
  pages={1383--1396},
  year={2020},
  publisher={IEEE}
}

@article{wang2022swinfuse,
  title={SwinFuse: A residual swin transformer fusion network for infrared and visible images},
  author={Wang, Zhishe and Chen, Yanlin and Shao, Wenyu and Li, Hui and Zhang, Lei},
  journal={IEEE Transactions on Instrumentation and Measurement},
  volume={71},
  pages={1--12},
  year={2022},
  publisher={IEEE}
}

@inproceedings{vs2022image,
  title={Image fusion transformer},
  author={Vs, Vibashan and Valanarasu, Jeya Maria Jose and Oza, Poojan and Patel, Vishal M},
  booktitle={2022 IEEE International conference on image processing (ICIP)},
  pages={3566--3570},
  year={2022},
  organization={IEEE}
}

@article{li2022cgtf,
  title={CGTF: Convolution-guided transformer for infrared and visible image fusion},
  author={Li, Jing and Zhu, Jianming and Li, Chang and Chen, Xun and Yang, Bin},
  journal={IEEE Transactions on Instrumentation and Measurement},
  volume={71},
  pages={1--14},
  year={2022},
  publisher={IEEE}
}

@inproceedings{zhao2024equivariant,
  title={Equivariant multi-modality image fusion},
  author={Zhao, Zixiang and Bai, Haowen and Zhang, Jiangshe and Zhang, Yulun and Zhang, Kai and Xu, Shuang and Chen, Dongdong and Timofte, Radu and Van Gool, Luc},
  booktitle={Proceedings of the IEEE/CVF conference on computer vision and pattern recognition},
  pages={25912--25921},
  year={2024}
}

@article{tang2022piafusion,
  title={PIAFusion: A progressive infrared and visible image fusion network based on illumination aware},
  author={Tang, Linfeng and Yuan, Jiteng and Zhang, Hao and Jiang, Xingyu and Ma, Jiayi},
  journal={Information Fusion},
  volume={83},
  pages={79--92},
  year={2022},
  publisher={Elsevier}
}

@inproceedings{xu2020fusiondn,
  title={Fusiondn: A unified densely connected network for image fusion},
  author={Xu, Han and Ma, Jiayi and Le, Zhuliang and Jiang, Junjun and Guo, Xiaojie},
  booktitle={Proceedings of the AAAI conference on artificial intelligence},
  volume={34},
  number={07},
  pages={12484--12491},
  year={2020}
}

@inproceedings{liu2022target,
  title={Target-aware dual adversarial learning and a multi-scenario multi-modality benchmark to fuse infrared and visible for object detection},
  author={Liu, Jinyuan and Fan, Xin and Huang, Zhanbo and Wu, Guanyao and Liu, Risheng and Zhong, Wei and Luo, Zhongxuan},
  booktitle={Proceedings of the IEEE/CVF conference on computer vision and pattern recognition},
  pages={5802--5811},
  year={2022}
}

@inproceedings{liang2022fusion,
  title={Fusion from decomposition: A self-supervised decomposition approach for image fusion},
  author={Liang, Pengwei and Jiang, Junjun and Liu, Xianming and Ma, Jiayi},
  booktitle={European conference on computer vision},
  pages={719--735},
  year={2022},
  organization={Springer}
}

@article{ma2020infrared,
  title={Infrared and visible image fusion via detail preserving adversarial learning},
  author={Ma, Jiayi and Liang, Pengwei and Yu, Wei and Chen, Chen and Guo, Xiaojie and Wu, Jia and Jiang, Junjun},
  journal={Information Fusion},
  volume={54},
  pages={85--98},
  year={2020},
  publisher={Elsevier}
}

@inproceedings{zhao2023metafusion,
  title={Metafusion: Infrared and visible image fusion via meta-feature embedding from object detection},
  author={Zhao, Wenda and Xie, Shigeng and Zhao, Fan and He, You and Lu, Huchuan},
  booktitle={Proceedings of the IEEE/CVF Conference on Computer Vision and Pattern Recognition},
  pages={13955--13965},
  year={2023}
}

@inproceedings{zhao2023cddfuse,
  title={Cddfuse: Correlation-driven dual-branch feature decomposition for multi-modality image fusion},
  author={Zhao, Zixiang and Bai, Haowen and Zhang, Jiangshe and Zhang, Yulun and Xu, Shuang and Lin, Zudi and Timofte, Radu and Van Gool, Luc},
  booktitle={Proceedings of the IEEE/CVF conference on computer vision and pattern recognition},
  pages={5906--5916},
  year={2023}
}

@article{li2023lrrnet,
  title={Lrrnet: A novel representation learning guided fusion network for infrared and visible images},
  author={Li, Hui and Xu, Tianyang and Wu, Xiao-Jun and Lu, Jiwen and Kittler, Josef},
  journal={IEEE transactions on pattern analysis and machine intelligence},
  volume={45},
  number={9},
  pages={11040--11052},
  year={2023},
  publisher={IEEE}
}

@article{xu2023murf,
  title={Murf: Mutually reinforcing multi-modal image registration and fusion},
  author={Xu, Han and Yuan, Jiteng and Ma, Jiayi},
  journal={IEEE transactions on pattern analysis and machine intelligence},
  volume={45},
  number={10},
  pages={12148--12166},
  year={2023},
  publisher={IEEE}
}

@inproceedings{zhao2023ddfm,
  title={DDFM: denoising diffusion model for multi-modality image fusion},
  author={Zhao, Zixiang and Bai, Haowen and Zhu, Yuanzhi and Zhang, Jiangshe and Xu, Shuang and Zhang, Yulun and Zhang, Kai and Meng, Deyu and Timofte, Radu and Van Gool, Luc},
  booktitle={Proceedings of the IEEE/CVF international conference on computer vision},
  pages={8082--8093},
  year={2023}
}

@inproceedings{liu2023multi,
  title={Multi-interactive feature learning and a full-time multi-modality benchmark for image fusion and segmentation},
  author={Liu, Jinyuan and Liu, Zhu and Wu, Guanyao and Ma, Long and Liu, Risheng and Zhong, Wei and Luo, Zhongxuan and Fan, Xin},
  booktitle={Proceedings of the IEEE/CVF international conference on computer vision},
  pages={8115--8124},
  year={2023}
}

@inproceedings{iqbal2010enhancing,
  author    = {K. Iqbal and M. Odetayo and A. James and R. A. Salam and A. Z. H. Talib},
  title     = {Enhancing the low quality images using unsupervised colour correction method},
  booktitle = {IEEE International Conference on Systems, Man and Cybernetics},
  year      = {2010},
  pages     = {1703--1709},
}

@article{ghani2015underwater,
  author    = {A. S. A. Ghani and N. A. M. Isa},
  title     = {Underwater image quality enhancement through integrated color model with Rayleigh distribution},
  journal   = {Applied Soft Computing},
  volume    = {27},
  pages     = {219--230},
  year      = {2015},
}

@article{ancuti2018color,
  author    = {C. O. Ancuti and C. Ancuti and C. De Vleeschouwer and P. Bekaert},
  title     = {Color balance and fusion for underwater image enhancement},
  journal   = {IEEE Trans. Image Process.},
  volume    = {27},
  number    = {1},
  pages     = {379--393},
  year      = {2018},
}

@article{gao2019underwater,
  author    = {S.-B. Gao and M. Zhang and Q. Zhao and X.-S. Zhang and Y.-J. Li},
  title     = {Underwater image enhancement using adaptive retinal mechanisms},
  journal   = {IEEE Trans. Image Process.},
  volume    = {28},
  number    = {11},
  pages     = {5580--5595},
  year      = {2019},
}

@inproceedings{ancuti2012enhancing,
  author    = {C. Ancuti and C. O. Ancuti and T. Haber and P. Bekaert},
  title     = {Enhancing underwater images and videos by fusion},
  booktitle = {Proc. IEEE/CVF Conf. Comput. Vis. Pattern Recognit. (CVPR)},
  year      = {2012},
  pages     = {81--88},
}

@article{galdran2015automatic,
  author    = {Galdran, A. and Pardo, D. and Picon, A. and Alvarez-Gila, A.},
  title     = {Automatic red-channel underwater image restoration},
  journal   = {Journal of Visual Communication and Image Representation},
  volume    = {26},
  pages     = {132--145},
  year      = {2015},
  publisher = {Elsevier}
}

@article{drews2016underwater,
  author    = {Drews, P. L. and Nascimento, E. R. and Botelho, S. S. and Campos, M. F. M.},
  title     = {Underwater depth estimation and image restoration based on single images},
  journal   = {IEEE Computer Graphics and Applications},
  volume    = {36},
  number    = {2},
  pages     = {24--35},
  year      = {2016},
  publisher = {IEEE}
}

@article{li2016single,
  author    = {Li, C. and Guo, J. and Wang, B. and Cong, R. and Zhang, Y. and Wang, J.},
  title     = {Single underwater image enhancement based on color cast removal and visibility restoration},
  journal   = {Journal of Electronic Imaging},
  volume    = {25},
  number    = {3},
  pages     = {033012},
  year      = {2016},
  publisher = {SPIE}
}

@article{li2016underwater,
  author    = {Li, C.-Y. and Guo, J.-C. and Cong, R.-M. and Pang, Y.-W. and Wang, B.},
  title     = {Underwater image enhancement by dehazing with minimum information loss and histogram distribution prior},
  journal   = {IEEE Transactions on Image Processing},
  volume    = {25},
  number    = {12},
  pages     = {5664--5677},
  year      = {2016},
  publisher = {IEEE}
}

@article{zhang2022underwater,
  author    = {Zhang, W. and Wang, Y. and Li, C.},
  title     = {Underwater image enhancement by attenuated color channel correction and detail preserved contrast enhancement},
  journal   = {IEEE Journal of Oceanic Engineering},
  pages     = {1--18},
  year      = {2022},
  publisher = {IEEE}
}

@article{li2019underwater,
  author    = {Li, Chongyi and Guo, Chunle and Ren, Wenqi and Cong, Runmin and Hou, Junhui and Kwong, Sam and Tao, Dacheng},
  title     = {An Underwater Image Enhancement Benchmark Dataset and Beyond},
  journal   = {IEEE Transactions on Image Processing},
  volume    = {29},
  pages     = {4376--4389},
  year      = {2019},
  month     = {nov},
  doi       = {10.1109/TIP.2019.2955241}
}

@article{peng2023ushapetransformer,
  author    = {Peng, L. and Zhu, C. and Bian, L.},
  title     = {U-Shape Transformer for Underwater Image Enhancement},
  journal   = {IEEE Transactions on Image Processing},
  volume    = {32},
  pages     = {3066--3079},
  year      = {2023},
  doi       = {10.1109/TIP.2023.3276332}
}

@article{Li2019Fusion,
  author  = {H. Li and J. Li and W. Wang},
  title   = {{A Fusion Adversarial Underwater Image Enhancement Network with a Public Test Dataset}},
  journal = {arXiv preprint},
  year    = {2019},
  doi     = {10.48550/arXiv.1906.06819},
  note    = {arXiv:1906.06819},
}

@inproceedings{tang2023transformerdiffusion,
  author    = {Tang, Yi and Kawasaki, Hiroshi and Iwaguchi, Takafumi},
  title     = {Underwater Image Enhancement by Transformer-based Diffusion Model with Non-uniform Sampling for Skip Strategy},
  booktitle = {Proceedings of the 31st ACM International Conference on Multimedia (MM '23)},
  address   = {New York, NY, USA},
  pages     = {5419--5427},
  year      = {2023},
  organization = {Association for Computing Machinery},
  doi       = {10.1145/3581783.3612475}
}

@inproceedings{zhao2024wfdiff,
  author    = {Zhao, C. and Cai, W. and Dong, C. and Hu, C.},
  title     = {Wavelet-based Fourier Information Interaction with Frequency Diffusion Adjustment for Underwater Image Restoration},
  booktitle = {2024 IEEE/CVF Conference on Computer Vision and Pattern Recognition (CVPR)},
  address   = {Seattle, WA, USA},
  pages     = {8281--8291},
  year      = {2024},
  month     = {jun},
  organization = {IEEE/CVF},
  doi       = {10.1109/CVPR52729.2024.00813}
}

@article{yang2015underwatercolor,
  author    = {Yang, M. and Sowmya, A.},
  title     = {An underwater color image quality evaluation metric},
  journal   = {IEEE Transactions on Image Processing},
  volume    = {24},
  number    = {12},
  pages     = {6062--6071},
  year      = {2015},
  doi       = {10.1109/TIP.2015.2480136}
}

@article{panetta2015hvsunderwater,
  author    = {Panetta, K. and Gao, C. and Agaian, S.},
  title     = {Human-visual-system-inspired underwater image quality measures},
  journal   = {IEEE Journal of Oceanic Engineering},
  volume    = {41},
  number    = {3},
  pages     = {541--551},
  year      = {2015},
  doi       = {10.1109/JOE.2015.2410644}
}

@article{cong2023pugan,
  author    = {Cong, R. and Yang, W. and Zhang, W. and Li, C. and Guo, C.-L. and Huang, Q. and Kwong, S.},
  title     = {PUGAN: Physical model-guided underwater image enhancement using GAN with dual-discriminators},
  journal   = {IEEE Transactions on Image Processing},
  volume    = {32},
  pages     = {4472--4485},
  year      = {2023}
}

@article{wang2021uiec2net,
  author    = {Wang, Yudong and Guo, Jichang and Gao, Huan and Yue, Huihui},
  title     = {UIEC²-Net: CNN-based underwater image enhancement using two color space},
  journal   = {Signal Processing: Image Communication},
  volume    = {96},
  pages     = {116250},
  year      = {2021},
  issn      = {0923-5965}
}

@INPROCEEDINGS{Bychkovsky2011FiveK,
  author={Bychkovsky, Vladimir and Paris, Sylvain and Chan, Eric and Durand, Fredo},
  booktitle={CVPR 2011}, 
  title={Learning photographic global tonal adjustment with a database of input / output image pairs}, 
  year={2011},
  volume={},
  number={},
  pages={97-104},
  doi={10.1109/CVPR.2011.5995332}}

@INPROCEEDINGS{Vinker2021HDR+,
  author={Vinker, Yael and Huberman-Spiegelglas, Inbar and Fattal, Raanan},
  booktitle={2021 IEEE/CVF International Conference on Computer Vision (ICCV)}, 
  title={Unpaired Learning for High Dynamic Range Image Tone Mapping}, 
  year={2021},
  volume={},
  number={},
  pages={14637-14646},
  doi={10.1109/ICCV48922.2021.01439}}

@inproceedings{wang2019underexposed,
  title={Underexposed photo enhancement using deep illumination estimation},
  author={Wang, Ruixing and Zhang, Qing and Fu, Chi-Wing and Shen, Xiaoyong and Zheng, Wei-Shi and Jia, Jiaya},
  booktitle={Proceedings of the IEEE/CVF Conference on Computer Vision and Pattern Recognition},
  pages={6849--6857},
  year={2019}
}

@article{gharbi2017deep,
  title={Deep bilateral learning for real-time image enhancement},
  author={Gharbi, Micha{\"e}l and Chen, Jiawen and Barron, Jonathan T and Hasinoff, Samuel W and Durand, Fr{\'e}do},
  journal={ACM Transactions on Graphics (TOG)},
  volume={36},
  number={4},
  pages={1--12},
  year={2017},
  publisher={ACM New York, NY, USA}
}

@inproceedings{he2020conditional,
  title={Conditional sequential modulation for efficient global image retouching},
  author={He, Jingwen and Liu, Yihao and Qiao, Yu and Dong, Chao},
  booktitle={European Conference on Computer Vision},
  pages={679--695},
  year={2020},
  publisher={Springer}
}

@inproceedings{moran2020deeplpf,
  title={Deeplpf: Deep local parametric filters for image enhancement},
  author={Moran, Sean and Marza, Pierre and McDonagh, Steven and Parisot, Sarah and Slabaugh, Gregory},
  booktitle={Proceedings of the IEEE/CVF Conference on Computer Vision and Pattern Recognition},
  pages={12826--12835},
  year={2020}
}

@article{zeng2020learning,
  title={Learning image-adaptive 3d lookup tables for high performance photo enhancement in real-time},
  author={Zeng, Hui and Cai, Jianrui and Li, Lida and Cao, Zisheng and Zhang, Lei},
  journal={IEEE Transactions on Pattern Analysis and Machine Intelligence},
  volume={44},
  number={4},
  pages={2058--2073},
  year={2020},
  publisher={IEEE}
}

@inproceedings{wang2021realtime,
  title={Real-time image enhancer via learnable spatial-aware 3d lookup tables},
  author={Wang, Tao and Li, Yong and Peng, Jingyang and Ma, Yipeng and Wang, Xian and Song, Fenglong and Yan, Youliang},
  booktitle={Proceedings of the IEEE/CVF International Conference on Computer Vision},
  pages={2471--2480},
  year={2021}
}

@inproceedings{zhang2022clut,
  title={Clut-net: Learning adaptively compressed representations of 3dluts for lightweight image enhancement},
  author={Zhang, Fengyi and Zeng, Hui and Zhang, Tianjun and Zhang, Lin},
  booktitle={Proceedings of the 30th ACM International Conference on Multimedia},
  pages={6493--6501},
  year={2022},
  publisher={ACM}
}

@article{zhang2025high,
  title={High-resolution Photo Enhancement in Real-time: A Laplacian Pyramid Network},
  author={Zhang, Feng and Deng, Haoyou and Li, Zhiqiang and Li, Lida and Xu, Bin and Lu, Qingbo and Cao, Zisheng and Wei, Minchen and Gao, Changxin and Sang, Nong and others},
  journal={IEEE Transactions on Pattern Analysis and Machine Intelligence},
  year={2025},
  publisher={IEEE}
}

@article{Hullin2011FlareRendering, author = {Hullin, Matthias and Eisemann, Elmar and Seidel, Hans-Peter and Lee, Sungkil}, title = {Physically-based real-time lens flare rendering}, year = {2011}, volume = {30}, number = {4}, journal = {ACM Trans. Graph.}}

@INPROCEEDINGS{Li2021CARMC,
  author={Li, Xiaoyu and Zhang, Bo and Liao, Jing and Sander, Pedro V.},
  booktitle={2021 IEEE/CVF International Conference on Computer Vision (ICCV)}, 
  title={Let’s See Clearly: Contaminant Artifact Removal for Moving Cameras}, 
  year={2021},
  volume={},
  number={},
  pages={1991-2000},
  doi={10.1109/ICCV48922.2021.00202}}

@INPROCEEDINGS{Wu2021NN4FR,
  author={Wu, Yicheng and He, Qiurui and Xue, Tianfan and Garg, Rahul and Chen, Jiawen and Veeraraghavan, Ashok and Barron, Jonathan T.},
  booktitle={2021 IEEE/CVF International Conference on Computer Vision (ICCV)}, 
  title={How to Train Neural Networks for Flare Removal}, 
  year={2021},
  volume={},
  number={},
  pages={2219-2227},
  doi={10.1109/ICCV48922.2021.00224}}

@ARTICLE{Dai2024Flare7Kpp,
  author={Dai, Yuekun and Li, Chongyi and Zhou, Shangchen and Feng, Ruicheng and Luo, Yihang and Loy, Chen Change},
  journal={IEEE Transactions on Pattern Analysis and Machine Intelligence}, 
  title={Flare7K++: Mixing Synthetic and Real Datasets for Nighttime Flare Removal and Beyond}, 
  year={2024},
  volume={46},
  number={11},
  pages={7041-7055},
  doi={10.1109/TPAMI.2024.3406821}}

@INPROCEEDINGS{Dai2024MIPI,
  author={Dai, Yuekun and Zhang, Dafeng and Li, Xiaoming and Yue, Zongsheng and Li, Chongyi and Zhou, Shangchen and Feng, Ruicheng and Yang, Peiqing and Jin, Zhezhu and Liu, Guanqun and Loy, Chen Change},
  booktitle={2024 IEEE/CVF Conference on Computer Vision and Pattern Recognition Workshops (CVPRW)}, 
  title={MIPI 2024 Challenge on Nighttime Flare Removal: Methods and Results}, 
  year={2024},
  volume={},
  number={},
  pages={1144-1152},
  doi={10.1109/CVPRW63382.2024.00121}}

@inproceedings{Huang2025DeflareMamba,
author = {Huang, Yihang and Huang, Yuanfei and Lin, Junhui and Huang, Hua},
title = {DeflareMamba: Hierarchical Vision Mamba for Contextually Consistent Lens Flare Removal},
year = {2025},
booktitle = {Proceedings of the 33rd ACM International Conference on Multimedia},
pages = {8028–8037},
numpages = {10},
}

@article{deng2024towards,
  title={Towards Blind Flare Removal Using Knowledge-driven Flare-level Estimator},
  author={Deng, Haoyou and Li, Lida and Zhang, Feng and Li, Zhiqiang and Xu, Bin and Lu, Qingbo and Gao, Changxin and Sang, Nong},
  journal={IEEE Transactions on Image Processing},
  year={2024},
  publisher={IEEE}
}

@article{caiF2T2HiTUShapedFFT2025,
  title={F2t2-hit: A u-shaped fft transformer and hierarchical transformer for reflection removal},
  author={Cai, Jie and Yang, Kangning and Ouyang, Ling and Fu, Lan and Ding, Jiaming and Sun, Huiming and Ho, Chiu Man and Meng, Zibo},
  journal={arXiv preprint arXiv:2506.05489},
  year={2025}
}

@inproceedings{dongLocationawareSingleImage2021,
  title={Location-aware single image reflection removal},
  author={Dong, Zheng and Xu, Ke and Yang, Yin and Bao, Hujun and Xu, Weiwei and Lau, Rynson WH},
  booktitle={Proceedings of the IEEE/CVF international conference on computer vision},
  pages={5017--5026},
  year={2021}
}

@article{guoSingleImageReflection2024,
  title={Single image reflection separation via dual-stream interactive transformers},
  author={Hu, Qiming and Wang, Hainuo and Guo, Xiaojie},
  journal={Advances in Neural Information Processing Systems},
  volume={37},
  pages={55228--55248},
  year={2024}
}

@inproceedings{hongLDiffERSingleImage2025,
  title={L-differ: Single image reflection removal with language-based diffusion model},
  author={Hong, Yuchen and Zhong, Haofeng and Weng, Shuchen and Liang, Jinxiu and Shi, Boxin},
  booktitle={European Conference on Computer Vision},
  pages={58--76},
  year={2024},
  organization={Springer}
}

@article{huangSingleImageReflection2025,
  title={Single Image Reflection Removal via inter-layer Complementarity},
  author={Huang, Yue and Li, Zi'ang and Hu, Tianle and Wen, Jie and Li, Guanbin and Zhang, Jinglin and Zhou, Guoxu and Fang, Xiaozhao},
  journal={arXiv preprint arXiv:2505.12641},
  year={2025}
}

@article{huDereflectionAnyImage2025,
  title={Dereflection Any Image with Diffusion Priors and Diversified Data},
  author={Hu, Jichen and Yang, Chen and Zhou, Zanwei and Fang, Jiemin and Yang, Xiaokang and Tian, Qi and Shen, Wei},
  journal={arXiv preprint arXiv:2503.17347},
  year={2025}
}

@inproceedings{huSingleImageReflection2023,
  title={Single image reflection separation via component synergy},
  author={Hu, Qiming and Guo, Xiaojie},
  booktitle={Proceedings of the IEEE/CVF international conference on computer vision},
  pages={13138--13147},
  year={2023}
}

@article{huTrashTreasureInteractive2021,
  title={Trash or treasure? an interactive dual-stream strategy for single image reflection separation},
  author={Hu, Qiming and Guo, Xiaojie},
  journal={Advances in Neural Information Processing Systems},
  volume={34},
  pages={24683--24694},
  year={2021}
}

@inproceedings{liSingleImageReflection2020,
  title={Single image reflection removal through cascaded refinement},
  author={Li, Chao and Yang, Yixiao and He, Kun and Lin, Stephen and Hopcroft, John E},
  booktitle={Proceedings of the IEEE/CVF conference on computer vision and pattern recognition},
  pages={3565--3574},
  year={2020}
}

@article{liTwostageSingleImage2023,
  title={Two-stage single image reflection removal with reflection-aware guidance},
  author={Li, Yu and Liu, Ming and Yi, Yaling and Li, Qince and Ren, Dongwei and Zuo, Wangmeng},
  journal={Applied Intelligence},
  volume={53},
  number={16},
  pages={19433--19448},
  year={2023},
  publisher={Springer}
}

@article{luSingleimageReflectionRemoval2025,
  title={Single-image reflection removal via self-supervised diffusion models},
  author={Lu, Zhengyang and Wang, Weifan and Guo, Tianhao and Wang, Feng},
  journal={The Journal of Supercomputing},
  volume={81},
  number={1},
  pages={338},
  year={2025},
  publisher={Springer}
}

@inproceedings{wanBenchmarkingSingleImageReflection2017,
  title={Benchmarking single-image reflection removal algorithms},
  author={Wan, Renjie and Shi, Boxin and Duan, Ling-Yu and Tan, Ah-Hwee and Kot, Alex C},
  booktitle={Proceedings of the IEEE international conference on computer vision},
  pages={3922--3930},
  year={2017}
}

@inproceedings{weiSingleImageReflection2019,
  title={Single image reflection removal exploiting misaligned training data and network enhancements},
  author={Wei, Kaixuan and Yang, Jiaolong and Fu, Ying and Wipf, David and Huang, Hua},
  booktitle={Proceedings of the IEEE/CVF Conference on Computer Vision and Pattern Recognition},
  pages={8178--8187},
  year={2019}
}

@inproceedings{yangSeeingDeeplyBidirectionally2018,
  title={Seeing deeply and bidirectionally: A deep learning approach for single image reflection removal},
  author={Yang, Jie and Gong, Dong and Liu, Lingqiao and Shi, Qinfeng},
  booktitle={Proceedings of the european conference on computer vision (ECCV)},
  pages={654--669},
  year={2018}
}

@article{zakarinReflectionRemovalEfficient2025,
  title={Reflection Removal through Efficient Adaptation of Diffusion Transformers},
  author={Zakarin, Daniyar and Wandel, Thiemo and Obukhov, Anton and Dai, Dengxin},
  journal={arXiv preprint arXiv:2512.05000},
  year={2025}
}

@inproceedings{zhangSingleImageReflection2018,
  title={Single image reflection separation with perceptual losses},
  author={Zhang, Xuaner and Ng, Ren and Chen, Qifeng},
  booktitle={Proceedings of the IEEE conference on computer vision and pattern recognition},
  pages={4786--4794},
  year={2018}
}

@inproceedings{zhaoReversibleDecouplingNetwork2025,
  title={Reversible decoupling network for single image reflection removal},
  author={Zhao, Hao and Li, Mingjia and Hu, Qiming and Guo, Xiaojie},
  booktitle={Proceedings of the Computer Vision and Pattern Recognition Conference},
  pages={26430--26439},
  year={2025}
}

@inproceedings{zhuRevisitingSingleImage2024,
  title={Revisiting single image reflection removal in the wild},
  author={Zhu, Yurui and Fu, Xueyang and Jiang, Peng-Tao and Zhang, Hao and Sun, Qibin and Chen, Jinwei and Zha, Zheng-Jun and Li, Bo},
  booktitle={Proceedings of the IEEE/CVF Conference on Computer Vision and Pattern Recognition},
  pages={25468--25478},
  year={2024}
}

@inproceedings{hu2018direction,
  title={Direction-aware spatial context features for shadow detection},
  author={Hu, Xiaowei and Zhu, Lei and Fu, Chi-Wing and Qin, Jing and Heng, Pheng-Ann},
  booktitle={Proceedings of the IEEE conference on computer vision and pattern recognition},
  pages={7454--7462},
  year={2018}
}

@inproceedings{cun2020towards,
  title={Towards ghost-free shadow removal via dual hierarchical aggregation network and shadow matting gan},
  author={Cun, Xiaodong and Pun, Chi-Man and Shi, Cheng},
  booktitle={Proceedings of the AAAI conference on artificial intelligence},
  volume={34},
  number={07},
  pages={10680--10687},
  year={2020}
}

@inproceedings{zhu2022bijective,
  title={Bijective mapping network for shadow removal},
  author={Zhu, Yurui and Huang, Jie and Fu, Xueyang and Zhao, Feng and Sun, Qibin and Zha, Zheng-Jun},
  booktitle={Proceedings of the IEEE/CVF Conference on Computer Vision and Pattern Recognition},
  pages={5627--5636},
  year={2022}
}

@inproceedings{guo2023shadowformer,
  title={ShadowFormer: Global context helps shadow removal},
  author={Guo, Lanqing and Huang, Siyu and Liu, Ding and Cheng, Hao and Wen, Bihan},
  booktitle={Proceedings of the AAAI conference on artificial intelligence},
  volume={37},
  number={1},
  pages={710--718},
  year={2023}
}

@inproceedings{guo2023shadowdiffusion,
  title={Shadowdiffusion: When degradation prior meets diffusion model for shadow removal},
  author={Guo, Lanqing and Wang, Chong and Yang, Wenhan and Huang, Siyu and Wang, Yufei and Pfister, Hanspeter and Wen, Bihan},
  booktitle={Proceedings of the IEEE/CVF conference on computer vision and pattern recognition},
  pages={14049--14058},
  year={2023}
}

@inproceedings{xiao2024homoformer,
  title={Homoformer: Homogenized transformer for image shadow removal},
  author={Xiao, Jie and Fu, Xueyang and Zhu, Yurui and Li, Dong and Huang, Jie and Zhu, Kai and Zha, Zheng-Jun},
  booktitle={Proceedings of the IEEE/CVF conference on computer vision and pattern recognition},
  pages={25617--25626},
  year={2024}
}

@article{liu2023visual,
  title={Visual instruction tuning},
  author={Liu, Haotian and Li, Chunyuan and Wu, Qingyang and Lee, Yong Jae},
  journal={Advances in neural information processing systems},
  volume={36},
  pages={34892--34916},
  year={2023}
}

@article{bai2025qwen2,
  title={Qwen2. 5-vl technical report},
  author={Bai, Shuai and Chen, Keqin and Liu, Xuejing and Wang, Jialin and Ge, Wenbin and Song, Sibo and Dang, Kai and Wang, Peng and Wang, Shijie and Tang, Jun and others},
  journal={arXiv preprint arXiv:2502.13923},
  year={2025}
}

@article{lu2025hyper,
  title={Hyper-bagel: A unified acceleration framework for multimodal understanding and generation},
  author={Lu, Yanzuo and Xia, Xin and Zhang, Manlin and Kuang, Huafeng and Zheng, Jianbin and Ren, Yuxi and Xiao, Xuefeng},
  journal={arXiv preprint arXiv:2509.18824},
  year={2025}
}

@incollection{murali2020image,
  title={Image denoising using DnCNN: An exploration study},
  author={Murali, Vineeth and Sudeep, PV},
  booktitle={Advances in Communication Systems and Networks: Select Proceedings of ComNet 2019},
  pages={847--859},
  year={2020},
  publisher={Springer}
}

@article{mcmaster,
  title={Color demosaicking by local directional interpolation and nonlocal adaptive thresholding},
  author={Zhang, Lei and Wu, Xiaolin and Buades, Antoni and Li, Xin},
  journal={Journal of Electronic imaging},
  volume={20},
  number={2},
  pages={023016--023016},
  year={2011},
  publisher={Society of Photo-Optical Instrumentation Engineers}
}

@book{Kodak,
  title={Kodak lossless true color image suite},
  author={Rich Franzen},
  volume={5},
  year={1999},
  publisher={https://r0k.us/graphics/kodak/}
}

@inproceedings{urban,
  title={Single image super-resolution from transformed self-exemplars},
  author={Huang, Jia-Bin and Singh, Abhishek and Ahuja, Narendra},
  booktitle={Proceedings of the IEEE conference on computer vision and pattern recognition},
  pages={5197--5206},
  year={2015}
}

@article{polyu,
  title={Real-world noisy image denoising: A new benchmark},
  author={Xu, Jun and Li, Hui and Liang, Zhetong and Zhang, David and Zhang, Lei},
  journal={arXiv preprint arXiv:1804.02603},
  year={2018}
}

@inproceedings{sidd,
  title={A high-quality denoising dataset for smartphone cameras},
  author={Abdelhamed, Abdelrahman and Lin, Stephen and Brown, Michael S},
  booktitle={Proceedings of the IEEE conference on computer vision and pattern recognition},
  pages={1692--1700},
  year={2018}
}

@article{guo2012paired,
  title={Paired regions for shadow detection and removal},
  author={Guo, Ruiqi and Dai, Qieyun and Hoiem, Derek},
  journal={IEEE transactions on pattern analysis and machine intelligence},
  volume={35},
  number={12},
  pages={2956--2967},
  year={2012},
  publisher={IEEE}
}

@article{dncnn,
  title={Beyond a gaussian denoiser: Residual learning of deep cnn for image denoising},
  author={Zhang, Kai and Zuo, Wangmeng and Chen, Yunjin and Meng, Deyu and Zhang, Lei},
  journal={IEEE transactions on image processing},
  volume={26},
  number={7},
  pages={3142--3155},
  year={2017},
  publisher={IEEE}
}

@inproceedings{maskdenoising,
  title={Masked image training for generalizable deep image denoising},
  author={Chen, Haoyu and Gu, Jinjin and Liu, Yihao and Magid, Salma Abdel and Dong, Chao and Wang, Qiong and Pfister, Hanspeter and Zhu, Lei},
  booktitle={Proceedings of the IEEE/CVF Conference on Computer Vision and Pattern Recognition},
  pages={1692--1703},
  year={2023}
}

@article{HAT,
  title={Towards adversarially robust deep image denoising},
  author={Yan, Hanshu and Zhang, Jingfeng and Feng, Jiashi and Sugiyama, Masashi and Tan, Vincent YF},
  journal={arXiv preprint arXiv:2201.04397},
  year={2022}
}

@inproceedings{DIL,
  title={Learning distortion invariant representation for image restoration from a causality perspective},
  author={Li, Xin and Li, Bingchen and Jin, Xin and Lan, Cuiling and Chen, Zhibo},
  booktitle={Proceedings of the IEEE/CVF Conference on Computer Vision and Pattern Recognition},
  pages={1714--1724},
  year={2023}
}

@article{lolv1,
  title={Deep retinex decomposition for low-light enhancement},
  author={Wei, Chen and Wang, Wenjing and Yang, Wenhan and Liu, Jiaying},
  journal={arXiv preprint arXiv:1808.04560},
  year={2018}
}

@article{lolv2,
  title={Sparse gradient regularized deep retinex network for robust low-light image enhancement},
  author={Yang, Wenhan and Wang, Wenjing and Huang, Haofeng and Wang, Shiqi and Liu, Jiaying},
  journal={IEEE Transactions on Image Processing},
  volume={30},
  pages={2072--2086},
  year={2021},
  publisher={IEEE}
}

@article{sice,
  title={Learning a deep single image contrast enhancer from multi-exposure images},
  author={Cai, Jianrui and Gu, Shuhang and Zhang, Lei},
  journal={IEEE Transactions on Image Processing},
  volume={27},
  number={4},
  pages={2049--2062},
  year={2018},
  publisher={IEEE}
}

@inproceedings{kind,
  title={Kindling the darkness: A practical low-light image enhancer},
  author={Zhang, Yonghua and Zhang, Jiawan and Guo, Xiaojie},
  booktitle={Proceedings of the 27th ACM international conference on multimedia},
  pages={1632--1640},
  year={2019}
}

@inproceedings{zerodce,
  title={Zero-reference deep curve estimation for low-light image enhancement},
  author={Guo, Chunle and Li, Chongyi and Guo, Jichang and Loy, Chen Change and Hou, Junhui and Kwong, Sam and Cong, Runmin},
  booktitle={Proceedings of the IEEE/CVF conference on computer vision and pattern recognition},
  pages={1780--1789},
  year={2020}
}

@inproceedings{ruas,
  title={Retinex-inspired unrolling with cooperative prior architecture search for low-light image enhancement},
  author={Liu, Risheng and Ma, Long and Zhang, Jiaao and Fan, Xin and Luo, Zhongxuan},
  booktitle={Proceedings of the IEEE/CVF conference on computer vision and pattern recognition},
  pages={10561--10570},
  year={2021}
}

@inproceedings{llflow,
  title={Low-light image enhancement with normalizing flow},
  author={Wang, Yufei and Wan, Renjie and Yang, Wenhan and Li, Haoliang and Chau, Lap-Pui and Kot, Alex},
  booktitle={Proceedings of the AAAI conference on artificial intelligence},
  volume={36},
  pages={2604--2612},
  year={2022}
}

@article{enlightengan,
  title={Enlightengan: Deep light enhancement without paired supervision},
  author={Jiang, Yifan and Gong, Xinyu and Liu, Ding and Cheng, Yu and Fang, Chen and Shen, Xiaohui and Yang, Jianchao and Zhou, Pan and Wang, Zhangyang},
  journal={IEEE transactions on image processing},
  volume={30},
  pages={2340--2349},
  year={2021},
  publisher={IEEE}
}

@inproceedings{snraware,
  title={Snr-aware low-light image enhancement},
  author={Xu, Xiaogang and Wang, Ruixing and Fu, Chi-Wing and Jia, Jiaya},
  booktitle={Proceedings of the IEEE/CVF conference on computer vision and pattern recognition},
  pages={17714--17724},
  year={2022}
}

@article{bread,
  title={Low-light image enhancement via breaking down the darkness},
  author={Guo, Xiaojie and Hu, Qiming},
  journal={International Journal of Computer Vision},
  volume={131},
  number={1},
  pages={48--66},
  year={2023},
  publisher={Springer}
}

@inproceedings{pairlie,
  title={Learning a simple low-light image enhancer from paired low-light instances},
  author={Fu, Zhenqi and Yang, Yan and Tu, Xiaotong and Huang, Yue and Ding, Xinghao and Ma, Kai-Kuang},
  booktitle={Proceedings of the IEEE/CVF conference on computer vision and pattern recognition},
  pages={22252--22261},
  year={2023}
}

@inproceedings{llformer,
  title={Ultra-high-definition low-light image enhancement: A benchmark and transformer-based method},
  author={Wang, Tao and Zhang, Kaihao and Shen, Tianrun and Luo, Wenhan and Stenger, Bjorn and Lu, Tong},
  booktitle={Proceedings of the AAAI conference on artificial intelligence},
  volume={37},
  pages={2654--2662},
  year={2023}
}

@inproceedings{retinexformer,
  title={Retinexformer: One-stage retinex-based transformer for low-light image enhancement},
  author={Cai, Yuanhao and Bian, Hao and Lin, Jing and Wang, Haoqian and Timofte, Radu and Zhang, Yulun},
  booktitle={Proceedings of the IEEE/CVF international conference on computer vision},
  pages={12504--12513},
  year={2023}
}

@article{GSAD,
  title={Global structure-aware diffusion process for low-light image enhancement},
  author={Hou, Jinhui and Zhu, Zhiyu and Hou, Junhui and Liu, Hui and Zeng, Huanqiang and Yuan, Hui},
  journal={Advances in Neural Information Processing Systems},
  volume={36},
  pages={79734--79747},
  year={2023}
}

@inproceedings{quadprior,
  title={Zero-reference low-light enhancement via physical quadruple priors},
  author={Wang, Wenjing and Yang, Huan and Fu, Jianlong and Liu, Jiaying},
  booktitle={Proceedings of the IEEE/CVF conference on computer vision and pattern recognition},
  pages={26057--26066},
  year={2024}
}

@inproceedings{guo2022image,
  title={Image dehazing transformer with transmission-aware 3d position embedding},
  author={Guo, Chun-Le and Yan, Qixin and Anwar, Saeed and Cong, Runmin and Ren, Wenqi and Li, Chongyi},
  booktitle={Proceedings of the IEEE/CVF conference on computer vision and pattern recognition},
  pages={5812--5820},
  year={2022}
}

@inproceedings{qiu2023mb,
  title={Mb-taylorformer: Multi-branch efficient transformer expanded by taylor formula for image dehazing},
  author={Qiu, Yuwei and Zhang, Kaihao and Wang, Chenxi and Luo, Wenhan and Li, Hongdong and Jin, Zhi},
  booktitle={Proceedings of the IEEE/CVF international conference on computer vision},
  pages={12802--12813},
  year={2023}
}

@article{song2023vision,
  title={Vision transformers for single image dehazing},
  author={Song, Yuda and He, Zhuqing and Qian, Hui and Du, Xin},
  journal={IEEE Transactions on Image Processing},
  volume={32},
  pages={1927--1941},
  year={2023},
  publisher={IEEE}
}

@inproceedings{dong2020multi,
  title={Multi-scale boosted dehazing network with dense feature fusion},
  author={Dong, Hang and Pan, Jinshan and Xiang, Lei and Hu, Zhe and Zhang, Xinyi and Wang, Fei and Yang, Ming-Hsuan},
  booktitle={Proceedings of the IEEE/CVF conference on computer vision and pattern recognition},
  pages={2157--2167},
  year={2020}
}

@inproceedings{shao2020domain,
  title={Domain adaptation for image dehazing},
  author={Shao, Yuanjie and Li, Lerenhan and Ren, Wenqi and Gao, Changxin and Sang, Nong},
  booktitle={Proceedings of the IEEE/CVF conference on computer vision and pattern recognition},
  pages={2808--2817},
  year={2020}
}

@inproceedings{chen2021psd,
  title={PSD: Principled synthetic-to-real dehazing guided by physical priors},
  author={Chen, Zeyuan and Wang, Yangchao and Yang, Yang and Liu, Dong},
  booktitle={Proceedings of the IEEE/CVF conference on computer vision and pattern recognition},
  pages={7180--7189},
  year={2021}
}

@inproceedings{yang2022self,
  title={Self-augmented unpaired image dehazing via density and depth decomposition},
  author={Yang, Yang and Wang, Chaoyue and Liu, Risheng and Zhang, Lin and Guo, Xiaojie and Tao, Dacheng},
  booktitle={Proceedings of the IEEE/CVF conference on computer vision and pattern recognition},
  pages={2037--2046},
  year={2022}
}

@inproceedings{wu2023ridcp,
  title={Ridcp: Revitalizing real image dehazing via high-quality codebook priors},
  author={Wu, Rui-Qi and Duan, Zheng-Peng and Guo, Chun-Le and Chai, Zhi and Li, Chongyi},
  booktitle={Proceedings of the IEEE/CVF conference on computer vision and pattern recognition},
  pages={22282--22291},
  year={2023}
}

@article{fang2024real,
  title={Real-world image dehazing with coherence-based pseudo labeling and cooperative unfolding network},
  author={Fang, Chengyu and He, Chunming and Xiao, Fengyang and Zhang, Yulun and Tang, Longxiang and Zhang, Yuelin and Li, Kai and Li, Xiu},
  journal={Advances in Neural Information Processing Systems},
  volume={37},
  pages={97859--97883},
  year={2024}
}

@article{he2010single,
  title={Single image haze removal using dark channel prior},
  author={He, Kaiming and Sun, Jian and Tang, Xiaoou},
  journal={IEEE transactions on pattern analysis and machine intelligence},
  volume={33},
  number={12},
  pages={2341--2353},
  year={2010},
  publisher={IEEE}
}

@inproceedings{berman2016non,
  title={Non-local image dehazing},
  author={Berman, Dana and Avidan, Shai and others},
  booktitle={Proceedings of the IEEE conference on computer vision and pattern recognition},
  pages={1674--1682},
  year={2016}
}

@inproceedings{yang2017deep,
  title={Deep joint rain detection and removal from a single image},
  author={Yang, Wenhan and Tan, Robby T and Feng, Jiashi and Liu, Jiaying and Guo, Zongming and Yan, Shuicheng},
  booktitle={Proceedings of the IEEE conference on computer vision and pattern recognition},
  pages={1357--1366},
  year={2017}
}

@InProceedings{Wang_2019_CVPR,
  author = {Wang, Tianyu and Yang, Xin and Xu, Ke and Chen, Shaozhe and Zhang, Qiang and Lau, Rynson W.H.},
  title = {Spatial Attentive Single-Image Deraining with a High Quality Real Rain Dataset},
  booktitle = {The IEEE Conference on Computer Vision and Pattern Recognition (CVPR)},
  month = {June},
  year = {2019}
}

@InProceedings{NeRD-Rain,
    author={Chen, Xiang and Pan, Jinshan and Dong, Jiangxin}, 
    title={Bidirectional Multi-Scale Implicit Neural Representations for Image Deraining},
    booktitle={Proceedings of the IEEE/CVF Conference on Computer Vision and Pattern Recognition (CVPR)},
    month={June},
    year={2024}
}

@inproceedings{luo2015removing,
  title={Removing rain from a single image via discriminative sparse coding},
  author={Luo, Yu and Xu, Yong and Ji, Hui},
  booktitle={Proceedings of the IEEE international conference on computer vision},
  pages={3397--3405},
  year={2015}
}

@inproceedings{li2016rain,
  title={Rain streak removal using layer priors},
  author={Li, Yu and Tan, Robby T and Guo, Xiaojie and Lu, Jiangbo and Brown, Michael S},
  booktitle={Proceedings of the IEEE conference on computer vision and pattern recognition},
  pages={2736--2744},
  year={2016}
}

@inproceedings{fu2017removing,
  title={Removing rain from single images via a deep detail network},
  author={Fu, Xueyang and Huang, Jiabin and Zeng, Delu and Huang, Yue and Ding, Xinghao and Paisley, John},
  booktitle={Proceedings of the IEEE conference on computer vision and pattern recognition},
  pages={3855--3863},
  year={2017}
}

@inproceedings{li2018recurrent,
  title={Recurrent squeeze-and-excitation context aggregation net for single image deraining},
  author={Li, Xia and Wu, Jianlong and Lin, Zhouchen and Liu, Hong and Zha, Hongbin},
  booktitle={Proceedings of the European conference on computer vision (ECCV)},
  pages={254--269},
  year={2018}
}

@inproceedings{chen2023learning,
  title={Learning a sparse transformer network for effective image deraining},
  author={Chen, Xiang and Li, Hao and Li, Mingqiang and Pan, Jinshan},
  booktitle={Proceedings of the IEEE/CVF conference on computer vision and pattern recognition},
  pages={5896--5905},
  year={2023}
}

@article{xiao2022image,
  title={Image de-raining transformer},
  author={Xiao, Jie and Fu, Xueyang and Liu, Aiping and Wu, Feng and Zha, Zheng-Jun},
  journal={IEEE transactions on pattern analysis and machine intelligence},
  volume={45},
  number={11},
  pages={12978--12995},
  year={2022},
  publisher={IEEE}
}

@inproceedings{zamir2022restormer,
  title={Restormer: Efficient transformer for high-resolution image restoration},
  author={Zamir, Syed Waqas and Arora, Aditya and Khan, Salman and Hayat, Munawar and Khan, Fahad Shahbaz and Yang, Ming-Hsuan},
  booktitle={Proceedings of the IEEE/CVF conference on computer vision and pattern recognition},
  pages={5728--5739},
  year={2022}
}

@inproceedings{yi2021structure,
  title={Structure-preserving deraining with residue channel prior guidance},
  author={Yi, Qiaosi and Li, Juncheng and Dai, Qinyan and Fang, Faming and Zhang, Guixu and Zeng, Tieyong},
  booktitle={Proceedings of the IEEE/CVF international conference on computer vision},
  pages={4238--4247},
  year={2021}
}

@inproceedings{fu2021rain,
  title={Rain streak removal via dual graph convolutional network},
  author={Fu, Xueyang and Qi, Qi and Zha, Zheng-Jun and Zhu, Yurui and Ding, Xinghao},
  booktitle={Proceedings of the AAAI Conference on Artificial Intelligence},
  volume={35},
  number={2},
  pages={1352--1360},
  year={2021}
}

@inproceedings{zamir2021multi,
  title={Multi-stage progressive image restoration},
  author={Zamir, Syed Waqas and Arora, Aditya and Khan, Salman and Hayat, Munawar and Khan, Fahad Shahbaz and Yang, Ming-Hsuan and Shao, Ling},
  booktitle={Proceedings of the IEEE/CVF conference on computer vision and pattern recognition},
  pages={14821--14831},
  year={2021}
}

@inproceedings{wang2020model,
  title={A model-driven deep neural network for single image rain removal},
  author={Wang, Hong and Xie, Qi and Zhao, Qian and Meng, Deyu},
  booktitle={Proceedings of the IEEE/CVF conference on computer vision and pattern recognition},
  pages={3103--3112},
  year={2020}
}

@inproceedings{jiang2020multi,
  title={Multi-scale progressive fusion network for single image deraining},
  author={Jiang, Kui and Wang, Zhongyuan and Yi, Peng and Chen, Chen and Huang, Baojin and Luo, Yimin and Ma, Jiayi and Jiang, Junjun},
  booktitle={Proceedings of the IEEE/CVF conference on computer vision and pattern recognition},
  pages={8346--8355},
  year={2020}
}

@inproceedings{ren2019progressive,
  title={Progressive image deraining networks: A better and simpler baseline},
  author={Ren, Dongwei and Zuo, Wangmeng and Hu, Qinghua and Zhu, Pengfei and Meng, Deyu},
  booktitle={Proceedings of the IEEE/CVF conference on computer vision and pattern recognition},
  pages={3937--3946},
  year={2019}
}

@inproceedings{abuolaim2020defocus,
  title={Defocus deblurring using dual-pixel data},
  author={Abuolaim, Abdullah and Brown, Michael S},
  booktitle={European conference on computer vision},
  pages={111--126},
  year={2020},
  organization={Springer}
}

@inproceedings{abuolaim2022improving,
  title={Improving single-image defocus deblurring: How dual-pixel images help through multi-task learning},
  author={Abuolaim, Abdullah and Afifi, Mahmoud and Brown, Michael S},
  booktitle={Proceedings of the IEEE/CVF Winter Conference on Applications of Computer Vision},
  pages={1231--1239},
  year={2022}
}

@inproceedings{lee2021iterative,
  title={Iterative filter adaptive network for single image defocus deblurring},
  author={Lee, Junyong and Son, Hyeongseok and Rim, Jaesung and Cho, Sunghyun and Lee, Seungyong},
  booktitle={Proceedings of the IEEE/CVF conference on computer vision and pattern recognition},
  pages={2034--2042},
  year={2021}
}

@inproceedings{mehri2021mprnet,
  title={MPRNet: Multi-path residual network for lightweight image super resolution},
  author={Mehri, Armin and Ardakani, Parichehr B and Sappa, Angel D},
  booktitle={Proceedings of the IEEE/CVF winter conference on applications of computer vision},
  pages={2704--2713},
  year={2021}
}

@article{quan2021gaussian,
  title={Gaussian kernel mixture network for single image defocus deblurring},
  author={Quan, Yuhui and Wu, Zicong and Ji, Hui},
  journal={Advances in Neural Information Processing Systems},
  volume={34},
  pages={20812--20824},
  year={2021}
}

@inproceedings{quan2023neumann,
  title={Neumann network with recursive kernels for single image defocus deblurring},
  author={Quan, Yuhui and Wu, Zicong and Ji, Hui},
  booktitle={Proceedings of the IEEE/CVF conference on computer vision and pattern recognition},
  pages={5754--5763},
  year={2023}
}

@inproceedings{quan2023single,
  title={Single image defocus deblurring via implicit neural inverse kernels},
  author={Quan, Yuhui and Yao, Xin and Ji, Hui},
  booktitle={Proceedings of the IEEE/CVF international conference on computer vision},
  pages={12600--12610},
  year={2023}
}

@article{quan2024deep,
  title={Deep single image defocus deblurring via gaussian kernel mixture learning},
  author={Quan, Yuhui and Wu, Zicong and Xu, Ruotao and Ji, Hui},
  journal={IEEE Transactions on Pattern Analysis and Machine Intelligence},
  year={2024},
  publisher={IEEE}
}

@article{ruan2021aifnet,
  title={Aifnet: All-in-focus image restoration network using a light field-based dataset},
  author={Ruan, Lingyan and Chen, Bin and Li, Jizhou and Lam, Miu-Ling},
  journal={IEEE Transactions on Computational Imaging},
  volume={7},
  pages={675--688},
  year={2021},
  publisher={IEEE}
}

@inproceedings{ruan2022learning,
  title={Learning to deblur using light field generated and real defocus images},
  author={Ruan, Lingyan and Chen, Bin and Li, Jizhou and Lam, Miuling},
  booktitle={Proceedings of the IEEE/CVF conference on computer vision and pattern recognition},
  pages={16304--16313},
  year={2022}
}

@inproceedings{son2021single,
  title={Single image defocus deblurring using kernel-sharing parallel atrous convolutions},
  author={Son, Hyeongseok and Lee, Junyong and Cho, Sunghyun and Lee, Seungyong},
  booktitle={Proceedings of the IEEE/CVF international conference on computer vision},
  pages={2642--2650},
  year={2021}
}

@inproceedings{shen2019human,
  title={Human-aware motion deblurring},
  author={Shen, Ziyi and Wang, Wenguan and Lu, Xiankai and Shen, Jianbing and Ling, Haibin and Xu, Tingfa and Shao, Ling},
  booktitle={Proceedings of the IEEE/CVF international conference on computer vision},
  pages={5572--5581},
  year={2019}
}

@inproceedings{rim2020real,
  title={Real-world blur dataset for learning and benchmarking deblurring algorithms},
  author={Rim, Jaesung and Lee, Haeyun and Won, Jucheol and Cho, Sunghyun},
  booktitle={European conference on computer vision},
  pages={184--201},
  year={2020},
  organization={Springer}
}

@inproceedings{nah2017deep,
  title={Deep multi-scale convolutional neural network for dynamic scene deblurring},
  author={Nah, Seungjun and Hyun Kim, Tae and Mu Lee, Kyoung},
  booktitle={Proceedings of the IEEE conference on computer vision and pattern recognition},
  pages={3883--3891},
  year={2017}
}

@article{luan2024gyroscope,
  title={Gyroscope-Assisted Motion Deblurring Network},
  author={Luan, Simin and Yang, Cong and Boukhers, Zeyd and Qin, Xue and Cheng, Dongfeng and Sui, Wei and Li, Zhijun},
  journal={CoRR},
  year={2024}
}

@inproceedings{kupyn2018deblurgan,
  title={Deblurgan: Blind motion deblurring using conditional adversarial networks},
  author={Kupyn, Orest and Budzan, Volodymyr and Mykhailych, Mykola and Mishkin, Dmytro and Matas, Ji{\v{r}}{\'\i}},
  booktitle={Proceedings of the IEEE conference on computer vision and pattern recognition},
  pages={8183--8192},
  year={2018}
}

@inproceedings{kupyn2019deblurgan,
  title={Deblurgan-v2: Deblurring (orders-of-magnitude) faster and better},
  author={Kupyn, Orest and Martyniuk, Tetiana and Wu, Junru and Wang, Zhangyang},
  booktitle={Proceedings of the IEEE/CVF international conference on computer vision},
  pages={8878--8887},
  year={2019}
}

@inproceedings{zhang2020deblurring,
  title={Deblurring by realistic blurring},
  author={Zhang, Kaihao and Luo, Wenhan and Zhong, Yiran and Ma, Lin and Stenger, Bjorn and Liu, Wei and Li, Hongdong},
  booktitle={Proceedings of the IEEE/CVF conference on computer vision and pattern recognition},
  pages={2737--2746},
  year={2020}
}

@inproceedings{wang2022uformer,
  title={Uformer: A general u-shaped transformer for image restoration},
  author={Wang, Zhendong and Cun, Xiaodong and Bao, Jianmin and Zhou, Wengang and Liu, Jianzhuang and Li, Houqiang},
  booktitle={Proceedings of the IEEE/CVF conference on computer vision and pattern recognition},
  pages={17683--17693},
  year={2022}
}

@inproceedings{tsai2022stripformer,
  title={Stripformer: Strip transformer for fast image deblurring},
  author={Tsai, Fu-Jen and Peng, Yan-Tsung and Lin, Yen-Yu and Tsai, Chung-Chi and Lin, Chia-Wen},
  booktitle={European conference on computer vision},
  pages={146--162},
  year={2022},
  organization={Springer}
}

@inproceedings{luo2023image,
  title={Image restoration with mean-reverting stochastic differential equations},
  author={Luo, Ziwei and Gustafsson, Fredrik K and Zhao, Zheng and Sj{\"o}lund, Jens and Sch{\"o}n, Thomas B},
  booktitle={Proceedings of the 40th International Conference on Machine Learning},
  pages={23045--23066},
  year={2023}
}

@inproceedings{xia2023diffir,
  title={Diffir: Efficient diffusion model for image restoration},
  author={Xia, Bin and Zhang, Yulun and Wang, Shiyin and Wang, Yitong and Wu, Xinglong and Tian, Yapeng and Yang, Wenming and Van Gool, Luc},
  booktitle={Proceedings of the IEEE/CVF international conference on computer vision},
  pages={13095--13105},
  year={2023}
}

@article{chen2023hierarchical,
  title={Hierarchical integration diffusion model for realistic image deblurring},
  author={Chen, Zheng and Zhang, Yulun and Liu, Ding and Gu, Jinjin and Kong, Linghe and Yuan, Xin and others},
  journal={Advances in neural information processing systems},
  volume={36},
  pages={29114--29125},
  year={2023}
}

@article{wang2026zero,
  title={Zero-shot realistic image deblurring with consistency model},
  author={Wang, Zhaohan and Chen, Chengjun and Dai, Chenggang},
  journal={Complex \& Intelligent Systems},
  volume={12},
  number={1},
  pages={29},
  year={2026},
  publisher={Springer}
}

@article{li2018benchmarking,
  title={Benchmarking single-image dehazing and beyond},
  author={Li, Boyi and Ren, Wenqi and Fu, Dengpan and Tao, Dacheng and Feng, Dan and Zeng, Wenjun and Wang, Zhangyang},
  journal={IEEE transactions on image processing},
  volume={28},
  number={1},
  pages={492--505},
  year={2018},
  publisher={IEEE}
}

@inproceedings{qu2017deshadownet,
  title={Deshadownet: A multi-context embedding deep network for shadow removal},
  author={Qu, Liangqiong and Tian, Jiandong and He, Shengfeng and Tang, Yandong and Lau, Rynson WH},
  booktitle={Proceedings of the IEEE conference on computer vision and pattern recognition},
  pages={4067--4075},
  year={2017}
}

@article{fattal2014dehazing,
  title={Dehazing using color-lines},
  author={Fattal, Raanan},
  journal={ACM transactions on graphics (TOG)},
  volume={34},
  number={1},
  pages={1--14},
  year={2014},
  publisher={ACM New York, NY, USA}
}

@inproceedings{cai2019toward,
  title={Toward real-world single image super-resolution: A new benchmark and a new model},
  author={Cai, Jianrui and Zeng, Hui and Yong, Hongwei and Cao, Zisheng and Zhang, Lei},
  booktitle={Proceedings of the IEEE/CVF International Conference on Computer Vision (ICCV)},
  pages={3086--3095},
  year={2019}
}

@inproceedings{wei2020component,
  title={Component divide-and-conquer for real-world image super-resolution},
  author={Wei, Pengxu and Xie, Ziwei and Lu, Hannan and Zhan, Zongyuan and Ye, Qixiang and Zuo, Wangmeng and Lin, Liang},
  booktitle={European Conference on Computer Vision (ECCV)},
  pages={101--117},
  year={2020}
}

@article{dong2014learning,
  title={Learning a deep convolutional network for image super-resolution},
  author={Dong, Chao and Loy, Chen Change and He, Kaiming and Tang, Xiaoou},
  journal={European conference on computer vision (ECCV)},
  year={2014}
}

@inproceedings{zhang2018image,
  title={Image super-resolution using very deep residual channel attention networks},
  author={Zhang, Yulun and Li, Kunpeng and Li, Kai and Wang, Lichen and Zhong, Bineng and Fu, Yun},
  booktitle={Proceedings of the European conference on computer vision (ECCV)},
  pages={286--301},
  year={2018}
}

@inproceedings{goodfellow2014generative,
  title={Generative adversarial nets},
  author={Goodfellow, Ian and Pouget-Abadie, Jean and Mirza, Mehdi and Xu, Bing and Warde-Farley, David and Ozair, Sherjil and Courville, Aaron and Bengio, Yoshua},
  booktitle={Advances in neural information processing systems (NeurIPS)},
  year={2014}
}

@inproceedings{ledig2017photo,
  title={Photo-realistic single image super-resolution using a generative adversarial network},
  author={Ledig, Christian and Theis, Lucas and Husz{\'a}r, Ferenc and Caballero, Jose and Cunningham, Andrew and Acosta, Alejandro and Aitken, Alykhan and Tejani, Alykhan and Totz, Johannes and Wang, Zehan and others},
  booktitle={Proceedings of the IEEE conference on computer vision and pattern recognition (CVPR)},
  pages={4681--4690},
  year={2017}
}

@inproceedings{zhang2021designing,
  title={Designing a practical degradation model for deep blind image super-resolution},
  author={Zhang, Kai and Liang, Jingyun and Van Gool, Luc and Timofte, Radu},
  booktitle={Proceedings of the IEEE/CVF International Conference on Computer Vision (ICCV)},
  pages={4791--4800},
  year={2021}
}

@inproceedings{wang2021real,
  title={Real-esrgan: Training real-world blind super-resolution with pure synthetic data},
  author={Wang, Xintao and Xie, Liangbin and Dong, Chao and Shan, Ying},
  booktitle={Proceedings of the IEEE/CVF International Conference on Computer Vision (ICCV) Workshops},
  pages={1905--1914},
  year={2021}
}

@inproceedings{ho2020denoising,
  title={Denoising diffusion probabilistic models},
  author={Ho, Jonathan and Jain, Ajay and Abbeel, Pieter},
  booktitle={Advances in Neural Information Processing Systems (NeurIPS)},
  volume={33},
  pages={6840--6851},
  year={2020}
}

@article{wang2023exploiting,
  title={Exploiting diffusion prior for real-world image super-resolution},
  author={Wang, Jianyi and Yue, Zongsheng and Zhou, Shangchen and Chan, Kelvin CK and Loy, Chen Change},
  journal={International Journal of Computer Vision (IJCV)},
  pages={1--21},
  year={2024}
}

@inproceedings{lin2023diffbir,
  title={Diffbir: Towards blind image restoration with generative diffusion prior},
  author={Lin, Xinqi and He, Jingwen and Chen, Ziyan and Lyu, Zhaoyang and Dai, Bo and Yu, Fanghua and Ouyang, Wanli and Qiao, Yu and Dong, Chao},
  booktitle={arXiv preprint arXiv:2308.15070},
  year={2023}
}

@inproceedings{wang2024sinsr,
  title={Sinsr: diffusion-based image super-resolution in a single step},
  author={Wang, Yufei and Yang, Wenhan and Chen, Xinyuan and Wang, Yaohui and Guo, Lanqing and Chau, Lap-Pui and Liu, Ziwei and Qiao, Yu and Kot, Alex C and Wen, Bihan},
  booktitle={Proceedings of the IEEE/CVF Conference on Computer Vision and Pattern Recognition (CVPR)},
  pages={25796--25805},
  year={2024}
}

@inproceedings{blau2018perception,
  title={The perception-distortion tradeoff},
  author={Blau, Yochai and Michaeli, Tomer},
  booktitle={Proceedings of the IEEE conference on computer vision and pattern recognition (CVPR)},
  pages={6228--6237},
  year={2018}
}

@inproceedings{agustsson2017ntire,
  title={Ntire 2017 challenge on single image super-resolution: Dataset and study},
  author={Agustsson, Eirikur and Timofte, Radu},
  booktitle={Proceedings of the IEEE conference on computer vision and pattern recognition workshops (CVPRW)},
  pages={126--135},
  year={2017}
}

@article{wang2004image,
  title={Image quality assessment: from error visibility to structural similarity},
  author={Wang, Zhou and Bovik, Alan C and Sheikh, Hamid R and Simoncelli, Eero P},
  journal={IEEE transactions on image processing},
  volume={13},
  number={4},
  pages={600--612},
  year={2004}
}

@inproceedings{zhang2018unreasonable,
  title={The unreasonable effectiveness of deep features as a perceptual metric},
  author={Zhang, Richard and Isola, Phillip and Efros, Alexei A and Shechtman, Eli and Wang, Oliver},
  booktitle={Proceedings of the IEEE conference on computer vision and pattern recognition (CVPR)},
  pages={586--595},
  year={2018}
}

@article{mittal2012making,
  title={Making a completely blind image quality analyzer},
  author={Mittal, Anish and Soundararajan, Rajiv and Bovik, Alan C},
  journal={IEEE Signal processing letters},
  volume={20},
  number={3},
  pages={209--212},
  year={2012}
}

@inproceedings{wang2023exploring,
  title={Exploring clip for assessing the look and feel of images},
  author={Wang, Jianyi and Chan, Kelvin CK and Loy, Chen Change},
  booktitle={Proceedings of the AAAI Conference on Artificial Intelligence},
  volume={37},
  number={2},
  pages={2555--2563},
  year={2023}
}

@article{team2023gemini,
  title={Gemini: a family of highly capable multimodal models},
  author={Team, Gemini and Anil, Rohan and Borgeaud, Sebastian and Alayrac, Jean-Baptiste and Yu, Jiahui and Soricut, Radu and Schalkwyk, Johan and Dai, Andrew M and Hauth, Anja and Millican, Katie and others},
  journal={arXiv preprint arXiv:2312.11805},
  year={2023}
}

\end{document}